\documentclass{article}

\usepackage[preprint]{neurips_2026}

\usepackage[utf8]{inputenc} 
\usepackage[T1]{fontenc}    
\usepackage{hyperref}       
\usepackage{url}            
\usepackage{booktabs}       
\usepackage{amsfonts}       
\usepackage{amsmath}        
\usepackage{amssymb}        
\usepackage{amsthm}         
\usepackage{mathtools}      
\usepackage{nicefrac}       
\usepackage{microtype}      
\usepackage{xcolor}         
\usepackage{graphicx}       
\newlength{\rdsfigh}
\newcommand{\rdsstretched}[1]{%
  \settoheight{\rdsfigh}{\includegraphics[width=0.9\linewidth]{#1}}%
  \includegraphics[width=0.9\linewidth,height=1.05\rdsfigh,keepaspectratio=false]{#1}%
}
\usepackage{tikz}           
\usetikzlibrary{arrows.meta,positioning,calc}

\title{Decomposing how prompting steers behavior}
\author{%
  Fan L.~Cheng
  \\Columbia University
  \\fan.cheng@columbia.edu
  \And
  Nikolaus Kriegeskorte
  \\Columbia University
  \\n.kriegeskorte@columbia.edu
}

\begin{document}

\maketitle

\begin{abstract}
Prompting steers large language models (LLMs) and vision--language models
(VLMs) without weight updates, but it remains unclear how a change in
instruction reshapes internal representations to produce a behavioral effect. We introduce a nested geometric decomposition framework that treats prompting as a transformation of the representational geometry for the content following the prompt. We ask what class of mathematical transformation best explains the effect of prompting by finding the best alignment between representations of the same stimulus set following different prompts. For each prompt pair, we fit a sequence of increasingly
expressive stimulus-invariant maps: translation, rigid transformation with uniform-scaling, sequential axis scaling, affine, and nonlinear transformations. We then test these maps causally by replacing a single layer's prompt-A hidden state for a new set of stimuli with its mapped counterpart and measuring recovery of prompt-B representational geometry and behavior. Across three LLMs, three VLMs, and six text or image datasets varying in style, emotion, scene content, and number, prompts consistently reshape representational geometry toward the instructed task structure. In the cross-validated nested variance decomposition, much of the prompt-induced activation change is explained by shape-preserving maps: translation and rigid transformation with uniform-scaling. The tier profiles reveal model- and task-specific routing strategies, differing in how much transformation classes explain variance and where along the layer hierarchy their contributions emerge. Crucially, although translation and rigid transformation tiers already improve behavioral agreement, affine transformation is the first tier to nearly recover target-prompt task geometry and produces corresponding gains in behavioral agreement. This suggests that cross-dimensional linear mixing may be a key contributor to how prompts reorganize representations toward the instructed task structure. Our framework provides a general way to decompose prompt-induced representational change into interpretable geometric components, revealing how a model routes task-relevant structure to produce prompt-driven behavior.
\end{abstract}

\section{Introduction}
\label{sec:intro}

Instruction prompts can alter model behavior at inference time without
updating parameters. For a fixed stimulus, changing the instruction from
``describe the object category'' to ``describe the artistic style,'' for example,
can shift which stimulus dimensions become behaviorally relevant and,
consequently, which output the model produces. This raises a mechanistic
question: how does a change in instruction reorganize internal
representations so as to steer downstream behavior?

Two complementary literatures bear on this question. First,
representational analyses show that prompts and context restructure
internal geometry, producing measurable changes in representational
similarity, manifold capacity, trajectory geometry, probing performance, and related
diagnostics
\citep{kirsanov2025geometry,hosseini2026context,park2025icl,polo2026emergent,gonzalezgutierrez2025do,davidson2025common,park2024linear}.
These effects can be depth-dependent, with some studies suggesting that intermediate layers provide especially informative or compressed representations and later layers more directly support prediction \citep{jiang2025compression,skean2024intermediate,skean2025layer}.
Second, work on activation steering shows that relatively simple
activation-space interventions, such as additive vectors, rotations,
and affine maps, can modulate prompt-induced behavioral
effects \citep{turner2023activation,zou2023representation,singh2024affine,vu2025angular}.
Taken together, these findings suggest that natural prompting may itself
be implemented through a low-complexity transformation in activation
space.

\begin{figure}[t]
  \centering
  \includegraphics[width=\linewidth]{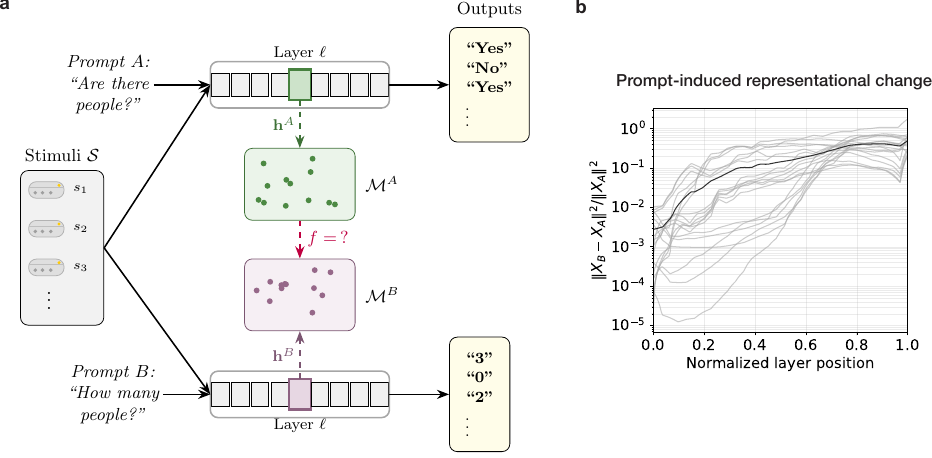}
  \caption{\textbf{(a)} Two prompts $A$ (``Are there people?'') and
  $B$ (``How many people?'') are presented to the same model together
  with a stimulus set $\mathcal{S}$. We tap the hidden state at a single
  transformer layer $\ell$ (highlighted) to obtain the layer-$\ell$
  manifolds $\mathcal{M}^{A}=\Phi^{A}(\mathcal{S})$ (green) and
  $\mathcal{M}^{B}=\Phi^{B}(\mathcal{S})$ (purple). The same forward pass continues past
  $\ell$ and yields a prompt-specific output per stimulus (``Yes/No''
  vs.\ counts). We ask whether a systematic map
  $f:\mathcal{M}^{A}\to\mathcal{M}^{B}$ exists, how complex it has to
  be, and whether it causally drives behaviour.
  \textbf{(b)}~Prompt-induced representational change as a function of normalised layer depth. Each gray curve is one pair of model and dataset from the three LLMs (OPT-2.7B, Llama3-8B, Qwen3-8B) on three text datasets and three VLMs (BLIP-2, LLaVA-OneVision, Qwen3-VL) on three image datasets; the black curve is their mean. The change is small but non-zero in early layers ($10^{-4}$--$10^{-2}$) and grows nearly monotonically with depth, reaching $10^{-1}$--$10^{0}$ at the top, with a consistent shape across architectures and modalities.}
  \label{fig:setup}
\end{figure}

What remains unclear, however, is the \emph{transformation class} of the
prompt-induced change. Existing work on the geometry of prompting has
established that prompts alter internal representations, but it has
typically characterized those changes through scalar metrics rather than decomposing them
into interpretable geometric components. Conversely, activation-steering
studies show that engineered low-complexity operators can influence
behavior, but they rarely estimate such operators from natural
prompting itself or test whether the recovered operators account for target-prompt geometry induced by an instruction change.


We address this gap with a nested geometric decomposition framework. For
each model, layer, dataset, and prompt pair, we fit a single
stimulus-invariant alignment map from source-prompt activations to
target-prompt activations and evaluate it on held-out stimuli. The
operator hierarchy separates centroid shifts, global Procrustes
alignment, axis-wise scaling in the Procrustes-aligned basis,
cross-dimensional feature mixing, and nonlinear residual structure. This
hierarchy is motivated by classical Procrustes analysis and by
shape-based approaches to comparing representations under explicit
transformation classes
\citep{gower1975procrustes,williams2021generalized,kornblith2019cka,williams2024equivalence,harvey2024similarity,barbosa2025shape}.
In parallel, we use RSA and silhouette analyses to test whether
prompting reshapes representational geometry toward the instructed task
structure; RSA compares representational dissimilarity matrices and is
designed to relate representational geometry to candidate task or model
structures \citep{kriegeskorte2008rsa,lin2024topological,cheng2025interpreting}. Finally, we
apply each fitted map as a causal activation intervention, replacing a
source-prompt hidden state with its mapped counterpart and measuring
whether the resulting representation and output recover those of the
target prompt. The resulting tier profiles reveal the prompt-routing
strategy a model uses to follow an instruction: which transformation
tiers carry the change, where they emerge across layers, and whether
they recover task-aligned geometry and behavior.

\paragraph{Contributions.}
\begin{enumerate}
\item We introduce a \emph{nested geometric decomposition} of prompt-induced hidden-state change across five transformation classes: translation, rigid transformation with uniform scaling, rigid transformation with axis-wise scaling, affine, and nonlinear, yielding a unique contribution for each tier (\S\ref{sec:method}).
\item We develop a \emph{causal intervention protocol} that applies the fitted map from each tier at a single layer on held-out stimuli and evaluates both representational geometry and behavioural recovery relative to the target prompt (\S\ref{sec:experiments}).
\item Across LLMs, VLMs, and six datasets, we find that prompting consistently reshapes representational geometry toward the instructed task structure, that much of the prompt effect is captured by low-complexity maps, and that model families, datasets, and prompt-pair groups differ systematically in the relative contribution of transformation tiers and in where these contributions emerge across depth, revealing distinct prompt-routing strategies (\S\ref{sec:results}).
\end{enumerate}

\section{Related work}
\label{sec:related}
\paragraph{Geometry of prompting and in-context representations.}
Prior work has characterized the geometric effects of prompting and
in-context structure primarily through changes in scalar metrics. A previous study analyzes zero-shot, few-shot, and soft prompting using manifold-capacity tools, showing that different prompting regimes can induce distinct representational mechanisms for task adaptation \citep{kirsanov2025geometry}. Other work characterizes how context and in-context examples reorganize internal representations into compact, identifiable substrates that drive in-context generalization, sucha as attention-head circuits, task vectors, and trajectory or phase-transition-like geometry \citep{hosseini2026context,park2025icl,yang2025unifying,hendel2023icl,todd2024fv}.
A complementary line documents the dynamic, distributed, and not always task-aligned representational effects of prompting \citep{gonzalezgutierrez2025do,davidson2025common,justintime2025tasks,polo2026emergent,rema2025,simhi2026old}.
Relative to this line of work, our contribution is to estimate explicit
transformation classes between prompt-conditioned representation
manifolds and to test whether the transformations causally
recover both representational geometry and behavior.

\paragraph{Activation steering and geometric interventions.}
We organize prior steering methods within the transformation hierarchy
$\mathcal{F}_{\mathrm{T}}\subset\mathcal{F}_{\mathrm{O_u}}\subset\mathcal{F}_{\mathrm{O_a}}\subset\mathcal{F}_{\mathrm{L}}\subset\mathcal{F}_{\mathrm{N}}$
(Fig.~\ref{fig:r2_decomposition}), distinguishing \emph{global}
operators, which is stimulus-invariant, from
\emph{input-conditioned} interventions, whose effective transformation
depends on the current stimuli. Most existing methods are naturally
described at the translation tier, where a stimulus-invariant direction
is added to the hidden state
\citep{turner2023activation,rimsky2024caa,li2023iti,zou2023representation,templeton2024scaling,wang2025srps,davidson2025common,singh2024affine}.
A smaller group of methods fits a single global linear or affine intervention \citep{postmus2024conceptors,sheng2026alphasteer,wu2024reft,wu2025reps,singh2024affine}.
Rotation-based methods and piecewise-affine optimal-transport edits are
also relevant, but they are generally not estimated as Procrustes maps
between paired representation clouds, and their target-angle or adaptive
variants are often input-conditioned
\citep{vu2025angular,pham2024householder,abdullaev2026chars,rodriguez2025act,scialanga2025sake}. The rigid transformation with uniform scaling $\mathcal{F}_{\mathrm{O_u}}$ therefore
remains comparatively underexplored as a stimulus-set mapping
estimator, while strictly nonlinear interventions appear relatively
rare \citep{zhao2026odesteer,raval2026curveball}. Independently of the
operator class, steering methods also differ in the stage of the
forward pass at which the intervention is applied
\citep{lee2025cast,nguyen2026pid,dang2026selective}, with a related line
operating in weight space rather than activation space
\citep{fierro2025weight},
and recent audits have argued for more systematic geometric and
mediator-typology characterizations
\citep{venkatesh2026nonidentifiability,tan2024generalize,imli2025unified,wehner2025taxonomy}.
In this paper, our goal is not to engineer operators that
induce a desired behavior, but to use the same hierarchy as a
measurement framework for natural prompting. The finding that
low-complexity linear maps recover most of the prompt-induced effect is
consistent with the strong empirical performance of translation- and
affine-class steering methods, which may succeed in part because they
approximate the activation changes and representational geometry induced by prompting itself.

\begin{figure}[!t]
  \centering
  \includegraphics[width=0.92\linewidth]{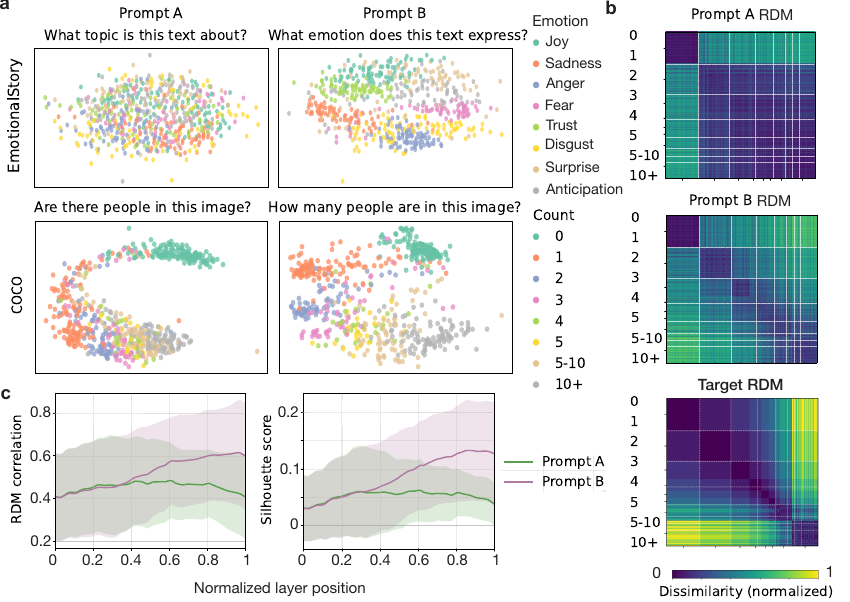}
\caption{\textbf{Prompting reshapes representational geometry toward the instructed task structure.}
\textbf{(a)} Multidimensional scaling (MDS) visualizations of prompt-conditioned representations.
(Top) Layer-32 representations from Llama3-8B-Instruct for 1{,}920 text stories under a topic prompt (Prompt A) and an emotion prompt (Prompt B).
(Bottom) Layer-27 representations from LLaVA-OneVision for 1{,}000 COCO images under a person-detection prompt (Prompt A) and a person-counting prompt (Prompt B).
Each point denotes one stimulus and is colored by the target label for Prompt B.\textbf{(b)} Representational dissimilarity matrix (RDM) for the same model, layer, and image set as in the bottom row of panel (a). The top two RDMs are computed from the prompt-A and prompt-B hidden states, respectively. The bottom RDM is the target task RDM for prompt B, constructed from numerical distances between count labels. \textbf{(c)} Layerwise alignment with the target task structure (pooled across models and datasets). Left: Spearman correlation between each prompt-induced RDM and the corresponding target RDM, computed separately for prompt A and prompt B. Right: silhouette score using the target labels, which measures whether stimuli with the same target label form compact, well-separated clusters in representation space. The shaded regions show $\pm1$ SD. }
  \label{fig:rep_structure}
\end{figure}

\section{Method: nested geometric decomposition}
\label{sec:method}

\paragraph{Preliminaries.}
Let $\mathcal{S}$ denote the stimulus space (a measurable set of images
or text excerpts), and let $\mathcal{P}$ be a finite set of prompts.
Let $M$ be a (vision--)language model with $L$ transformer blocks; fix
an analysis layer $\ell\in\{1,\dots,L\}$.
For each prompt $p\in\mathcal{P}$, the model defines a
\emph{feature map}
\begin{equation}
  \Phi^{p} \;:\; \mathcal{S} \;\longrightarrow\; \mathbb{R}^{D},
  \qquad
  s \;\longmapsto\; \mathbf{h}^{p}(s),
\end{equation}
returning the layer-$\ell$ hidden state at the final input-prompt token.
We call its image
\[
  \mathcal{M}^{p}\;\coloneqq\;\Phi^{p}(\mathcal{S})\;\subset\;\mathbb{R}^{D}
\]
the \emph{manifold} of prompt $p$. The ambient space
$\mathbb{R}^{D}$ carries the canonical Euclidean inner product and
the induced Frobenius metric on $\mathbb{R}^{N\times D}$.

\paragraph{Paired observations across prompts.}
Fix a pair of prompts $A, B\in\mathcal{P}$ and a finite stimulus set
$\{s_{i}\}_{i=1}^{N}\subset\mathcal{S}$ (indices $i,j$ range over
stimuli). For each stimulus $i$ we obtain a paired observation
$\big(\Phi^{A}(s_{i}),\,\Phi^{B}(s_{i})\big)\in\mathbb{R}^{D}\times\mathbb{R}^{D}$,
and stack these into representation matrices
\begin{equation}
  \mathbf{X}^{A},\;\mathbf{X}^{B}\;\in\;\mathbb{R}^{N\times D},
  \qquad
  \big[\mathbf{X}^{p}\big]_{i,:}\;=\;\Phi^{p}(s_{i})^{\!\top}.
\end{equation}

\paragraph{Problem statement.}
We ask whether there is a map $f:\mathbb{R}^{D}\!\to\!\mathbb{R}^{D}$
that approximately satisfies $f\circ\Phi^{A}\approx\Phi^{B}$ on
$\mathcal{S}$ (Fig.~\ref{fig:setup}\textbf{a}). We call a fitted map \emph{global} if its parameters are \emph{stimulus-invariant}. We contrast
this with \emph{input-conditioned} steering rules, whose effective
direction, angle, coefficient, or application mask varies with the
current stimuli. We ask how expressive a global map must be before
it can explain, and causally reproduce, the representational effect of
changing prompts.

\paragraph{Hypothesis classes.}
We approximate $f$ by elements of an increasing chain of transformation families on \(\mathbb{R}^{D}\), ranging from classical Lie-group actions to more general affine and nonlinear families:
\begin{equation}
\small
\begin{gathered}
  \underbrace{\mathbb{R}^{D}}_{\text{translations }T(D)}
  \;\hookrightarrow\;
  \underbrace{\mathbb{R}^{D}\rtimes\big(\mathbb{R}_{>0}\!\times\! O(D)\big)}_{\text{similarity group }\mathrm{Sim}(D)}
  \;\hookrightarrow\;
  \underbrace{\mathbb{R}^{D}\rtimes\big(\mathrm{Diag}(D)\!\times\! O(D)\big)}_{\text{axis-scaled rigid maps}}
  \;\hookrightarrow\;
  \underbrace{\mathbb{R}^{D}\rtimes\mathrm{GL}(D)}_{\text{affine group }\mathrm{Aff}(D)}
  \;\hookrightarrow\;
  \underbrace{\mathcal{F}_{\mathrm{N}}}_{\text{nonlinear maps}}.
\end{gathered}
\end{equation}
Each level corresponds to a hypothesis class of mappings on hidden vectors
$\mathbf{x}\in\mathbb{R}^{1\times D}$, with free intercept parameters
$\mathbf{a},\mathbf{b}\in\mathbb{R}^{D}$:
\begin{center}
\renewcommand{\arraystretch}{1.55}
\resizebox{\linewidth}{!}{%
\begin{tabular}{@{}lll@{}}
\toprule
Class $\mathcal{F}_{k}$ & $f_{k}(\mathbf{x})$ & Parameters \\
\midrule
$\mathcal{F}_{\mathrm{T}}$ : translation
   & $\mathbf{x} + \mathbf{b}$
   & $\mathbf{b}\in\mathbb{R}^{D}$ \\
$\mathcal{F}_{\mathrm{O_{u}}}$ : rigid transformation and uniform scaling
   & $\mathbf{b} + c\,(\mathbf{x}-\mathbf{a})\,\mathbf{Q}$
   & $\mathbf{a},\mathbf{b}\in\mathbb{R}^{D},\;\mathbf{Q}\in O(D),\; c\in\mathbb{R}_{>0}$ \\
$\mathcal{F}_{\mathrm{O_{a}}}$ : rigid transformation and axis-wise scaling
   & $\mathbf{b} + (\mathbf{x}-\mathbf{a})\,\mathbf{Q}\,\mathbf{D}$
   & $\mathbf{a},\mathbf{b}\in\mathbb{R}^{D},\;\mathbf{Q}\in O(D),\;\mathbf{D}=\mathrm{diag}(d_{1},\dots,d_{D})$ \\
$\mathcal{F}_{\mathrm{L}}$ : affine transformation
   & $\mathbf{b} + (\mathbf{x}-\mathbf{a})\,\mathbf{M}$
   & $\mathbf{a},\mathbf{b}\in\mathbb{R}^{D},\;\mathbf{M}\in\mathbb{R}^{D\times D}$ \\
$\mathcal{F}_{\mathrm{N}}$ : nonlinear transformation
   & $\mathbf{b} + g_{\boldsymbol{\theta}}(\mathbf{x}-\mathbf{a})$
   & $\mathbf{a},\mathbf{b}\in\mathbb{R}^{D},\;\boldsymbol{\theta}\in\mathbb{R}^{P}$ \\
\bottomrule
\end{tabular}%
}
\end{center}
\noindent
$\mathcal{F}_{\mathrm{N}}$ is instantiated in our experiments as a
single shared multilayer perceptron
$g_{\boldsymbol{\theta}}:\mathbb{R}^{D}\!\to\!\mathbb{R}^{D}$;
reported $\Delta R^{2}_{\mathrm{N}}$ should therefore be read as a
lower bound on what an arbitrarily expressive nonlinear map could
explain. We use the right-multiplication convention
$\mathbf{x}\mathbf{M}$ to match the row-major data layout. The five
classes form a strictly nested chain
$\mathcal{F}_{\mathrm{T}}\subset\mathcal{F}_{\mathrm{O_u}}\subset\mathcal{F}_{\mathrm{O_a}}\subset\mathcal{F}_{\mathrm{L}}\subset\mathcal{F}_{\mathrm{N}}$;
$\mathcal{F}_{\mathrm{O_a}}$ preserves the rigid alignment found by
$\mathcal{F}_{\mathrm{O_u}}$ but allows each aligned axis to be
independently rescaled, isolating axis-wise gain modulation from
arbitrary linear feature mixing. Inclusion identifications are in
Appendix~\ref{app:methods}.

\paragraph{Parameter estimation.}
Each class is fit on the training fold by minimizing the squared
Frobenius reconstruction error
\begin{equation}
  \widehat{f}_{k}\;=\;
  \arg\min_{f\in\mathcal{F}_{k}}\;
  \big\|\,\mathbf{X}^{B}-f(\mathbf{X}^{A})\,\big\|_{F}^{2},
\end{equation}
after centring each prompt condition by its training-fold mean.
$\mathcal{F}_{\mathrm{T}}$ has a closed-form mean-shift solution;
$\mathcal{F}_{\mathrm{O_u}}$ is solved by orthogonal Procrustes,
$\mathcal{F}_{\mathrm{O_a}}$ by Procrustes followed by per-axis least
squares on the rigidly-aligned features, and $\mathcal{F}_{\mathrm{L}}$
by ridge regression. The nonlinear class $\mathcal{F}_{\mathrm{N}}$ is
instantiated as a shared one-hidden-layer MLP fit by stochastic
gradient descent on the same Frobenius criterion. All estimators are
evaluated under stratified $K$-fold cross-validation across stimuli.
Closed-form derivations, the MLP capacity and
optimizer, and the cross-validation procedure are in
Appendix~\ref{app:methods}.

\paragraph{Causal intervention.}
\label{method:intervention}
To probe whether the fitted geometric transformation also \emph{causally}
reproduces the effect of prompt $B$, we run model on stimulus
$s_{i}$ under prompt $A$ and replace the layer-$\ell$ hidden state at
the final input token by $\widehat{f}_{k}\!\big(\Phi^{A}(s_{i})\big)$
before continuing the autoregressive forward pass. Let $y_{k}(s_{i})$
denote the resulting output token sequence. As a no-fit oracle
reference we also include a level that patches the
held-out prompt-$B$ representation directly,
$\widehat{f}_{\mathrm{prompt_B}}(\Phi^{A}(s_{i})):=\Phi^{B}(s_{i})$, with
the surrounding prompt-$A$ context unchanged. This is distinct from
running model end-to-end under prompt $B$: the prompt tokens, the
attention pattern up to layer $\ell$, and the post-layer-$\ell$
processing all start from prompt $A$'s context, so $y_{\mathrm{prompt_B}}$
upper-bounds what \emph{any} fitted single-layer replacement
can recover from prompt $A$ alone.

\section{Evaluation}
For each transform $k\in\{\mathrm{T},\mathrm{O_u},\mathrm{O_a},\mathrm{L},\mathrm{N}\}$ we report three families of measures:
\begin{enumerate}
  \item \textbf{Incremental explained variance:} Let
    $\mathrm{RSS}_{k}\;\coloneqq\;\big\|\mathbf{X}^{B}-\widehat{f}_{k}(\mathbf{X}^{A})\big\|_{F}^{2}$
    on held-out stimuli for $k\in\{\mathrm{T},\mathrm{O_u},\mathrm{O_a},\mathrm{L},\mathrm{N}\}$,
    and let $\mathrm{RSS}_{0}\coloneqq\big\|\mathbf{X}^{B}-\mathbf{X}^{A}\big\|_{F}^{2}$
    be the no-transform residual. We define the \emph{cumulative}
    cross-validated $R^{2}$ of tier $k$ as \(R^{2}_{k} = (\mathrm{RSS}_{0}-\mathrm{RSS}_{k})/\mathrm{RSS}_{0}\) and the \emph{incremental} $R^{2}$ contribution of tier $k$ over its predecessor $k-1$ in the nested chain as $\Delta R^{2}_{k}\;=\;R^{2}_{k}-R^{2}_{k-1}$. These increments together with the residual unexplained fraction $R^{2}_{\mathrm{resid}}=\mathrm{RSS}_{\mathrm{N}}/\mathrm{RSS}_{0}$
    sum to $1$ by construction. Individual increments may be negative
    under cross-validation when a more expressive class generalizes
    worse on held-out stimuli;
  \item \textbf{representational geometry metrics:}
    Spearman correlation between the data RDM of
    $\widehat{f}_{k}(\mathbf{X}^{A})$ and a category-derived target RDM
    over $\{s_{i}\}$ based on the task structure of prompt B, and the silhouette score
    $s_{i}=(b_{i}-a_{i})/\max(a_{i},b_{i})$, where $a_{i}$ is the mean
    within-category distance and $b_{i}$ is the mean distance to the
    nearest other category;
  \item \textbf{behavioral recovery:}
    a per-dataset keyword evaluator scores each intervened output
    $y_{k}(s_{i})$ on prompt $B$'s ground-truth attribute label
    (e.g.\ presence of the target style word, correct numeric count),
    yielding two metrics: \emph{relevance} (does the text address
    the target attribute) and \emph{accuracy} (is the answer correct
    for that stimulus). The oracle-patched output
    $y_{\mathrm{prompt_B}}(s_{i})$ provides the upper-bound reference.
\end{enumerate}

\begin{figure}[!t]
  \centering
  \includegraphics[width=\linewidth]{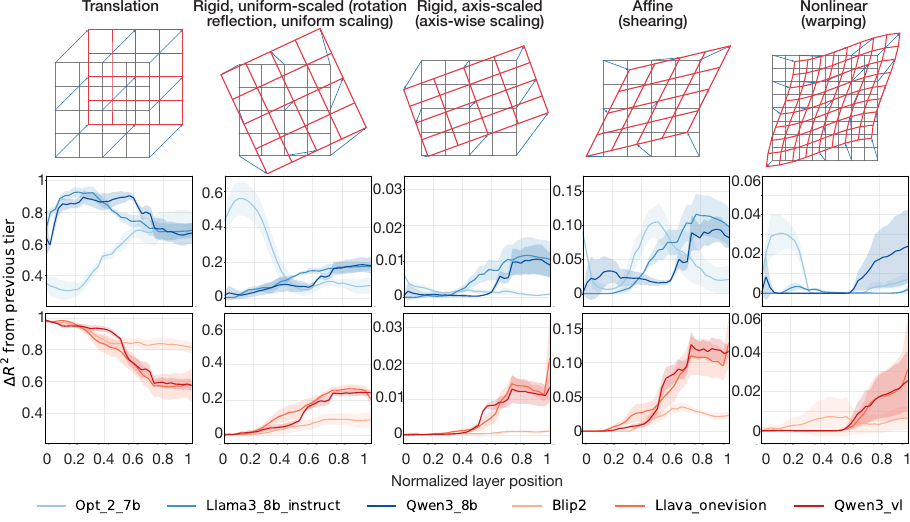}
  \caption{Nested geometric decomposition of prompt-induced representational maps: translation ($\mathcal{F}_{\mathrm{T}}$), rigid transformation with uniform scaling ($\mathcal{F}_{\mathrm{O_u}}$), rigid transformation with axis-wise scaling ($\mathcal{F}_{\mathrm{O_a}}$), affine transformation ($\mathcal{F}_{\mathrm{L}}$), and nonlinear transformation($\mathcal{F}_{\mathrm{N}}$). Top: Schematic of the additional geometric freedom introduced by each tier (listed in parentheses); blue and red grids show the source and transformed representations, respectively. Middle: Incremental explained variance over the preceding tier, $\Delta R^{2}$, as a function of normalised layer depth for LLMs, pooled across datasets and prompt pairs. Bottom: The same decomposition for VLMs. Lines denote model-specific means across prompt pairs; shaded bands denote $\pm 1$ SD.}
  \label{fig:r2_decomposition}
\end{figure}

\section{Experiments}
\label{sec:experiments}

\paragraph{Models.}
We evaluate six open-weight transformer models.
On the language models: \textbf{OPT-2.7B},
\textbf{Meta-Llama-3-8B-Instruct}, and \textbf{Qwen3-8B}. On the vision language models: \textbf{BLIP-2} (OPT-2.7B backbone, Q-Former bridging), \textbf{LLaVA-OneVision-7B}
(vision encoder/projector with Qwen2 language backbone, multimodal instruction-tuned), and
\textbf{Qwen3-VL-8B} (native vision-language model with interleaved text-image/video pretraining).

\paragraph{Datasets.}
The three text datasets (EmotionalStory, WritingStyle, Number) are used by the LLMs; the three
image datasets (EmoSet, StyleTransfer, COCO) are used by the VLMs. Each
dataset has a two-attribute factorial structure ($\text{attr}_A\!\times\!\text{attr}_B$) that the prompt taxonomy exploits for cross- and within-attribute comparisons. Dataset sizes,
attribute levels, construction procedures, and prompt templates are in
Appendix~\ref{app:datasets}. Prompts are written at three levels: \emph{open} (free-form attribute query), \emph{specific} (single-value yes/no query), and \emph{irrelevant}
(task-unrelated factual question), and combined into six pair groups
(Table~\ref{tab:prompt_groups}). G3 and G4 are restricted to the
secondary attribute axis; all pairs are evaluated in the forward
direction only. See details in Appendix~\ref{app:datasets}.

\begin{table}[!t]                                                                             
  \centering                                                    
  \caption{Prompt-pair groups. Examples show the prompt pairs for EmotionalStory dataset.}                                                                                 
  \label{tab:prompt_groups}
  \small  
  \begin{tabular}{@{}lllp{0.46\linewidth}@{}}                   
  \toprule  
  Group & Relation & Pair type & Example (Prompt $A \to$ Prompt $B$, EmotionalStory) \\
  \midrule  
  G1 & Cross-attribute   & open $\to$ open           & ``What topic is this text about?'' $\to$ ``What emotion does this text express?'' \\
  G2 & Cross-attribute   & specific $\to$ specific   & ``Is this about a career?'' $\to$ ``Does this text express joy?'' \\                                                                       
  G3 & Within-attribute  & open $\to$ specific       & ``What emotion does this text express?'' $\to$ ``Does this text express joy?'' \\                                                          
  G4 & Within-attribute  & specific $\to$ specific   & ``Does this text express joy?'' $\to$ ``Does this text express sadness?'' \\                                                               
  G5 & Irrelevant source & irrelevant $\to$ open     & ``What is the capital of France?'' $\to$ ``What emotion does this text express?'' \\                                                       
  G6 & Irrelevant source & irrelevant $\to$ specific & ``What is the capital of France?'' $\to$ ``Does this text express joy?'' \\                                                                
  \bottomrule     
  \end{tabular}                                                                                              
  \end{table}       

\paragraph{Implementation.}
We extract hidden states at the last input-prompt token from all
transformer layers, and fit and evaluate every transform under
five-fold cross-validation. For causal interventions we inject the fitted transform at a single layer and decode greedily up to $50$ tokens. Hyperparameters, layer indexing per model family, MLP architecture, and evaluator dictionaries are in Appendix.~\ref{app:impl}.

\begin{figure}[!t]
  \centering
  \includegraphics[width=0.92\linewidth]{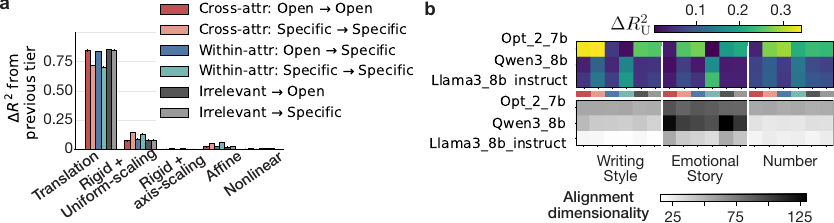}
  \caption{
\textbf{Comparison of prompt-pair groups and datasets.}
\textbf{(a)} Cross-validated $\Delta R^{2}$, averaged over datasets, models, and layers for each prompt-pair group. Specific prompt-pair groups show distinct decomposition profiles, regardless of whether the paired prompts differ across attributes or within an attribute.
\textbf{(b)}
\textbf{Top:} $\Delta R^{2}_{\mathrm{O_u}}$, the additional variance explained by rotation/reflection and uniform scaling beyond translation.
\textbf{Bottom:} alignment dimensionality $D$. Larger $D$ indicates that more dimensions contribute appreciably to the prompt-pair alignment. OPT-2.7B shows larger and more dataset-dependent $\Delta R^{2}_{\mathrm{O_u}}$, whereas Llama3-8B shows the lowest alignment dimensionality. EmotionalStory (story topic $\to$ emotion) shows higher alignment dimensionality across models.
}
  \label{fig:strategy_rank}
\end{figure}

\section{Results}
\label{sec:results}


\paragraph{Prompting reshapes representational geometry toward the instructed task structure.}
\label{sec:results_norm_change}%
\label{sec:results_rep_structure}%
We first ask how much change in the hidden activations is induced by prompting. We feed the same stimulus set under two cross-attribute open-ended prompts (G1) and measure the normalised squared Frobenius distance between the prompted hidden states (Fig.~\ref{fig:setup}\textbf{b}). The change grows nearly monotonically with depth, which is consistent across other prompt-pair groups (Appendix~\ref{app:norm_change}).

We then analyzed whether this prompt-induced change carries the geometrical structure required by prompt B (Fig.~\ref{fig:rep_structure}). The example 2D MDS visualization of the prompted hidden states of LLaVA-OneVision-7B for 1000 COCO images demonstrates that the representations of images were separated along the attribute the prompt targets (Fig.~\ref{fig:rep_structure}\textbf{a}; see Appendix~\ref{app:mds} for other models and datasets). The
representational dissimilariy matrix (RDM)s that computes pair-wise distances between image representations make this explicit (Fig.~\ref{fig:rep_structure}\textbf{b}): when stimuli are sorted by count, prompt $B$'s RDM displays the count-graded ordinal block
structure of the target count-RDM (bottom panel), while prompt $A$'s
does not. The layer-wise alignment with each prompt's target attribute grows monotonically
with depth and clearly separates prompt $B$ from prompt $A$ in the
top layers (Fig.~\ref{fig:rep_structure}\textbf{c}; pooled across models and datasets, see Appendix~\ref{app:rsa} for individual results). Higher RDM correlations indicate higher alignment with the task structure of prompt B and higher silhouette scores indicate better separation among categories of the target attribute. 

\begin{figure}[!t]
  \centering
  \includegraphics[width=\linewidth]{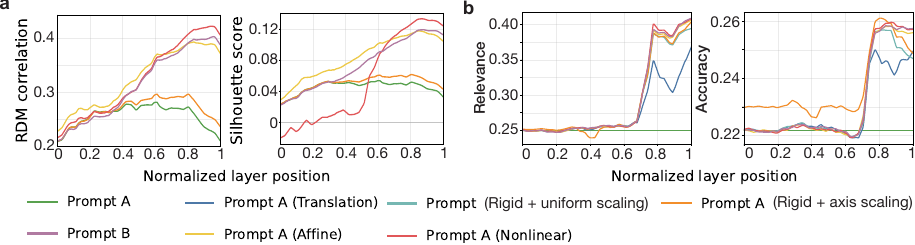}
  \caption{Causal validation of the nested geometric decomposition. At each layer, we replace prompt-$A$'s hidden states with the transformed counterpart for one fitted transformation type and then run the model forward through the remaining layers. \textbf{(a)} RDM correlation (left) and silhouette score (right). Axis-scaled rigid transformations improve representational alignment, affine maps produce a larger gain, and nonlinear maps yield the strongest recovery. \textbf{(b)}~Behavior evaluation on prompt-$B$'s task, measured by Relevance (left) and Accuracy (right). Beyond the translation tier, each successive tier improves behavioural recovery and progressively brings performance closer to prompt $B$.}
  \label{fig:intervention}
\end{figure}

\paragraph{Nested geometric decomposition.}
\label{sec:results_decomp}%
\label{sec:results_r2_decomposition}%
\label{sec:results_strategy_rank}%
To quantify the extent to which prompt-induced representational changes are captured at each level of transformation complexity, we fit the nested family of models
$\mathcal{F}_{\mathrm{T}}\subset\mathcal{F}_{\mathrm{O_u}}\subset
 \mathcal{F}_{\mathrm{O_{a}}}\subset\mathcal{F}_{\mathrm{L}}\subset
 \mathcal{F}_{\mathrm{N}}$
at every layer of each model for prompt pairs across six groups, and compute the \emph{incremental} $R^{2}$ contributed by each tier on held-out images (Fig.~\ref{fig:r2_decomposition}; Fig.~\ref{fig:strategy_rank}\textbf{a}; see Appendix \ref{app:r2bars} for group-specific results). Pure translation accounts for the largest share of explained variance, dominating in early layers and remaining substantial throughout the hierarchy. Higher-complexity tiers contribute additional, smaller increments primarily in mid-to-deep layers, with an overall increasing trend that peaks in the final layers. 

Across LLMs, compared with Llama3 and Qwen3, OPT-2.7B relies less on translation in early layers and more on rotation, reflection, scaling, and nonlinear transformations, as well as more on affine transformations in intermediate layers (Fig.~\ref{fig:r2_decomposition}). This pattern suggests that the models implement distinct early-layer encoding strategies: OPT-2.7B appears to represent prompts as more distributed perturbations that require rotational and higher-dimensional mixing components, whereas Llama3 and Qwen3 may rely on a lower-dimensional ``instruction code.'' Among VLMs, BLIP2 shows a stronger dependence on translation in late layers than LLaVA-OneVision and Qwen3-VL. In BLIP2, multimodal information may be encoded by the Q-Former as a steering vector that persists into deeper layers, whereas LLaVA-OneVision and Qwen3-VL show greater flexibility in information mixing, potentially because of differences in their training protocols.

We additionally compare the explained incremental variance of each tier across different
prompt-pair groups (Fig.~\ref{fig:strategy_rank}\textbf{a}). The five-tier incremental contribution hierarchy ($\mathcal{F}_{\mathrm{T}}>\mathcal{F}_{\mathrm{L}}>
 \mathcal{F}_{\mathrm{O_u}}>\mathcal{F}_{\mathrm{O_a}}\approx
 \mathcal{F}_{\mathrm{N}}$) is consistent across prompt-pair groups, indicating that it is not an artifact of any
particular pairing regime. Models rely more on rotation, reflection, and affine transformations for paired specific prompts, regardless of whether the pairs are cross-attribute or within-attribute. Appendix~\ref{app:para} further evaluates generalization across prompt paraphrases and out-of-distribution datasets, showing that fitted transformations are largely stable to semantic rewordings of the target prompt and partially transferable across stimulus distributions, although the relative contributions of transformations remain dataset-dependent.

\paragraph{Where rotation matters: explained variance and alignment dimensionality.}
$\Delta R^{2}_{\mathrm{O_u}}$ across datasets and prompt-pair groups (Fig.~\ref{fig:strategy_rank}\textbf{b}, top) shows that OPT-2.7B generally exhibits larger $\Delta R^{2}_{\mathrm{O_u}}$ than Llama3 and Qwen3. OPT also shows distinct profiles across datasets and groups, with much higher $\Delta R^{2}_{\mathrm{O_u}}$ for cross-attribute, open-ended prompt pairs in WritingStyle than in EmotionalStory or Number, which are more translation-dominated (see Appendix~\ref{app:r2bars}). By contrast, Llama3 and Qwen3 show higher $\Delta R^{2}_{\mathrm{O_u}}$ for within-attribute, specific prompt pairs than for other groups, whereas OPT shows the opposite trend, relying more on translation for within-attribute, specific prompt pairs.

We further computed the alignment dimensionality, defined as the number of singular values of the centered cross-covariance used by orthogonal Procrustes that exceed $0.01\,\sigma_{\max}$ (Fig.~\ref{fig:strategy_rank}\textbf{b}, bottom; Appendix~\ref{app:rank10} shows the same qualitative pattern under a $0.1$ threshold). Higher alignment dimensionality suggests that more dimensions participate in the Procrustes alignment between prompt pairs. EmotionalStory repeatedly shows higher dimensionality than WritingStyle or Number, suggesting that topic-versus-emotion prompt pairs induce a richer, more distributed alignment structure, even when the incremental explained variance is small. Llama3-8B-Instruct tends to have the smallest alignment dimensionalities across models, indicating that its alignment is concentrated in fewer dominant directions than that of OPT-2.7B or Qwen3-8B. The dimensionality of rotation and reflection therefore varies with both task structure and model family.

\paragraph{Causal validation: representational geometry and behavior.}
\label{sec:results_intervention}
We performed an interventional analysis (\S\ref{method:intervention}) and computed alignment metrics on held-out stimuli to test whether the fitted transformations can causally recover the representational geometry and behaviour induced by prompt $B$ (Fig.~\ref{fig:intervention}; pooled across models, datasets, and prompt-pair group G1). For the interventional analysis, we focus on the open-ended prompt $B$ (G1 and G5) because they preserve non-degenerate multi-label structure for RSA, avoid yes-bias artefacts in behavioural evaluation, and provide a matched comparison between task-relevant and task-irrelevant source prompts. 

Translation and uniform-scaled rigid transformations are omitted from Fig.~\ref{fig:intervention}\textbf{a} because they leave distance-based RDM rankings and silhouette scores unchanged (App.~\ref{app:rsa} for additional results). Axis-scaled rigid transformations improve geometric alignment but remain clearly below affine and nonlinear transformations, especially in later layers. This indicates that allowing each Procrustes-aligned axis to be rescaled independently captures part of the prompt-pair transformation, but does not fully account for the representational reorganisation. The additional improvement from affine maps suggests that feature mixing is important for matching prompt-$B$ geometry. The nonlinear map gives the strongest recovery, but its advantage over the affine map is smaller than the gain from the axis-scaled tier to the affine tier. Downstream behavior shows a sharper divide between translation and the higher-capacity transformations (Fig.~\ref{fig:intervention}\textbf{b}; see Appendix~\ref{app:behavior} for text outputs). As translation tier can substantially steer behavior, each successive tier improves behavioural performance and progressively brings it closer to prompt $B$. 

\section{Conclusion}
\label{sec:conclusion}
Our paper recast prompting as a low-complexity geometric transformation: a translation that does most of the work, rotation and shearing fixes the residual structured component. The effect of the prompts studied here, thus, can be understood as effecting an affine transform of the representation of the content following the prompt. Allowing a nonlinear transform did not explain either the resulting geometry or the resulting behavior substantially better. We see this as evidence that the hidden-state effects of instruction prompting are well approximated by an affine map at the population level, and that the residual fine structure --- the rotation, and the small nonlinear part --- is where the operational consequences of prompting are disproportionately concentrated. 

\textbf{Limitations and broader impacts.}
Our analysis is limited to zero-shot instruction prompts and does not
evaluate few-shot demonstrations, soft prompts, or longer-context
conditioning. Our intervention replaces the residual stream
at a single layer and final input-prompt token; multi-layer or multi-token
interventions and more systematic tests under input-distribution shift
remain future work. The proposed framework provides an interpretability
benefit by decomposing prompt-induced representational change into
explicit transformation classes and quantifying their contributions to variance, geometry, and behavior. Such measurements could also inform more efficient activation-level steering. We therefore view the method
both as a diagnostic tool for transparency and controlled evaluation, and
as a potential basis for deployment-oriented steering techniques.

\begin{ack}
\end{ack}

\bibliographystyle{plainnat}
\bibliography{references}


\clearpage
\appendix

\section{Methods: Nested Geometric Decomposition}
\label{app:methods}

\paragraph{Nested-chain inclusions.}
$\mathcal{F}_{\mathrm{T}}$ is recovered from $\mathcal{F}_{\mathrm{O_u}}$
at $(\mathbf{a},c,\mathbf{Q})=(\mathbf{0},1,\mathbf{I})$;
$\mathcal{F}_{\mathrm{O_u}}$ from $\mathcal{F}_{\mathrm{O_a}}$ at
$\mathbf{D}=c\,\mathbf{I}$; $\mathcal{F}_{\mathrm{O_a}}$ from
$\mathcal{F}_{\mathrm{L}}$ at $\mathbf{M}=\mathbf{Q}\,\mathbf{D}$ with
$\mathbf{Q}\in O(D)$ and $\mathbf{D}$ diagonal; and
$\mathcal{F}_{\mathrm{L}}$ from $\mathcal{F}_{\mathrm{N}}$ when
$g_{\boldsymbol{\theta}}$ implements a linear map.

\paragraph{Parameter estimation.}
The four parametric tiers all operate on the centred matrices                                 $\widetilde{\mathbf{X}}^{A}=\mathbf{X}^{A}-\boldsymbol{\mu}^{A}$ and                          $\widetilde{\mathbf{X}}^{B}=\mathbf{X}^{B}-\boldsymbol{\mu}^{B}$.                             smallskip                                                                                     \noindent\emph{Translation~($\mathcal{F}_{\mathrm{T}}$).}     
The mean-shift solution is                                                                    \begin{equation}                                              
    \widehat{\mathbf{b}} \;=\; \boldsymbol{\mu}^{B}-\boldsymbol{\mu}^{A}.                 \end{equation}                                                                             
  \smallskip                                                                                         
  \noindent\emph{Uniform-scaled rigid~($\mathcal{F}_{\mathrm{O_u}}$).}
  Compute the cross-covariance                                                                      
  $\mathbf{C}=\widetilde{\mathbf{X}}^{A\,\top}\widetilde{\mathbf{X}}^{B}$
  and its SVD                                                                                      
  $\mathbf{C}=\mathbf{U}\boldsymbol{\Sigma}\mathbf{V}^{\!\top}$.  The
  orthogonal-Procrustes solution is                                                                   
  \begin{equation}                                              
    \widehat{\mathbf{Q}}\;=\;\mathbf{U}\mathbf{V}^{\!\top},                                   \qquad  
    \widehat{c}\;=\; \frac{\bigl\langle\widetilde{\mathbf{X}}^{A}\widehat{\mathbf{Q}},\,\widetilde{\mathbf{X}}^{B}\bigr\rangle_{F}}    
         {\bigl\|\widetilde{\mathbf{X}}^{A}\bigr\|_{F}^{2}}.                                          
  \end{equation}    
  \smallskip                                                                                  
  \noindent\emph{Axis-scaled rigid~($\mathcal{F}_{\mathrm{O_a}}$).}
  Re-use $\widehat{\mathbf{Q}}$ from the $\mathcal{F}_{\mathrm{O_u}}$ fit, form the rigidly-aligned features $\mathbf{Z}=\widetilde{\mathbf{X}}^{A}\widehat{\mathbf{Q}}$, and fit each                                                                                           diagonal entry by per-axis least squares,                                                          
  \begin{equation}                                                                                  
    \widehat{d}_{j}\;=\;    
    \frac{\bigl\langle\mathbf{Z}_{:,j},\,\widetilde{\mathbf{X}}^{B}_{:,j}\bigr\rangle} 
         {\bigl\|\mathbf{Z}_{:,j}\bigr\|_{2}^{2}},    
    \qquad j=1,\dots,D. 
  \end{equation}     
  \smallskip                                                    
  \noindent\emph{Affine~($\mathcal{F}_{\mathrm{L}}$).}
  Ridge regression gives the closed form   
  \begin{equation}
    \widehat{\mathbf{M}}\;=\;       \bigl(\widetilde{\mathbf{X}}^{A\,\top}\widetilde{\mathbf{X}}^{A}+\lambda\mathbf{I}\bigr)^{-1}        
    \,\widetilde{\mathbf{X}}^{A\,\top}\widetilde{\mathbf{X}}^{B},                                                                                            
    \qquad \lambda=1.         
  \end{equation} 
  
  $g_{\boldsymbol{\theta}}$ is a two-layer MLP with one hidden layer of
  size $H{=}512$ and GELU activation. It is trained for 200 epochs over  
  the training-fold rows with mini-batch size $\min(256, N_{\text{train}})$,
  using AdamW (learning rate $10^{-3}$, weight decay $10^{-3}$) on the   
  mean-squared reconstruction error (equivalent to the Frobenius
  criterion up to a $1/(N_{\text{train}}\,D)$ normalization). The seed  
  is set per fold for reproducibility. 

For $k\in\{\mathrm{O_u},\mathrm{O_a},\mathrm{L}\}$ the empirical risk
is invariant under
$(\mathbf{a},\mathbf{b})\mapsto(\mathbf{a}+\mathbf{v},\mathbf{b}+f_{k}(\mathbf{v})-f_{k}(\mathbf{0}))$;
we adopt the canonical choice $\widehat{\mathbf{a}}=\boldsymbol{\mu}^{A}$
and $\widehat{\mathbf{b}}=\boldsymbol{\mu}^{B}$, reducing the remaining
estimation to a problem on the centred matrices
$\widetilde{\mathbf{X}}^{A},\widetilde{\mathbf{X}}^{B}$. For
$\mathcal{F}_{\mathrm{N}}$ we use the same centring convention, although
the risk is not invariant under arbitrary translations due to the
nonlinearity of $g_{\boldsymbol{\theta}}$.

\paragraph{Cross-validation.}
We use stratified five-fold cross-validation across each stimuli set; per-fold
estimators are fit on the training partition and evaluated on the
held-out partition. Cumulative $R^{2}_k$ and incremental
$\Delta R^{2}_k$ are computed from held-out residual sums of squares.

\paragraph{Hypothesis interpretation.}
Each pure transformation class produces a characteristic decomposition
profile, summarized in Table~\ref{tab:hypothesis_interp}.

\begin{table}[!h]
\centering
\caption{Decomposition signatures of pure transformation classes.}
\label{tab:hypothesis_interp}
\small
\begin{tabular}{@{}lll@{}}
\toprule
Pure transformation & Dominant component & Interpretation \\
\midrule
Mean shift $\mathbf{b}$
   & $\Delta R^{2}_{\mathrm{T}}\to 1$
   & concept-vector / steering hypothesis \\
Rigid + uniform scaling
   & $\Delta R^{2}_{\mathrm{O_u}}\to 1$
   & readout realignment / shape-preserving \\
Rigid + axis-wise scaling
   & $\Delta R^{2}_{\mathrm{O_a}}\to 1$
   & per-axis gain modulation \\
General linear map
   & $\Delta R^{2}_{\mathrm{L}}\to 1$
   & arbitrary linear feature mixing \\
Nonlinear map
   & $\Delta R^{2}_{\mathrm{N}}>0$
   & genuine nonlinear restructuring \\
Residual / noise
   & $R^{2}_{\mathrm{resid}}>0$
   & stimuli-dependent / beyond-MLP capacity \\
\bottomrule
\end{tabular}
\end{table}


\clearpage
\section{Datasets and prompt design}
\label{app:datasets}

\paragraph{Dataset design.}
\textbf{EmotionalStory} ($N{=}1{,}920$) is a corpus of
$8\text{ emotions}\!\times\!8\text{ topics}\!\times\!30$ short stories,
with name-pool and scenario-pool diversity injections inspired by \citep{sofroniew2026emotion}.
\textbf{WritingStyle} ($N{=}1{,}440$) is a synthetically generated
corpus of $6\text{ styles}\!\times\!4\text{ topics}\!\times\!60$ short
passages; style and content vary factorially by rendering the same
neutral content seed in each of the six styles.
\textbf{Number} ($N{=}1{,}247$) uses a mix of pseudo-stimuli, real
Wikipedia sentences, and task-template stimuli spanning five
cognitive-task framings, with set-property prompts (e.g.\ ``Are there
any prime numbers?'') that require scanning all numerical tokens
rather than matching a single value. Tables \ref{tab:example_emostory}--\ref{tab:example_number} show
representative stimuli from the three text datasets.
\textbf{EmoSet} ($N{=}1{,}600$) is a balanced
$8\text{ emotions}\!\times\!4\text{ content categories}$ (person,
animal, nature, object) subset of the EmoSet benchmark
\citep{yang2023emoset}.
\textbf{StyleTransfer} ($N{=}1{,}920$;
\citealp{boger2025style}) consists of photographs rendered in seven
styles (six artistic styles plus the original photographs) across four
scenes (beach, bedroom, library, mountain), with multiple images per
style$\times$scene cell.
\textbf{COCO} ($N{=}1{,}000$) is a subset of COCO val2017
\citep{lin2014coco} selected for high multi-supercategory coverage
($\geq 3$ of 12 supercategories per image), probing people-detection
and people-count prompts.

\paragraph{Diversity controls in synthetic LLM stimuli.}
Without explicit controls, the LLM seed generator collapses on a few
canonical patterns (e.g.\ defaulting to ``Maya'' for joy stories and
``Marcus'' for anger stories, or repeating one factual seed across all
\emph{nature} passages). To prevent this we (i) curate a balanced
$100$-name pool for EmotionalStory and assign names by a coprime stride
so each emotion uses $\sim 95$ distinct names with $\le 4$ occurrences
per (name, emotion) cell; (ii) curate a $15$-scenario pool per topic
($120$ total) and rotate as $\text{scenarios}[k\bmod 15]$, capping
each scenario at $\le 2$ occurrences per cell; (iii) for WritingStyle,
curate $15$ subthemes per topic and rotate round-robin so each
subtheme contributes $4$ stimuli at full scale.

\paragraph{Number-stimulus design.}
The Number dataset extends the design in \citet{hu2026} in two ways. First,
prompts are \emph{set-property} queries (``Are there any prime
numbers?'', ``Are there any numbers greater than 5?'') rather than
value-identification queries, because lexical surface-form matching
trivially solves ``Is the number exactly 3?'' on the input
``I have 3 apples'' without engaging numerical processing. Second, the
single-number random insertion is generalized to a balanced
1-/2-/3-number multi-number variant. Stimuli combine three types:
(a)~pseudo-stimuli ($7$-word \texttt{wikitext-103}
chunks \citep{merity2016wikitext} with $c\!\in\!\{1,2,3\}$ target
numbers inserted at random positions; $N{=}90$); (b)~real Wikipedia
sentences containing one or more target numbers $1$--$9$
($N{\approx}675$); and (c)~task-template stimuli reproducing five
cognitive tasks:quantity, comparison, arithmetic, property,
ordinal, with five phrasings per (task, number, format) cell
($N{\approx}442$). Anchor values in comparison/arithmetic templates are
mathematically validated against the target before acceptance.

\begin{table}[!h]
\centering
\caption{Example stimuli from \textbf{EmotionalStory}
( topic $\times$ emotion, with one example per emotion
and per topic, repeated to balance both axes; first $\sim\!130$
characters of each story shown). The full corpus is $8\,\text{emotions}
\times 8\,\text{topics}\times 30 = 1{,}920$ stories.}
\label{tab:example_emostory}
\footnotesize
\begin{tabular}{@{}llp{0.21\linewidth}p{0.42\linewidth}@{}}
\toprule
Emotion & Topic & Scenario & First $\sim\!130$ chars \\
\midrule
joy          & career     & performance review meeting    & ``Aaliyah sat across from her manager, hands folded in her lap, bracing for the usual mix of praise and critique. Instead, her$\ldots$'' \\
sadness      & education  & a thesis defense              & ``Devon had spent five years on this research, and as the committee chair slid the marked-up pages across the table and quietly$\ldots$'' \\
anger        & family     & a kitchen scene               & ``Iris had asked her younger brother three times to stop leaving his dirty dishes in the sink, and yet there they were again --- a$\ldots$'' \\
fear         & friendship & a coffee shop meetup          & ``Nasir arrived at the coffee shop early, rehearsing what he needed to say to his best friend of ten years --- that he had$\ldots$'' \\
trust        & health     & a doctor's visit              & ``Saif sat across from Dr.\ Okafor, who had been his physician for over a decade, and listened as she carefully explained his new$\ldots$'' \\
disgust      & hobbies    & a painting session            & ``Wilhelm set his brush down and leaned in to examine the still-life arrangement he'd been painting for the past hour --- a bowl of$\ldots$'' \\
surprise     & finance    & buying a house                & ``Anika sat across from the loan officer, bracing herself for the familiar sting of rejection, when he slid a folder across the$\ldots$'' \\
anticipation & travel     & an airport delay              & ``Felipe checked the departures board for the fourth time in ten minutes, watching the blinking status next to his flight number$\ldots$'' \\
joy          & education  & a thesis defense              & ``Beatriz stood at the front of the conference room, her hands trembling slightly as the last committee member lowered his pen and$\ldots$'' \\
sadness      & family     & a kitchen scene               & ``Gerwin stood at the kitchen counter, turning his mother's old recipe card over in his hands, the handwriting so familiar it made$\ldots$'' \\
anger        & friendship & a coffee shop meetup          & ``Lila sat across from her best friend at their usual corner table, watching her scroll through her phone for the third time since$\ldots$'' \\
fear         & health     & a doctor's visit              & ``Ren sat on the crinkled paper of the examination table, hands gripping the edge until the knuckles went pale, waiting for the$\ldots$'' \\
trust        & hobbies    & a painting session            & ``Thandiwe set her half-finished canvas on the easel beside her mentor's, nervously watching as the older woman leaned in to study$\ldots$'' \\
disgust      & finance    & buying a house                & ``Adaeze flipped through the inspection report for the house she had been dreaming about for months, and her stomach turned when$\ldots$'' \\
surprise     & travel     & an airport delay              & ``Diego had settled into his gate seat with a coffee and a book, fully resigned to the four-hour delay the departures board had$\ldots$'' \\
anticipation & career     & performance review meeting    & ``Magnus straightened his tie for the third time as he sat outside his manager's closed office door, the muffled voices inside$\ldots$'' \\
\bottomrule
\end{tabular}
\end{table}

\begin{table}[!h]
\centering
\caption{Example stimuli from \textbf{WritingStyle}
(topic $\times$ style). The same neutral content seed is rendered in
each of the six styles within a topic, so within-topic rows differ
\emph{only} in style. Each style and topic appears at least once;
the full corpus is $6\,\text{styles}\times 4\,\text{topics}\times
60 = 1{,}440$ short passages.}
\label{tab:example_writingstyle}
\footnotesize
\begin{tabular}{@{}llp{0.65\linewidth}@{}}
\toprule
Style & Topic & First sentence \\
\midrule
formal       & nature     & ``Forests occupy approximately 31 percent of Earth's total land surface and constitute the primary habitat for an estimated 80 percent of the planet's terrestrial biodiversity.'' \\
casual       & nature     & ``Forests take up about 31\% of all the land on Earth, which is pretty impressive when you think about it.'' \\
poetic       & nature     & ``Like a living mantle draped across nearly a third of Earth's ancient skin, forests breathe and pulse with the world's wild abundance.'' \\
technical    & technology & ``A smartphone is a handheld electronic device that integrates mobile telephony with general-purpose computing capabilities, incorporating$\ldots$'' \\
journalistic & technology & ``Smartphones are handheld devices that integrate mobile phone capabilities with computing functions, including internet access and$\ldots$'' \\
archaic      & technology & ``Behold the smartphone, that marvellous handheld contrivance which doth unite within its slender form the offices of the common$\ldots$'' \\
formal       & food       & ``Macronutrients constitute a class of essential dietary compounds required by the human body in substantial quantities to support$\ldots$'' \\
casual       & food       & ``Macronutrients are basically the nutrients your body needs a lot of to keep you energized and help you grow.'' \\
poetic       & food       & ``Like pillars holding up the temple of the body, macronutrients are the great sustaining forces that fuel our every breath, our$\ldots$'' \\
technical    & sports     & ``Each match consists of two opposing teams, each fielding exactly eleven players on the pitch simultaneously.'' \\
journalistic & sports     & ``Two teams of eleven players each compete in soccer, a sport in which the objective is to advance the ball into the opposing team's goal.'' \\
archaic      & sports     & ``Verily, the noble sport of soccer is contested betwixt two sides, each comprising eleven players of valiant constitution.'' \\
\bottomrule
\end{tabular}
\end{table}

\begin{table}[!h]
\centering
\caption{Example stimuli from \textbf{Number}, spanning the three
stimulus types and the five task framings of \citet{hu2026}, each shown
in both \emph{digit} and \emph{word} numerical formats. The full corpus
has $N{\approx}1{,}247$ stimuli.}
\label{tab:example_number}
\footnotesize
\begin{tabular}{@{}lllp{0.55\linewidth}@{}}
\toprule
Stimulus type & Task framing & Format & Text \\
\midrule
pseudo-stimulus & ---        & digit & ``Sold at a lower price than 7 the.'' \\
pseudo-stimulus & ---        & word  & ``Museum remained eight in the tower building for.'' \\
real Wikipedia  & ---        & digit & ``An academic study found that Jordan's 1st NBA comeback resulted in an increase in the market$\ldots$'' \\
real Wikipedia  & ---        & word  & ``In 2014, Fernandez played the leading lady in Sajid Nadiadwala's \emph{Kick}, which is one of the$\ldots$'' \\
task-template   & quantity   & digit & ``I have a total of 1 apple.'' \\
task-template   & quantity   & word  & ``I have a total of one apple.'' \\
task-template   & comparison & digit & ``A number smaller than 15 is 1.'' \\
task-template   & comparison & word  & ``A number smaller than nine is one.'' \\
task-template   & arithmetic & digit & ``The sum of 0 and 1 is 1.'' \\
task-template   & arithmetic & word  & ``The sum of zero and one is one.'' \\
task-template   & property   & digit & ``A perfect square is 1.'' \\
task-template   & property   & word  & ``An example of an odd number is one.'' \\
task-template   & ordinal    & digit & ``1 comes after 0.'' \\
task-template   & ordinal    & word  & ``Before two comes one.'' \\
\bottomrule
\end{tabular}
\end{table}

\paragraph{Prompt templates.}
EmotionalStory uses \emph{``What emotion does this text express?''} and
\emph{``What topic is this text about?''} as the two open prompts; the
sixteen specific prompts are yes/no queries
\emph{``Does this text express \{value\}?''} (joy, sadness, anger, fear,
trust, disgust, surprise, anticipation) and \emph{``Is this about
\{value\}?''} (career, education, family, friendship, health, hobbies,
finance, travel). WritingStyle uses the analogous open pair
\emph{``What writing style is this in?''} / \emph{``What topic is this
text about?''} with the six styles ($\{$formal, casual, poetic,
technical, journalistic, archaic$\}$) and four topics ($\{$nature,
technology, food, sports$\}$). Number uses the nine set-property
prompts distributed across the five cognitive tasks (1 quantity / 3
comparison / 2 arithmetic / 2 property / 1 ordinal). Irrelevant prompts
(\emph{``What is the capital of France?''} and similar) are shared
across all datasets. \textbf{StyleTransfer} uses \emph{``What scene does the image depict?''} and \emph{``What artistic style does the image belong to?''} as the open pair, with eleven specific prompts: four scenes (\emph{``Is this a \{beach, bedroom, library, mountain\}?''}) and seven styles (\emph{``Is this in the style of \{Demuth, Klimt, Monet, Pollock\}?''}). 
  \emph{``Are there people in this image?''} (detection) and
  \emph{``How many people are in this image?''} (count) as the open
  pair, with eight count-specific yes/no queries of the form
  \emph{``Are there exactly \{0, 1, 2, 3, 4, 5\} people in this image?''}
  plus the two range bins \emph{``Are there between 5 and 10 people?''}
  and \emph{``Are there more than 10 people?''}.
  \textbf{EmoSet} (VLM) uses \emph{``What is depicted in this image?''}
  (content) and \emph{``What emotion does this image evoke?''} (emotion)
  as the open pair, with twelve specific prompts: four content categories
  (\emph{``Are there people / animals?''}, \emph{``Is this a natural
  landscape?''}, \emph{``Is this primarily an inanimate scene?''}) and
  eight emotions (\emph{``Does this image evoke \{amusement, awe, contentment,
  excitement, anger, disgust, fear, sadness\}?''}). Irrelevant prompts
  (\emph{``What is the capital of France?''}, \emph{``What is 2+2?''},
  \emph{``What day of the week is it?''}, and seven similar
  factual-knowledge questions) are shared across all six datasets. Full prompt JSONs will be in the released code.

\clearpage
\section{Implementation details}
\label{app:impl}

\paragraph{Layer indexing.}
For each model we extract hidden states at every transformer block at
the last input-prompt token. Layer indices are reported on a normalized
$[0,1]$ scale so that models of different depth can be overlaid;
absolute layer counts are: OPT-2.7B (32), Llama-3-8B-Instruct (32),
Qwen3-8B (36), BLIP-2 (32),
LLaVA-OneVision-7B (28), Qwen3-VL-8B (36).

\paragraph{Feature extraction.}
For each (model, dataset, prompt) cell we feed every stimulus through
the model under the prompt and store the residual-stream hidden state
at the \emph{last input-prompt token} from every transformer block;
this token is the position at which the model commits to a continuation
and is the conventional probe site for prompt-conditioned reads. LLM
inputs are formatted using the model's training-time convention: chat
models (Llama-3-Instruct, Qwen-2-Instruct, Qwen3) use
\texttt{tokenizer.apply\_chat\_template(\dots, add\_generation\_prompt=True)};
OPT-2.7B (a non-instruction-tuned base model) uses the QA template
(\verb|Question: {prompt}\nText: {text}\nAnswer:|). VLM inputs are
processed by each model's native processor (BLIP-2, LLaVA-OneVision,
Qwen3-VL). Hidden states are stored as zarr arrays of shape
$(1,\,n_{\text{stim}},\,D)$ per layer with metadata recording the
pooling convention, model identity, prompt string, sequence length, and
the number of truncated stimuli (typically zero).

\paragraph{Cross-validation, ridge, and MLP.}
Five-fold stratified cross-validation across stimuli, ridge
$\lambda{=}1$, and a one-hidden-layer MLP ($H{=}512$, GELU, AdamW with weight
decay $10^{-3}$, 200 epochs at mini-batch size $\le 256$), see Appendix~\ref{app:methods} for the estimator definitions.

\paragraph{Behavioural evaluation.}
Per-dataset keyword evaluators check whether generated text satisfies
the target prompt's semantic constraint, scoring two metrics: a
\emph{relevance} flag (does the text address the target attribute at
all, e.g.\ does it mention any artistic-style or emotion vocabulary)
and an \emph{accuracy} flag (does the text identify the correct
ground-truth label, e.g.\ ``Monet'' for a Monet-style image, ``joy''
for a joy-emotion story, the correct count for COCO). For style and
emotion prompts the evaluator combines artist-name keywords, art-school
keywords, and synonym lists (e.g.\ ``post-impressionism'',
``van gogh'', ``starry night'' all map to \texttt{vangogh}). Evaluator
dictionaries will be in the released code.

\clearpage
\section{Normalized prompt-induced activation changes}
\label{app:norm_change}

\begin{figure}[h]
  \centering
  \includegraphics[width=0.75\linewidth]{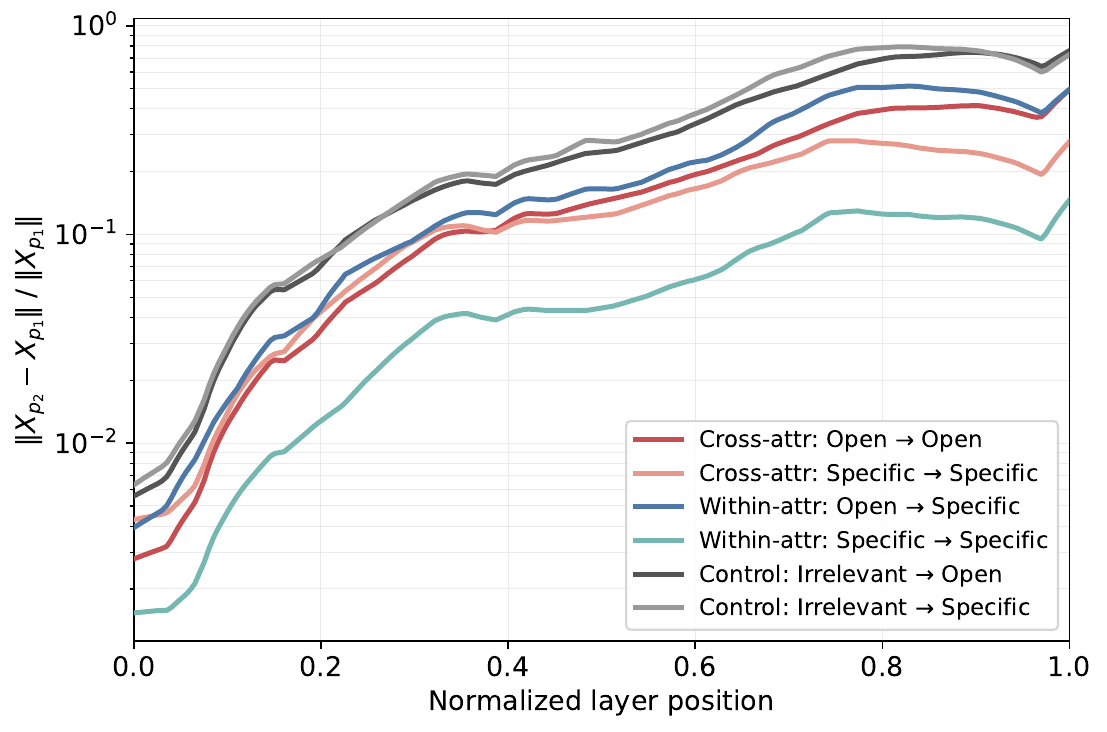}
  \caption{Normalized prompt-induced activation change across the six
  prompt-pair groups, pooled across all (model, dataset) cells. The
  depth-graded growth of the change is consistent across groups.}
  \label{fig:norm_change_groups}
\end{figure}

\begin{figure}[h]
  \centering
  \includegraphics[width=0.75\linewidth]{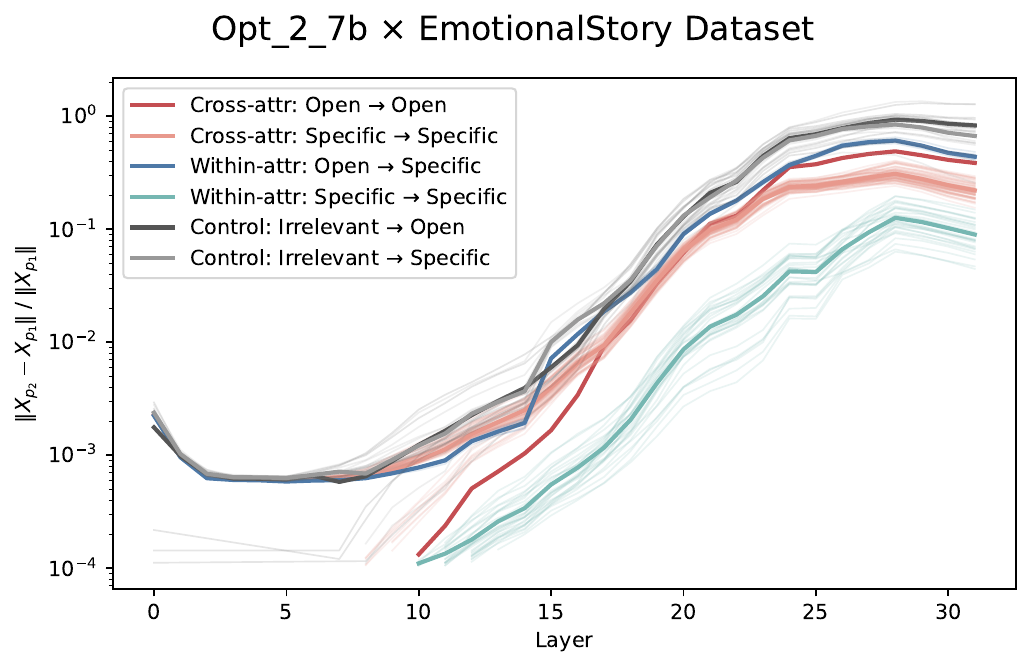}\\[2pt]
  \includegraphics[width=0.75\linewidth]{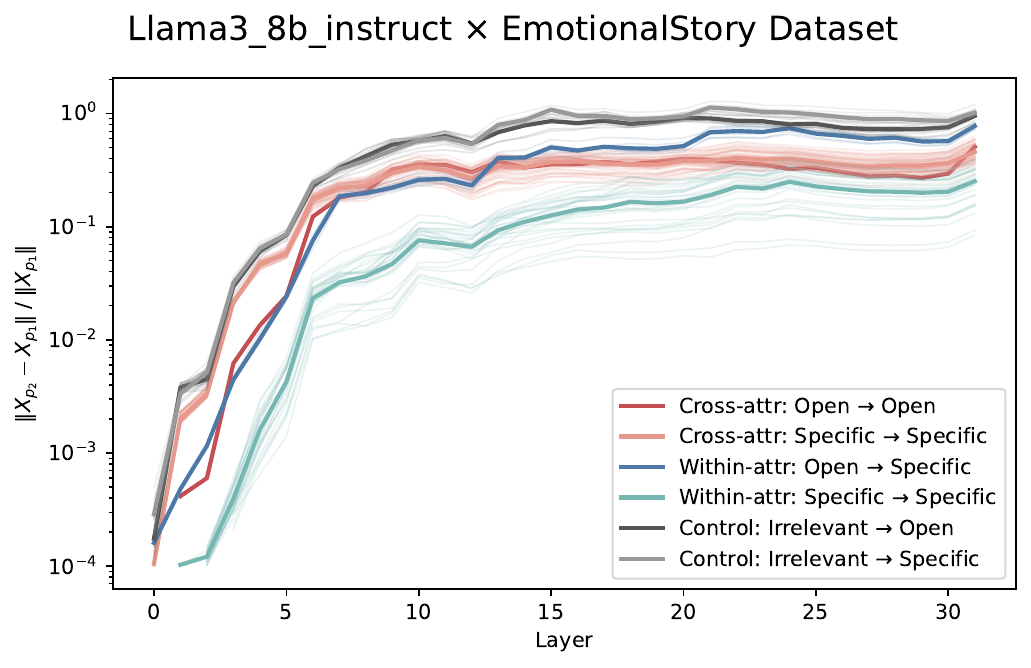}\\[2pt]
  \includegraphics[width=0.75\linewidth]{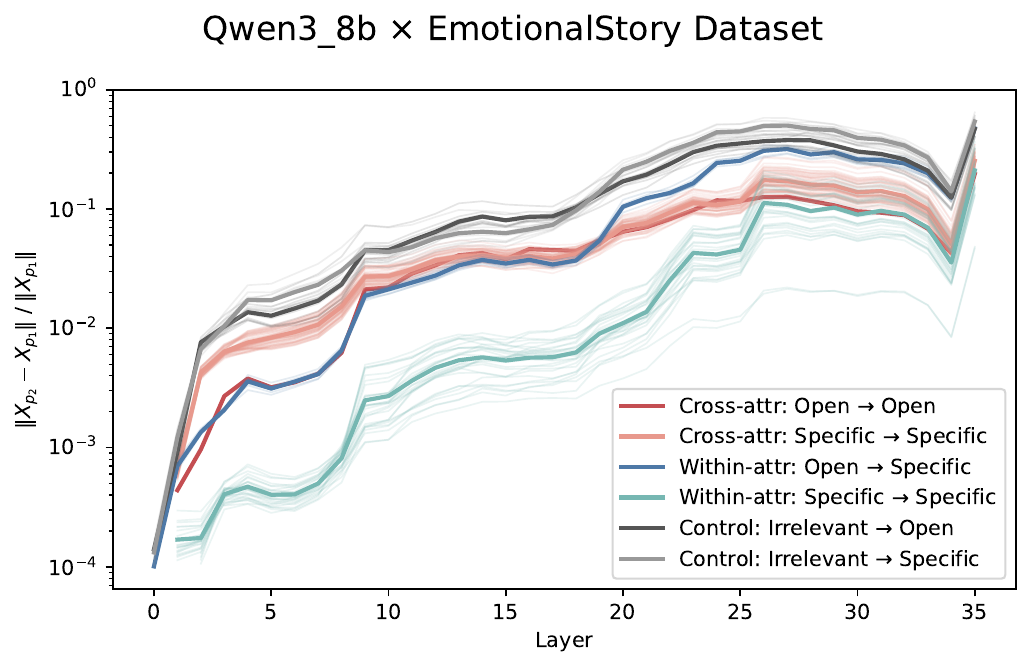}
  \caption{Normalized activation change for EmotionalStory across
  layers for OPT-2.7B (top),
  Llama-3-8B-Instruct (middle), Qwen3-8B (bottom).}
  \label{fig:norm_change_emostory}
\end{figure}

\begin{figure}[h]
  \centering
  \includegraphics[width=0.75\linewidth]{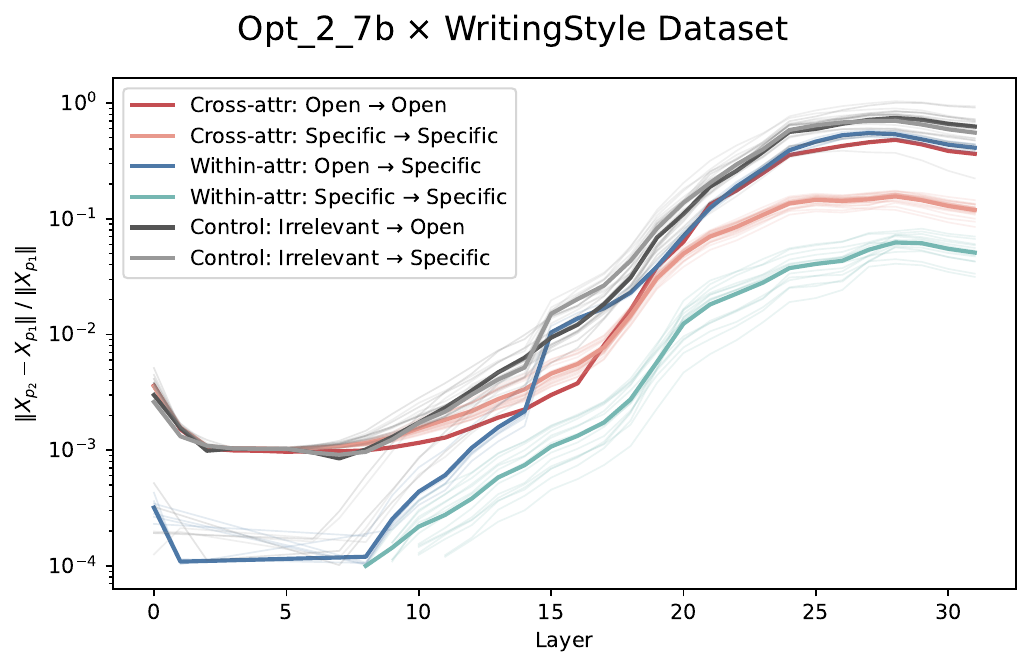}\\[2pt]
  \includegraphics[width=0.75\linewidth]{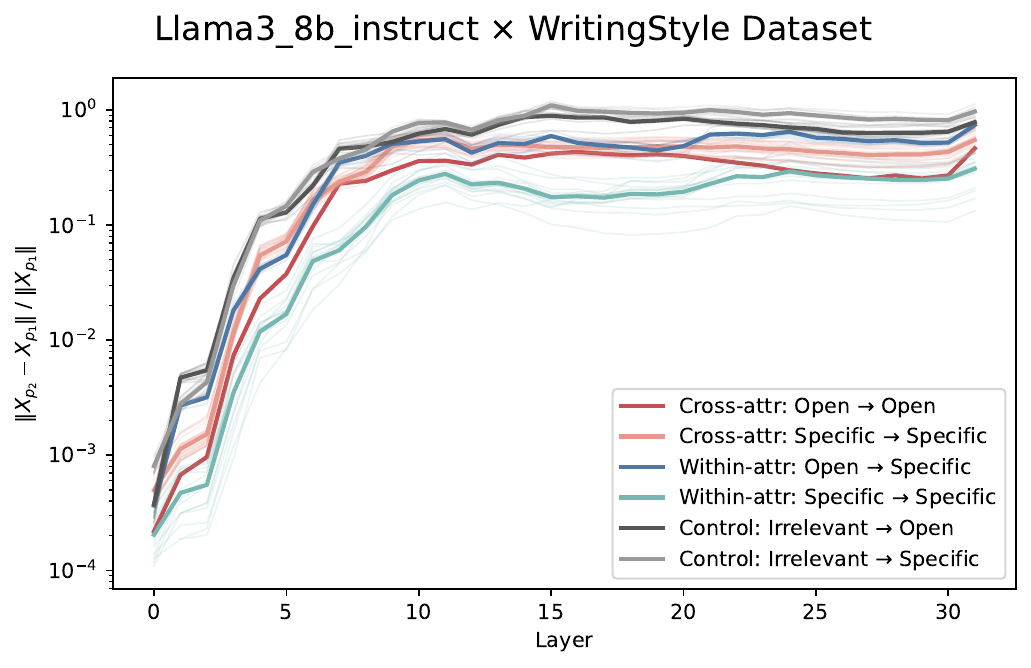}\\[2pt]
  \includegraphics[width=0.75\linewidth]{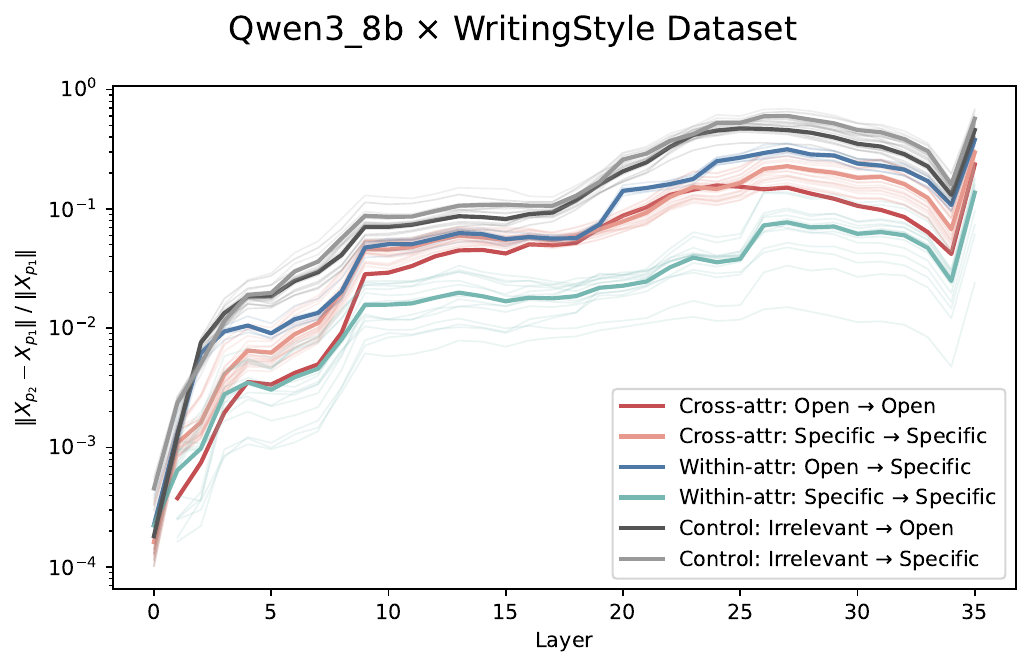}
  \caption{Normalized activation change for WritingStyle across
  layers for OPT-2.7B (top),
  Llama-3-8B-Instruct (middle), Qwen3-8B (bottom).}
  \label{fig:norm_change_writingstyle}
\end{figure}

\begin{figure}[h]
  \centering
  \includegraphics[width=0.75\linewidth]{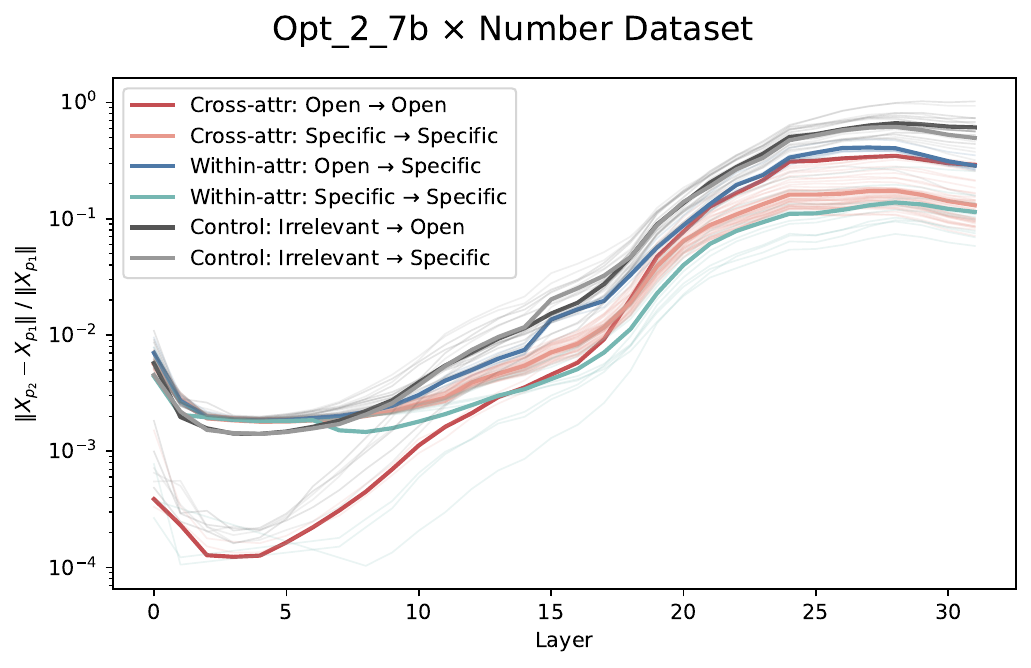}\\[2pt]
  \includegraphics[width=0.75\linewidth]{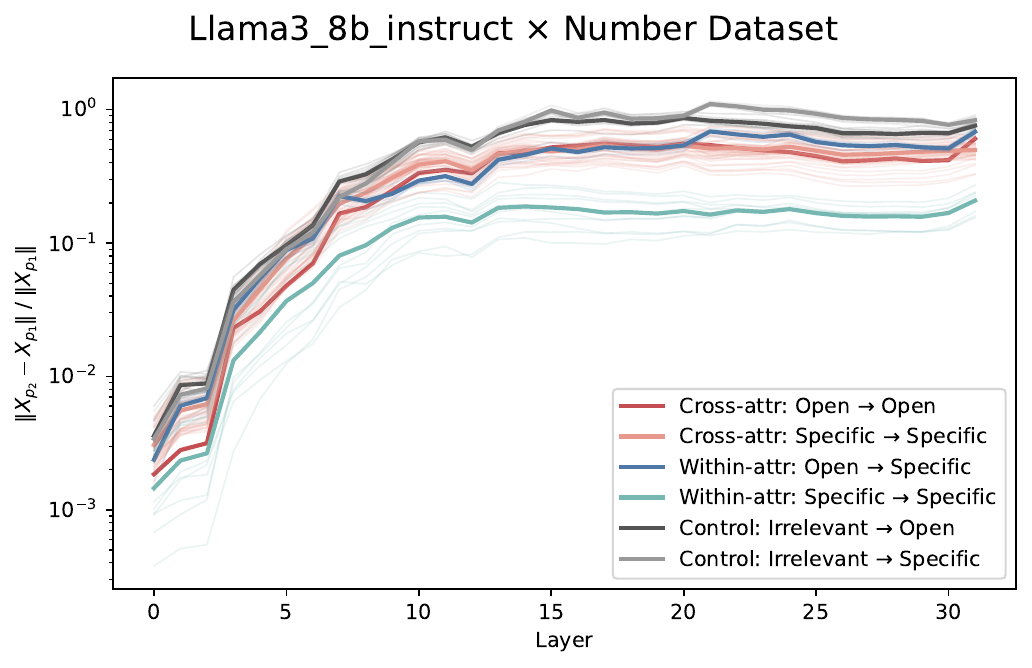}\\[2pt]
  \includegraphics[width=0.75\linewidth]{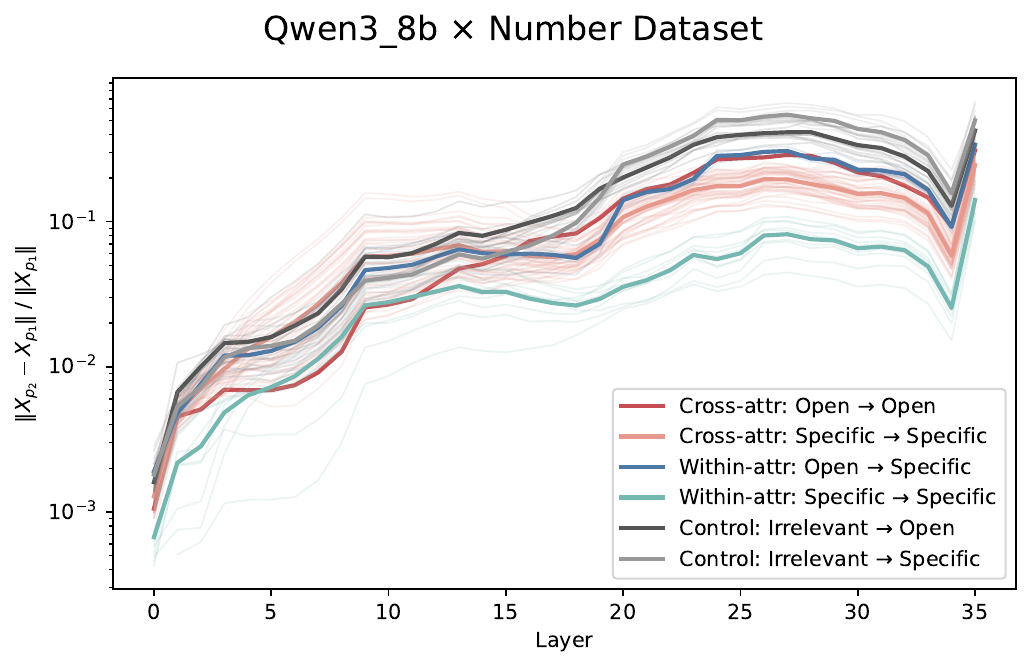}
  \caption{Normalized activation change for Number across layers for OPT-2.7B (top),
  Llama-3-8B-Instruct (middle), Qwen3-8B (bottom).}
  \label{fig:norm_change_number}
\end{figure}

\begin{figure}[h]
  \centering
  \includegraphics[width=0.75\linewidth]{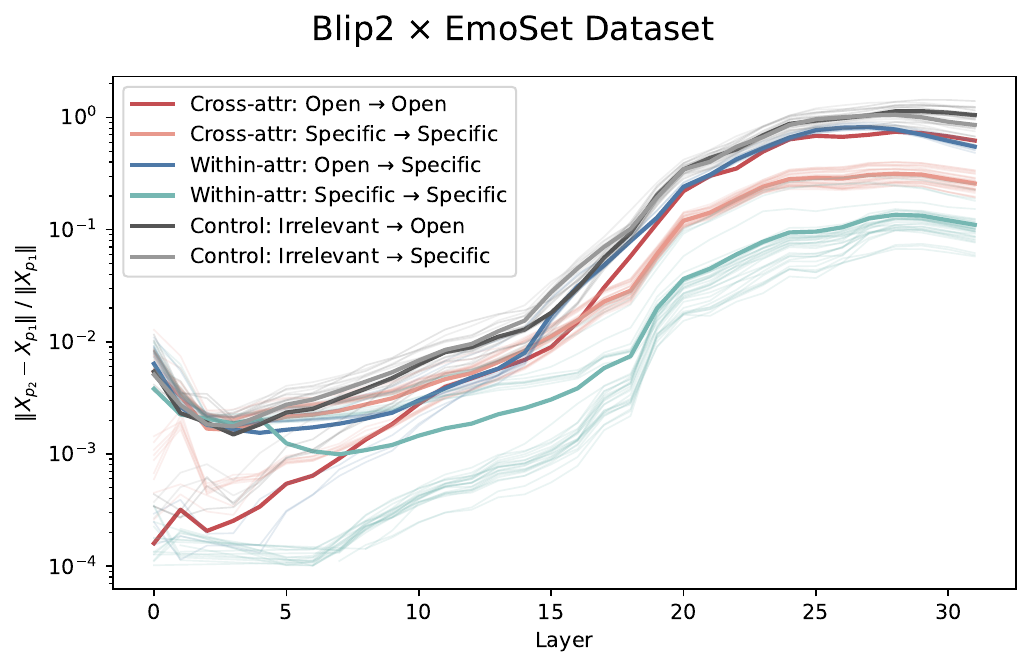}\\[2pt]
  \includegraphics[width=0.75\linewidth]{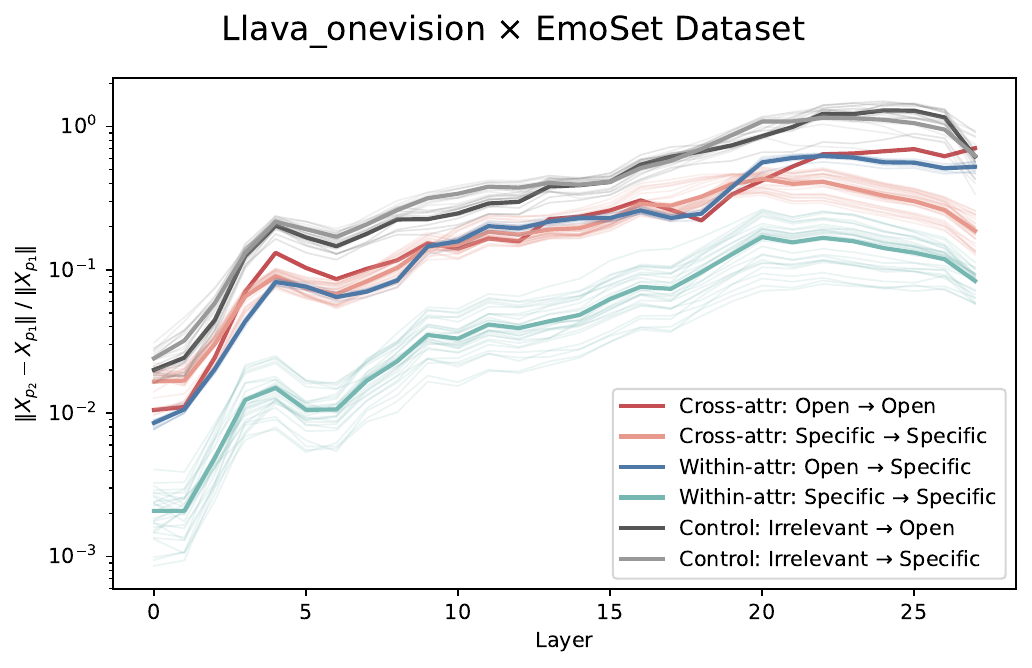}\\[2pt]
  \includegraphics[width=0.75\linewidth]{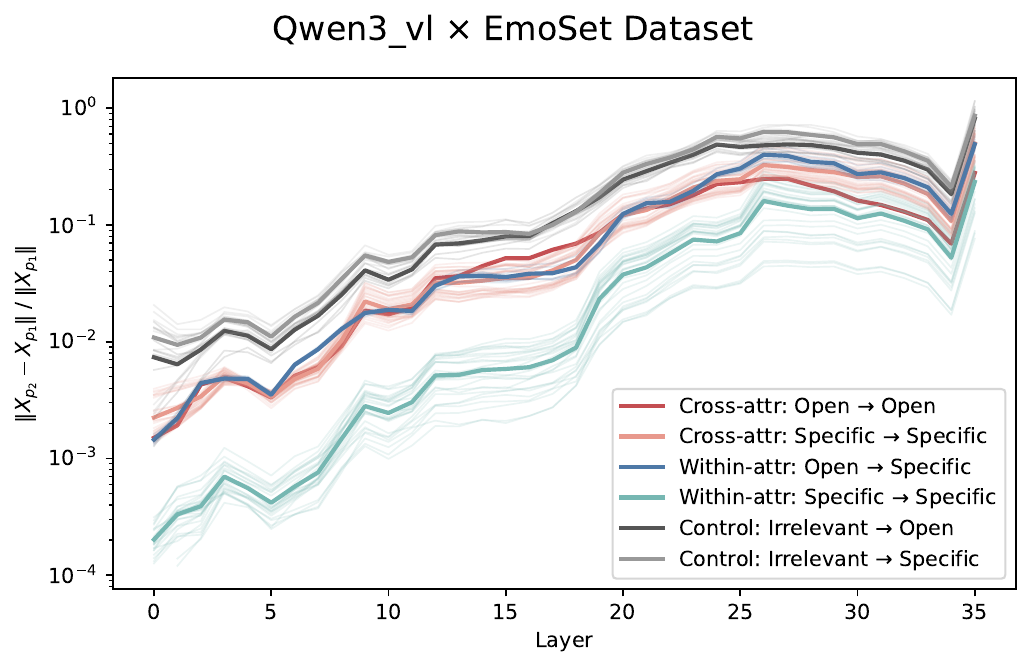}
  \caption{Normalized activation change for EmoSet across layers for BLIP-2 (top),
  LLaVA-OneVision-7B (middle), Qwen3-VL-8B (bottom).}
  \label{fig:norm_change_emoset}
\end{figure}

\begin{figure}[h]
  \centering
  \includegraphics[width=0.75\linewidth]{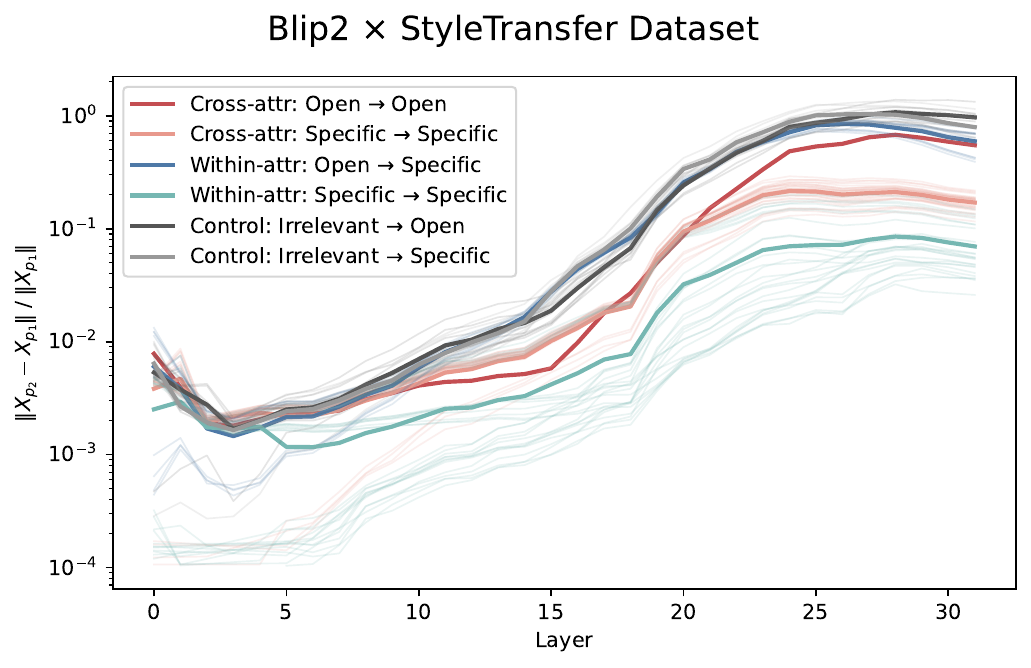}\\[2pt]
  \includegraphics[width=0.75\linewidth]{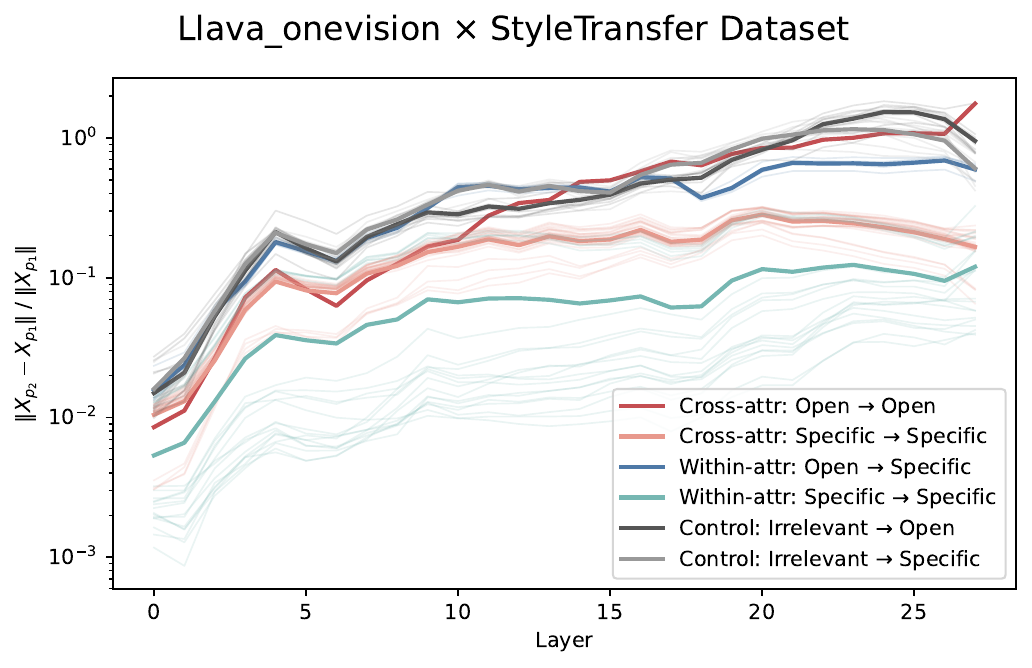}\\[2pt]
  \includegraphics[width=0.75\linewidth]{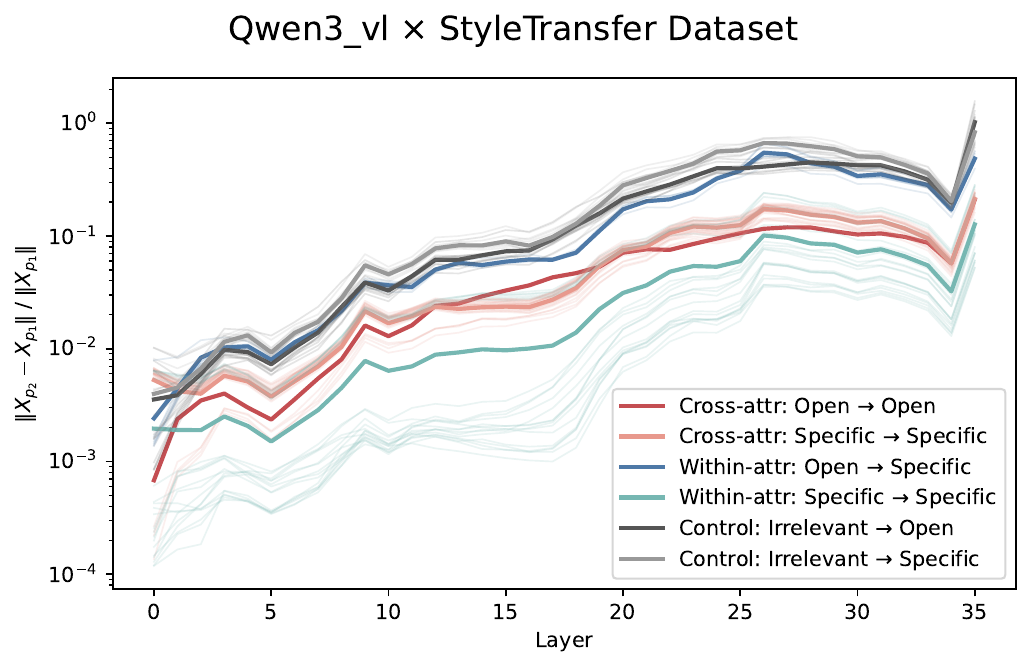}
  \caption{Normalised activation change for StyleTransfer across
  layers for BLIP-2 (top),
  LLaVA-OneVision-7B (middle), Qwen3-VL-8B (bottom).}
  \label{fig:norm_change_styletransfer}
\end{figure}

\begin{figure}[h]
  \centering
  \includegraphics[width=0.75\linewidth]{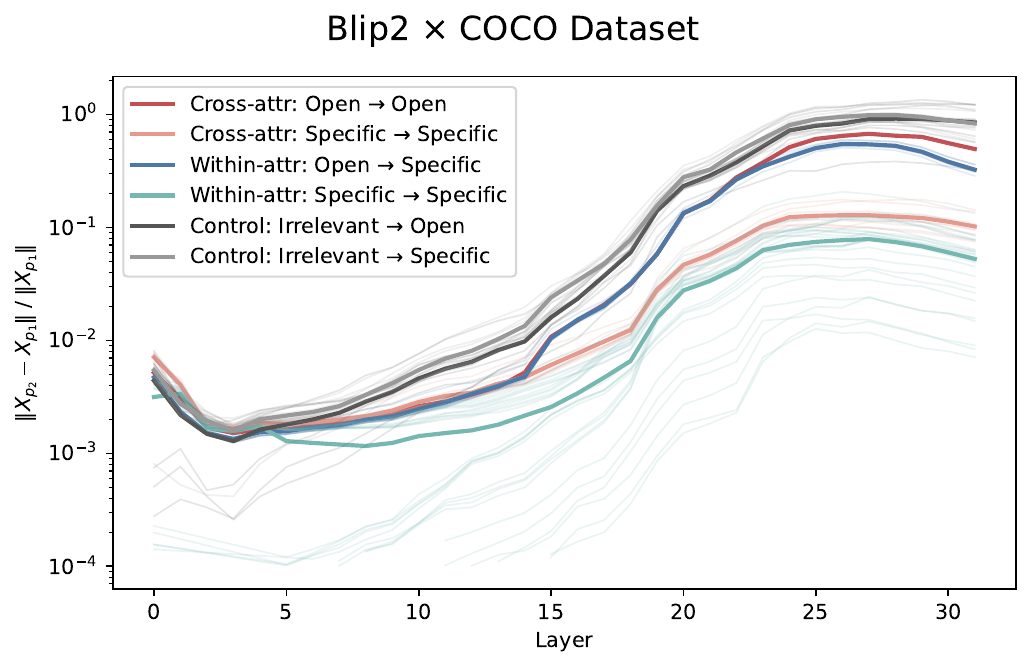}\\[2pt]
  \includegraphics[width=0.75\linewidth]{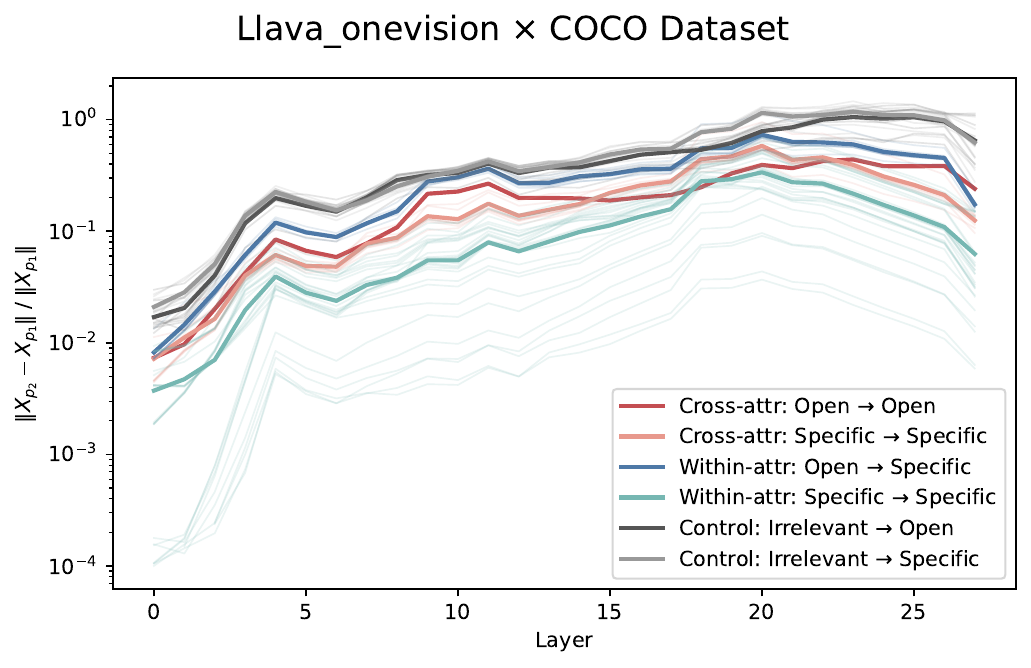}\\[2pt]
  \includegraphics[width=0.75\linewidth]{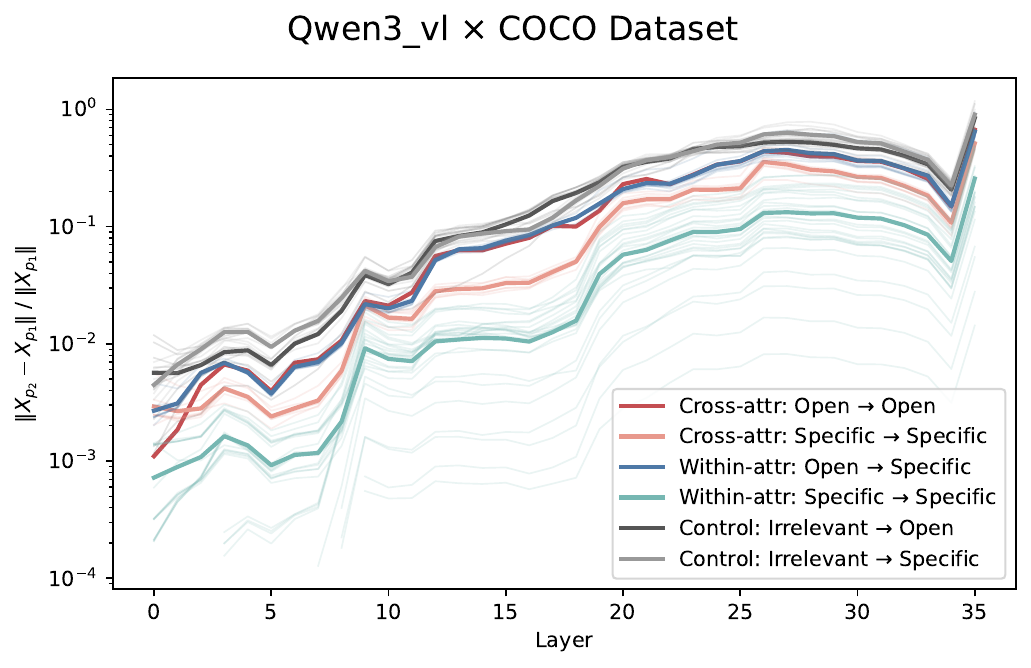}
  \caption{Normalized activation change for COCO across layers for BLIP-2 (top),
  LLaVA-OneVision-7B (middle), Qwen3-VL-8B (bottom).}
  \label{fig:norm_change_coco}
\end{figure}

The normalized squared Frobenius distance
$\|\mathbf{X}^{B}-\mathbf{X}^{A}\|_{F}^{2}/\|\mathbf{X}^{A}\|_{F}^{2}$
are shown separately for the six
prompt-pair groups (G1--G6). The summary across models and datasets is
shown in Fig.~\ref{fig:norm_change_groups}, and the individual (model, dataset) pair results are in Figs.~\ref{fig:norm_change_emostory}--\ref{fig:norm_change_coco}. The two irrelevant-source prompt groups (G5--G6) produce the largest activation change, followed by the within-attribute open$\to$specific pair (G3) and the two cross-attribute pairs (G1: open$\to$open; G2: specific$\to$specific). The within-attribute specific$\to$specific pair (G4) produces the smallest change, consistent with both prompts querying the same attribute axis at the same specificity level.

\clearpage
\section{MDS visualizations}
\label{app:mds}

\begin{figure}[h]
  \centering
  \rdsstretched{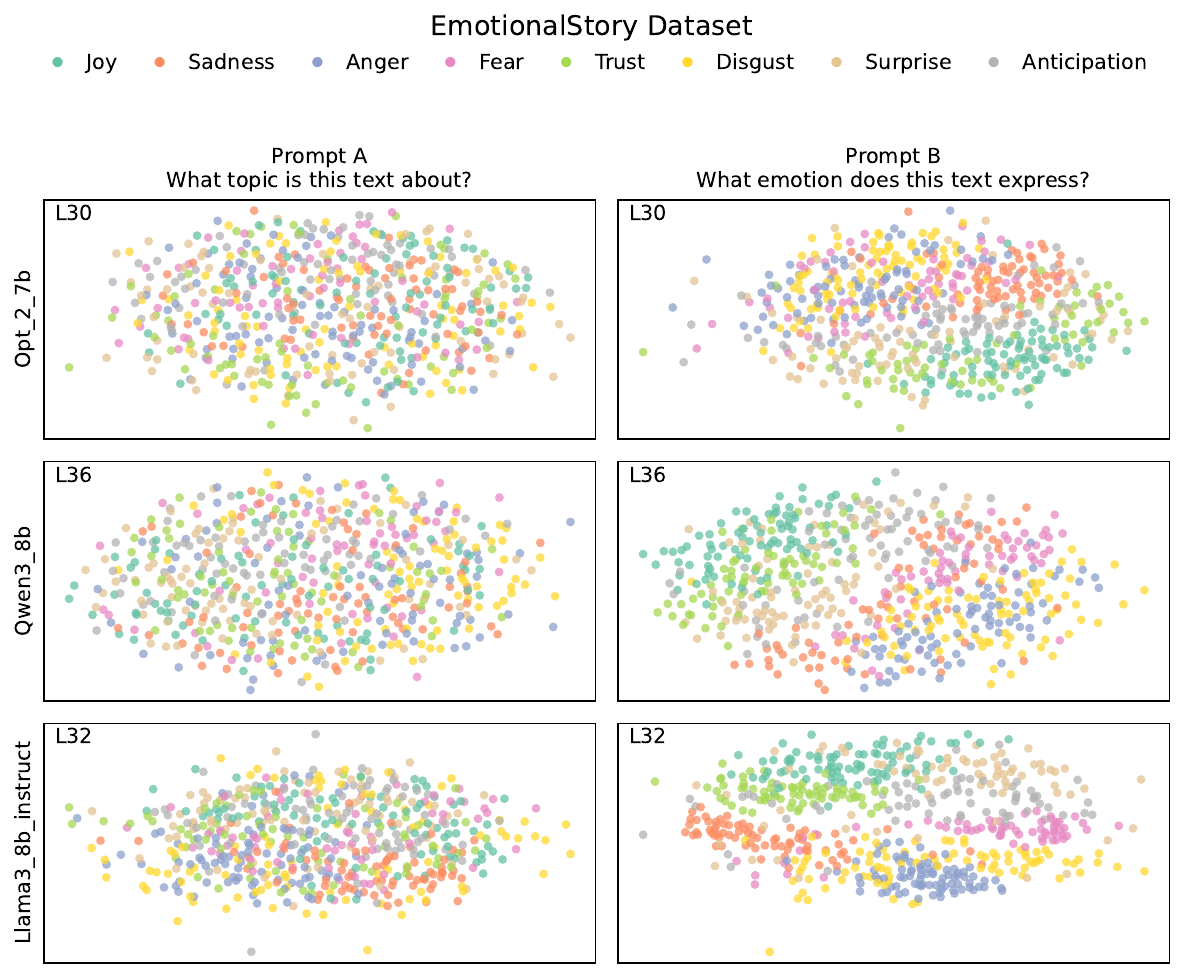}
  \caption{EmotionalStory dataset, prompt $A$ (topic) vs.\ prompt $B$
  (emotion); stimuli coloured by ground-truth emotion.}
  \label{fig:rds_emotionalstory}
\end{figure}

\begin{figure}[h]
  \centering
  \rdsstretched{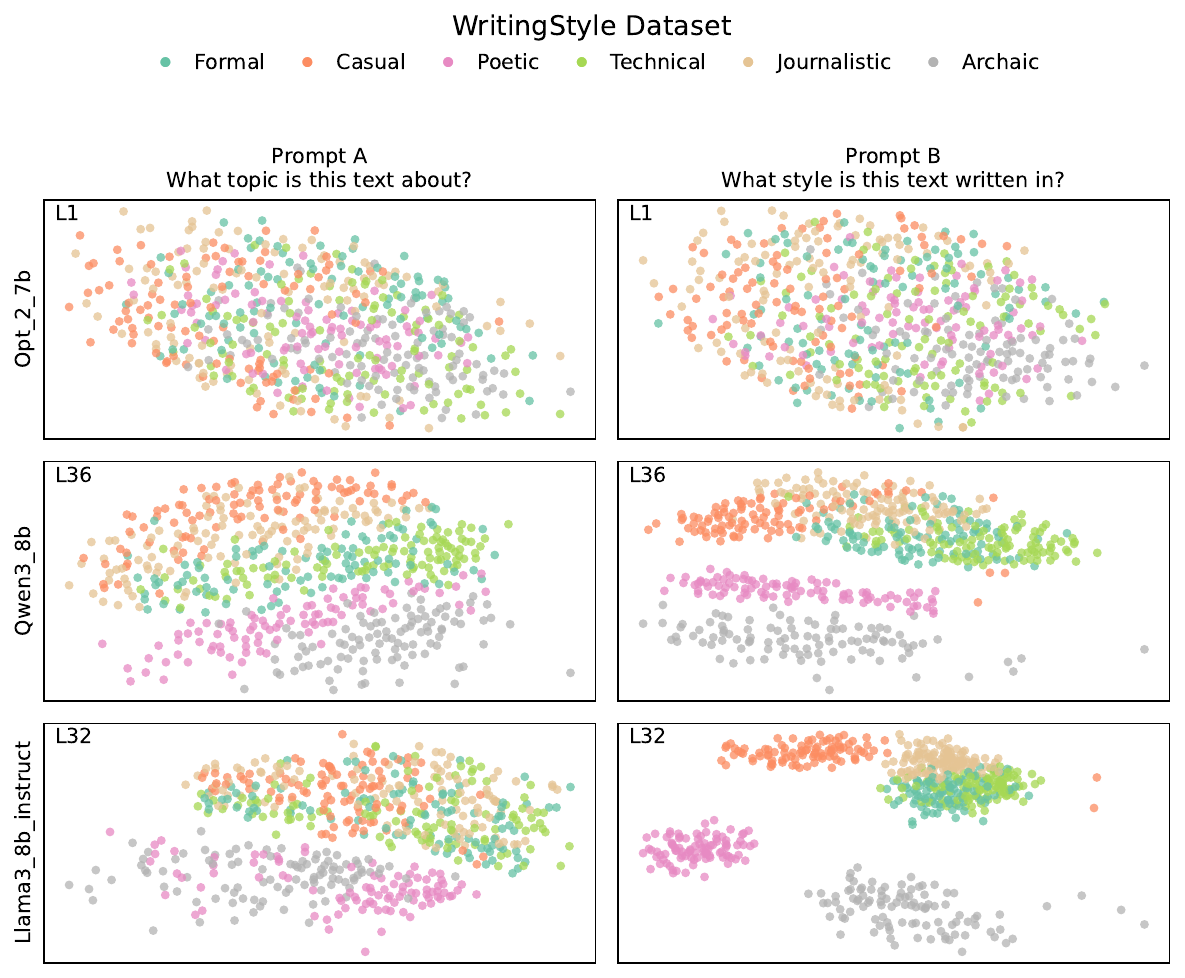}
  \caption{WritingStyle dataset, prompt $A$ (topic) vs.\ prompt $B$ (writing
  style); stimuli coloured by ground-truth style.}
  \label{fig:rds_writingstyle}
\end{figure}

\begin{figure}[h]
  \centering
  \rdsstretched{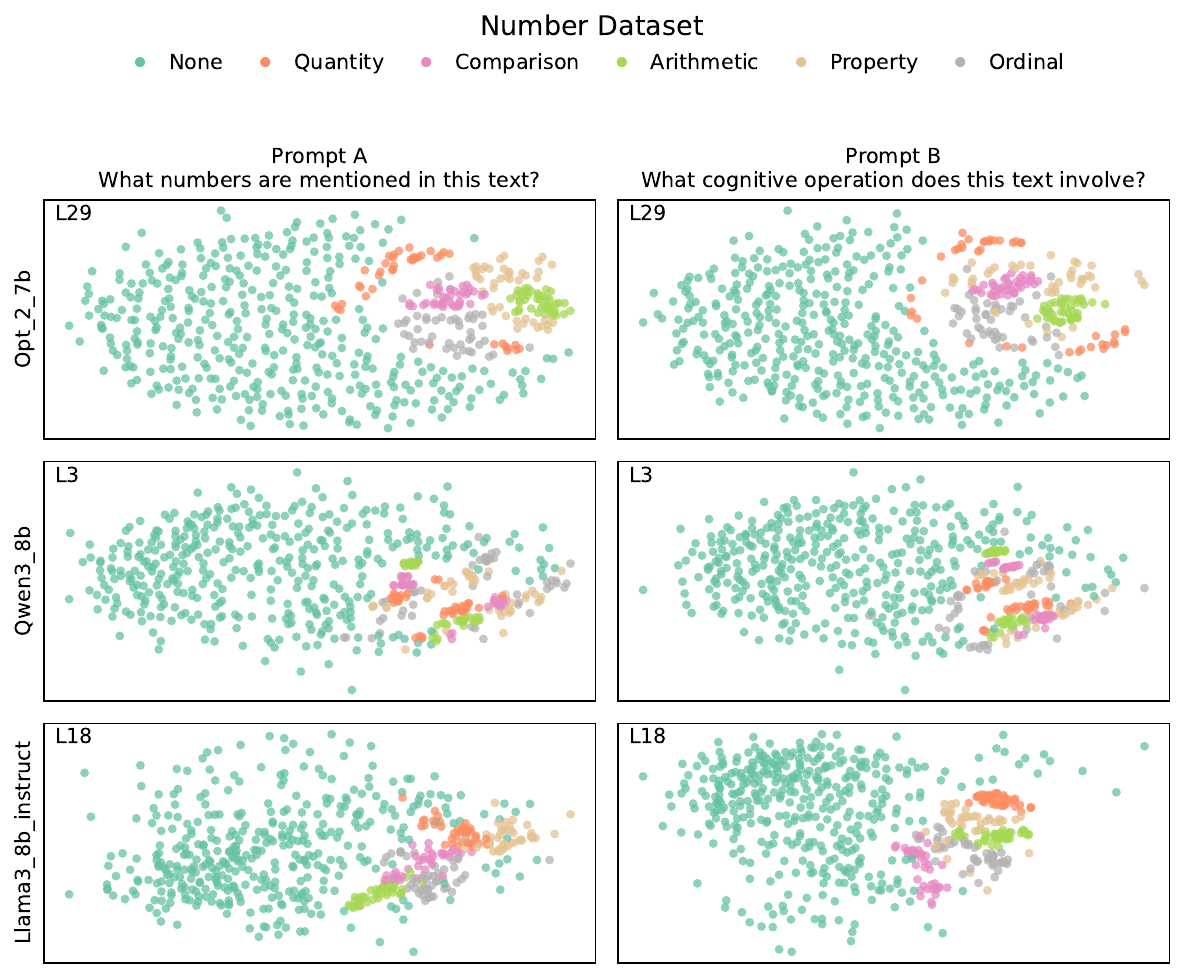}
  \caption{Number dataset, prompt $A$ (numbers mentioned) vs.\ prompt $B$
  (cognitive operation); stimuli coloured by ground-truth task
  framing.}
  \label{fig:rds_number}
\end{figure}

\begin{figure}[h]
  \centering
  \rdsstretched{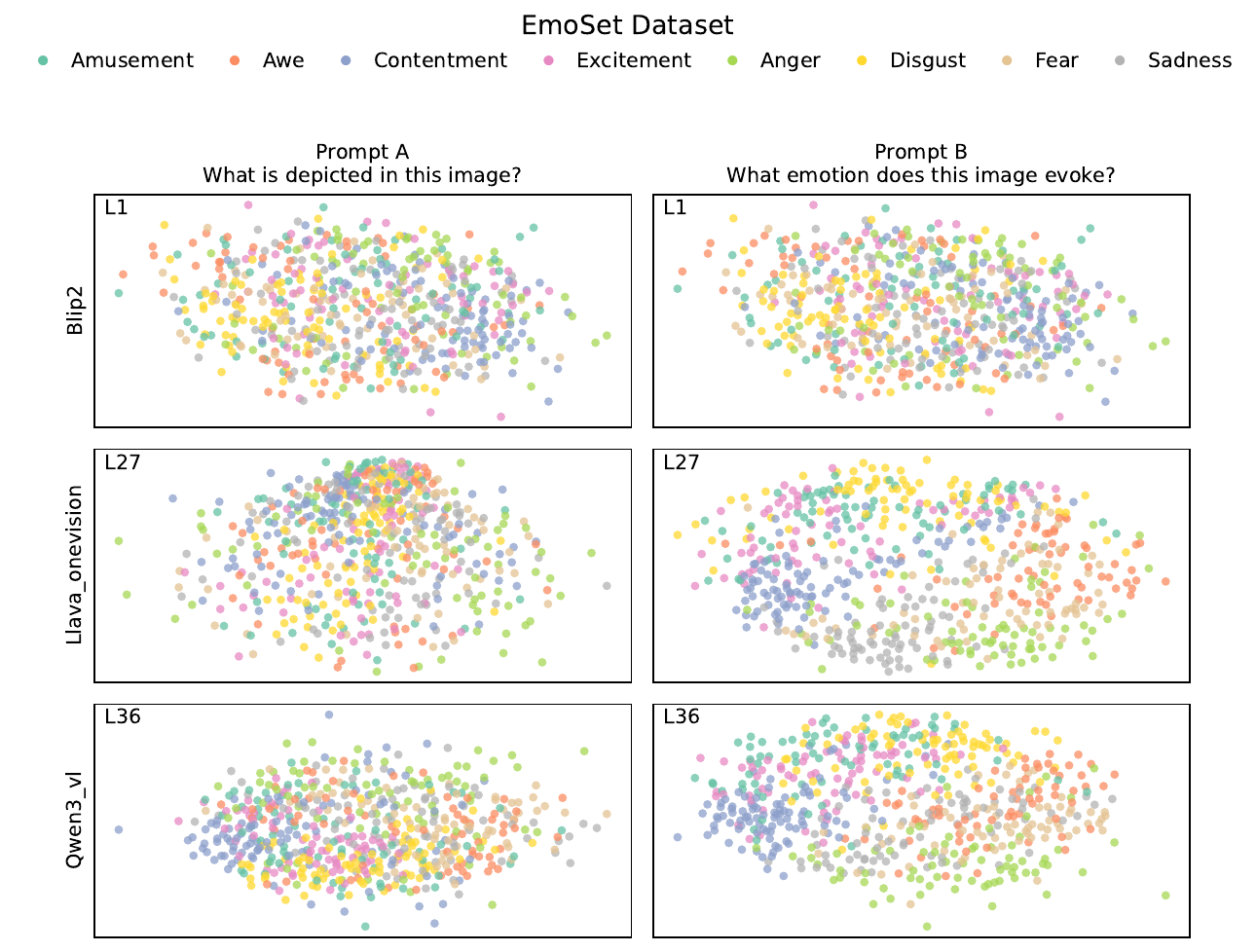}
  \caption{EmoSet dataset, prompt $A$ (image content) vs.\ prompt $B$ (emotion
  evoked); stimuli coloured by ground-truth emotion.}
  \label{fig:rds_emoset}
\end{figure}

\begin{figure}[h]
  \centering
  \includegraphics[width=\linewidth]{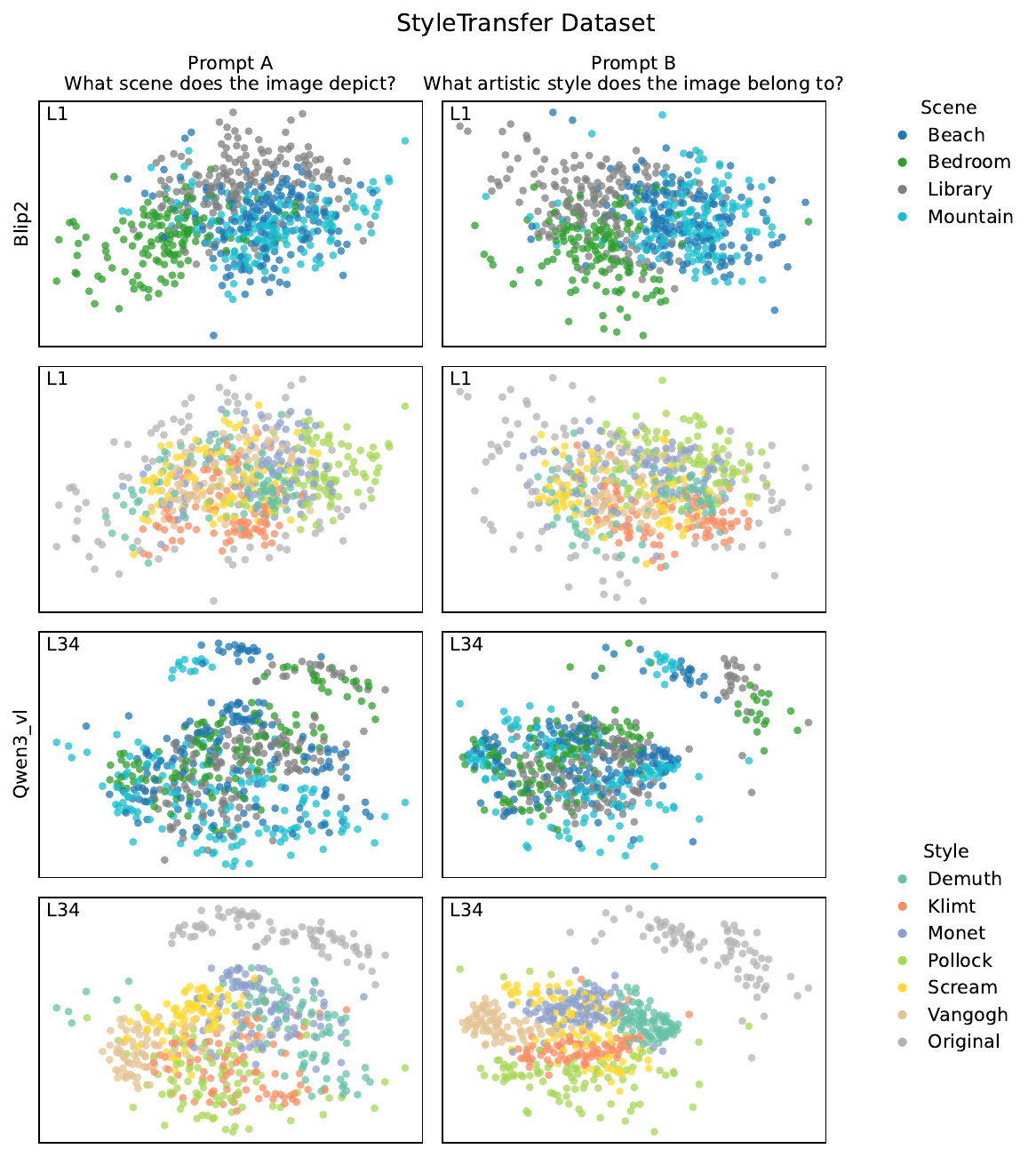}
  \caption{StyleTransfer dataset, prompt $A$ (scene) vs.\ prompt $B$
  (artistic style); stimuli coloured by ground-truth scene (top
  legend) and style (bottom legend).}
  \label{fig:rds_styletransfer}
\end{figure}

\begin{figure}[h]
  \centering
  \includegraphics[width=\linewidth]{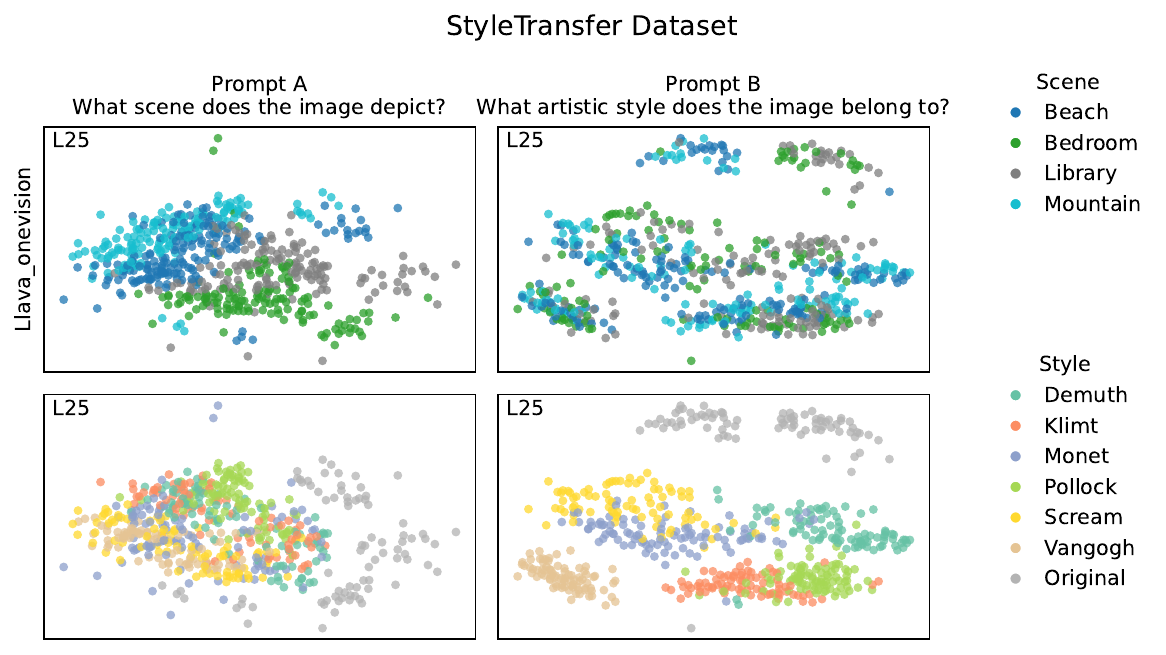}
  \caption{StyleTransfer dataset (continued), prompt $A$ (scene) vs.\
  prompt $B$ (artistic style); stimuli coloured by ground-truth scene
  (top) and style (bottom).}
  \label{fig:rds_styletransfer_llava}
\end{figure}

\begin{figure}[h]
  \centering
  \includegraphics[width=\linewidth]{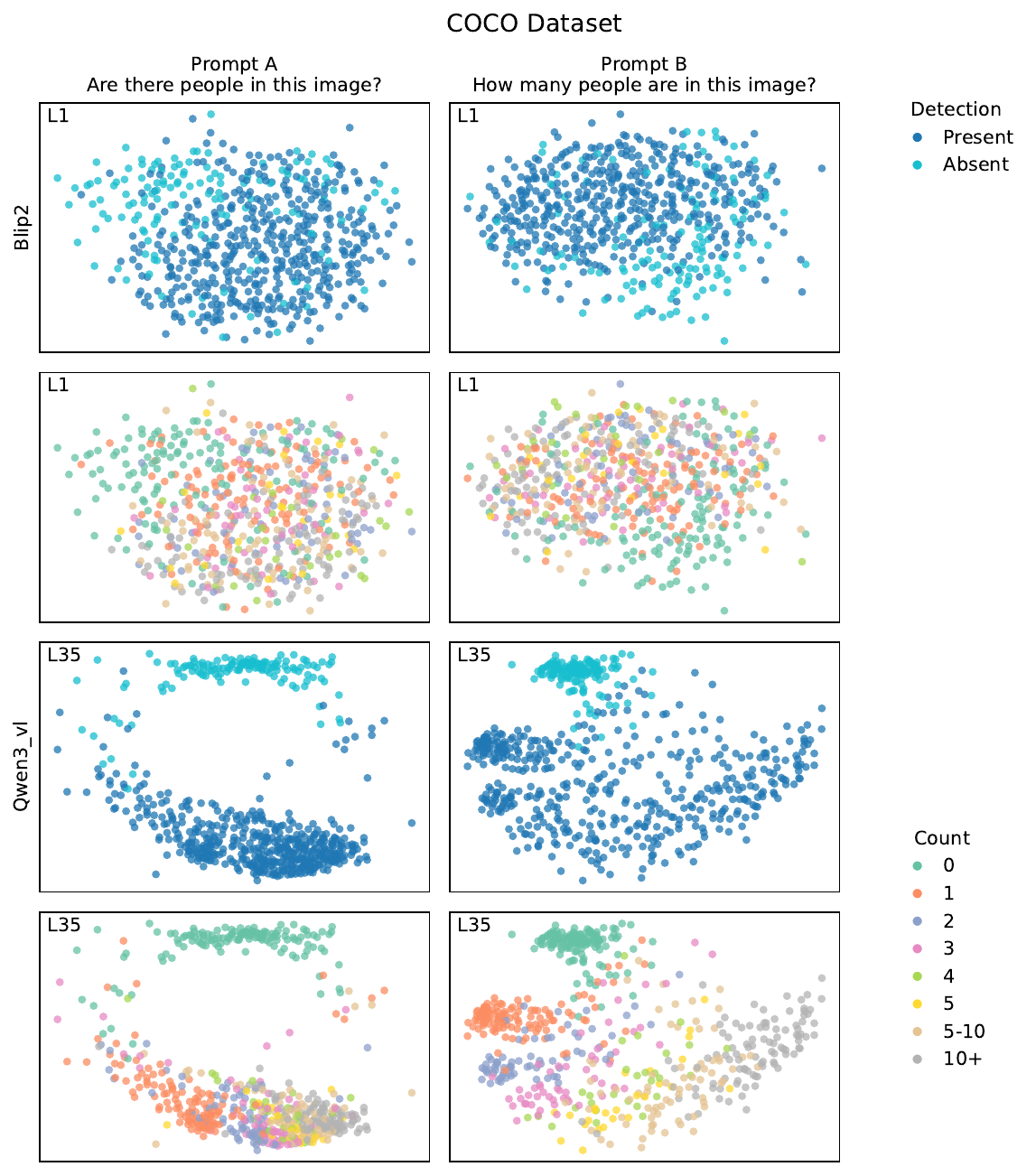}
  \caption{COCO dataset, prompt $A$ (people detection) vs.\ prompt $B$ (people
  count); stimuli coloured by ground-truth detection (top) and count
  bin (bottom).}
  \label{fig:rds_coco}
\end{figure}

\clearpage
\section{RSA and silhouette score}
\label{app:rsa}

For each pair of model and dataset, we report the layerwise Spearman
correlation between the data RDM of the prompt-induced hidden
states and the prompt-$B$ target attribute RDM, alongside the
silhouette score based on the target-labels.  

\begin{figure}[h]                                                                             
    \centering
    \includegraphics[width=\linewidth]{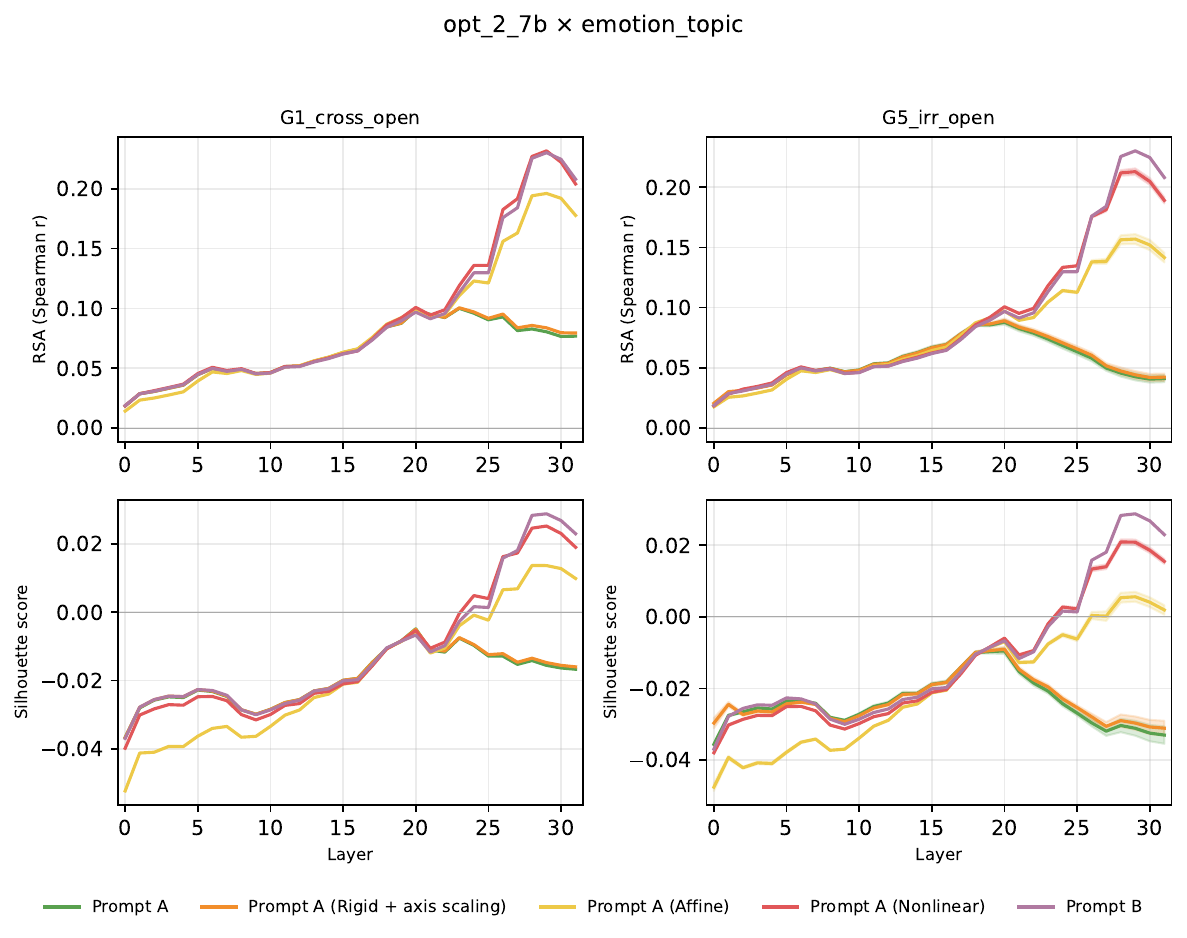}\\[2pt]       
    \includegraphics[width=\linewidth]{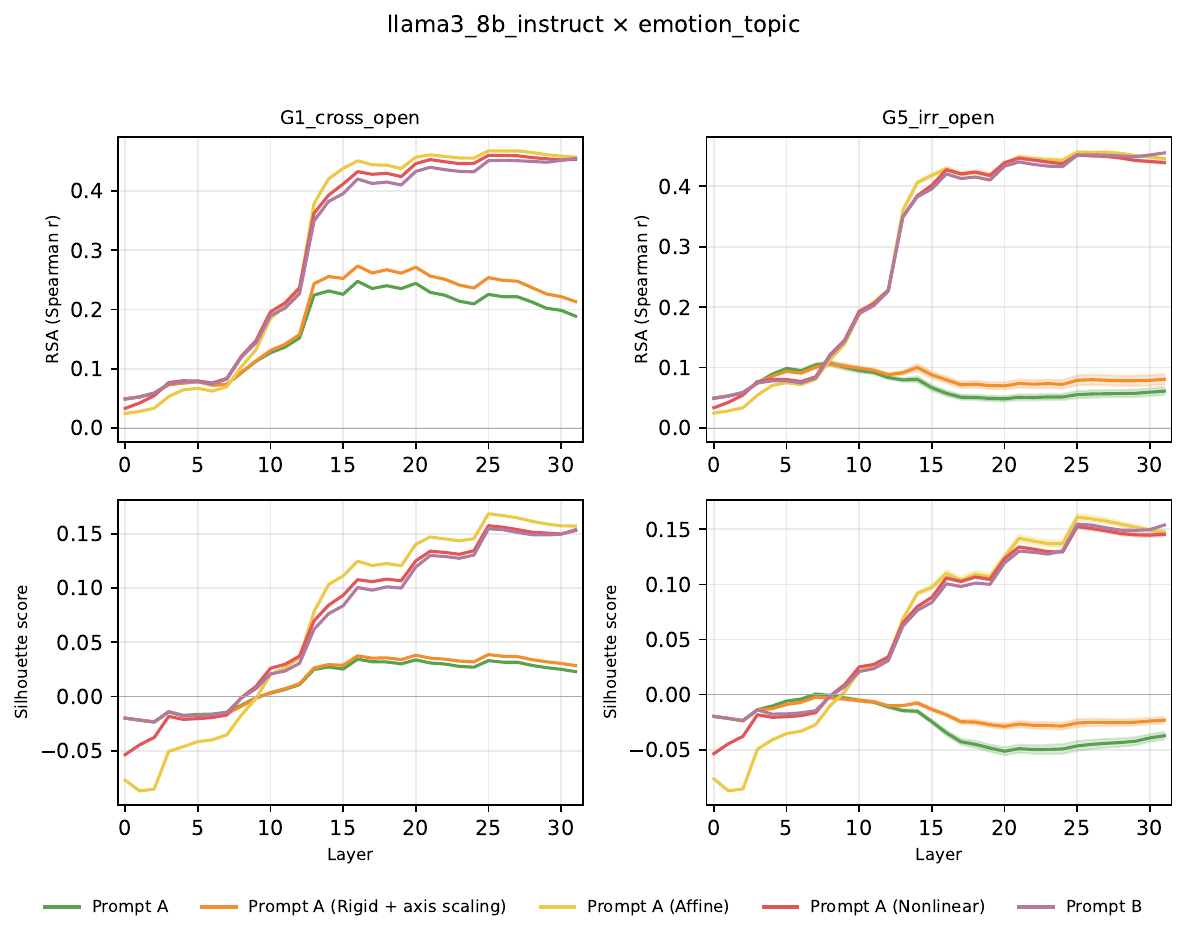}      
    \caption{Layerwise RDM correlation (left) 
    and silhouette score (right) for EmotionalStory (1/2) under prompt $A$ vs.\ prompt $B$ for OPT-2.7B (top) and Llama-3-8B-Instruct (bottom).}         
    \label{fig:rsa_emostory_a}      
  \end{figure}      
  \begin{figure}[h]                                             
    \centering
    \includegraphics[width=\linewidth]{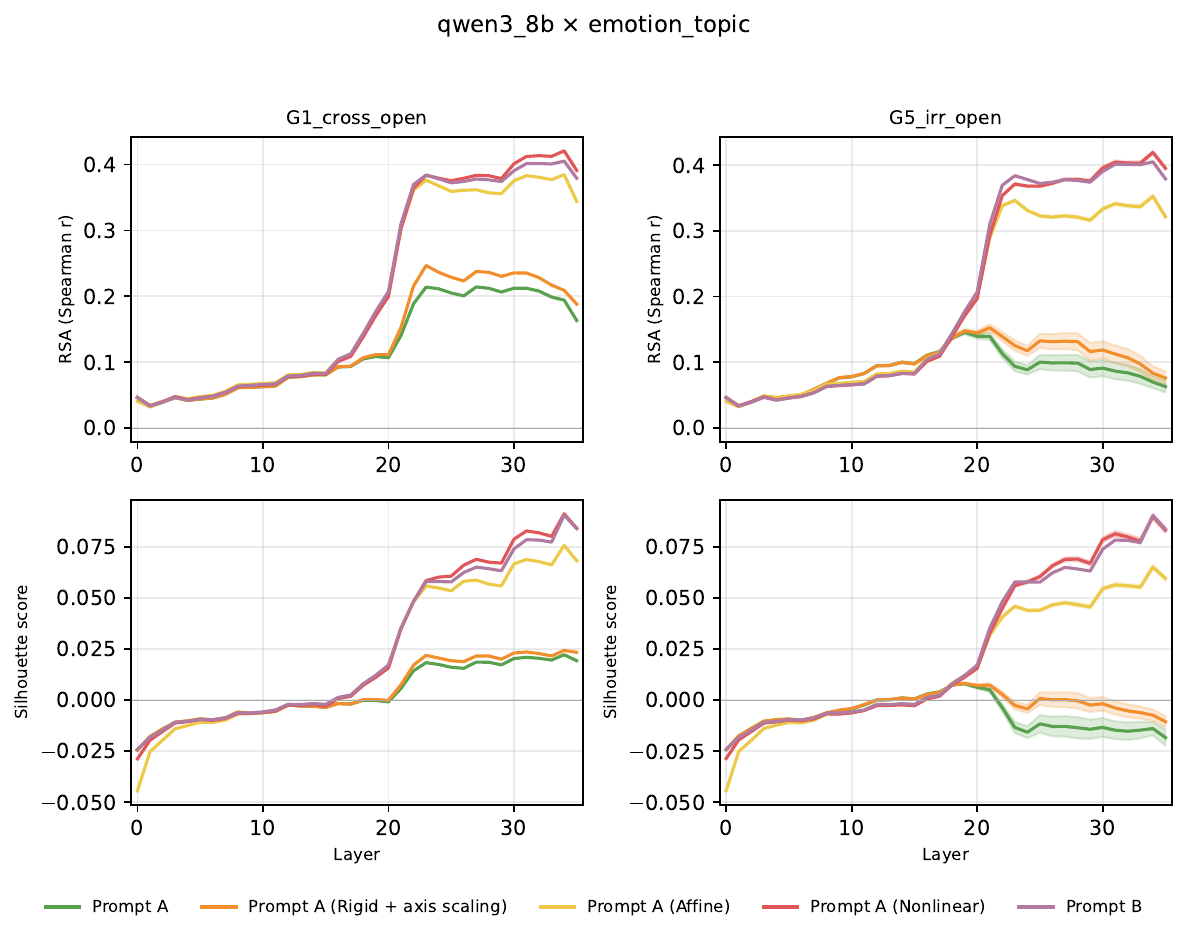}
    \caption{Layerwise RDM correlation (left) 
    and silhouette score (right) for EmotionalStory (2/2) under prompt $A$ vs.\ prompt $B$ for for Qwen3-8B.}                                                                                                                                                          
    \label{fig:rsa_emostory_b} 
  \end{figure}   

\begin{figure}[h]
  \centering
  \includegraphics[width=\linewidth]{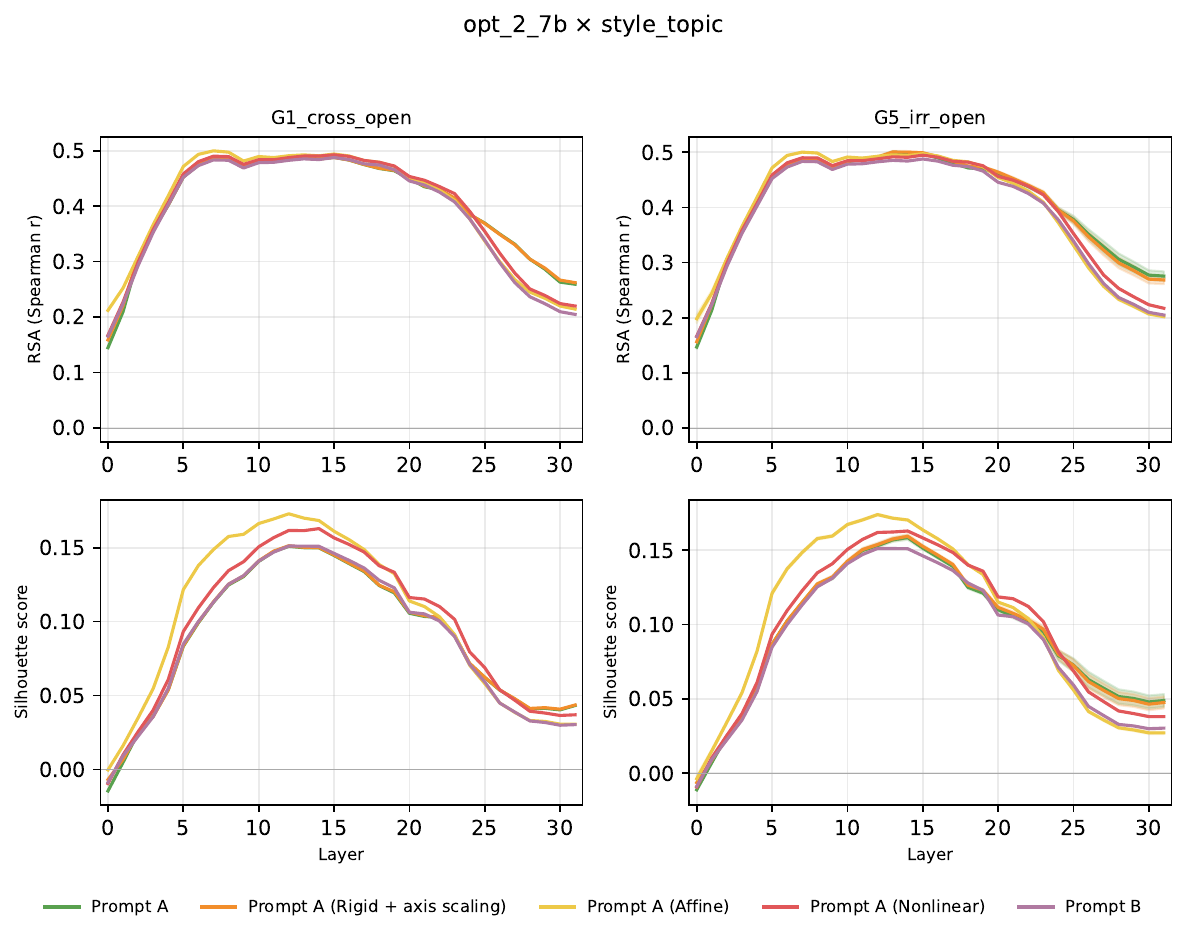}\\[2pt]
  \includegraphics[width=\linewidth]{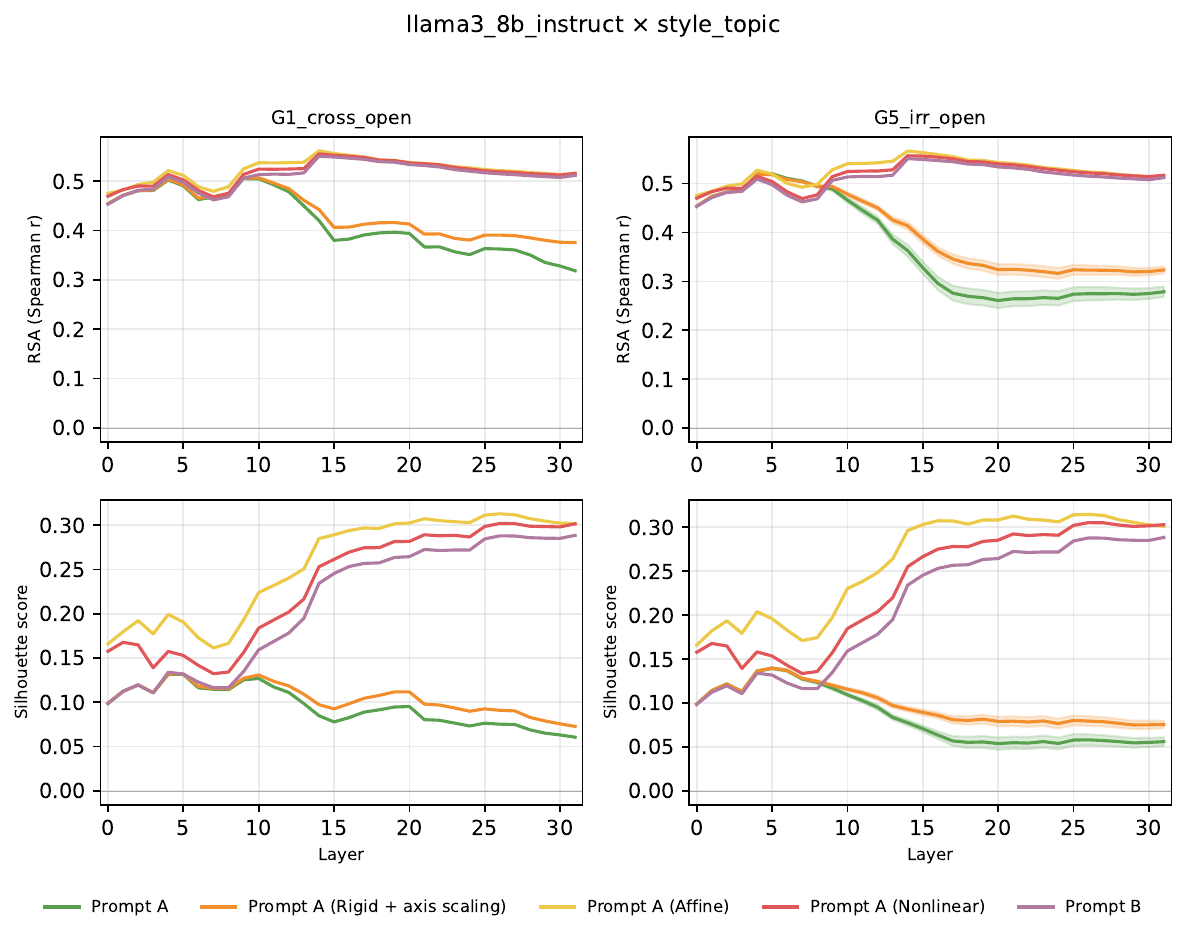}
  \caption{Layerwise RDM correlation (left) 
    and silhouette score (right) for WritingStyle (1/2) under prompt $A$ vs.\ prompt $B$ for OPT-2.7B (top) and Llama-3-8B-Instruct (bottom).}
  \label{fig:rsa_writingstyle}
\end{figure}
\begin{figure}[h]
  \centering
  \includegraphics[width=\linewidth]{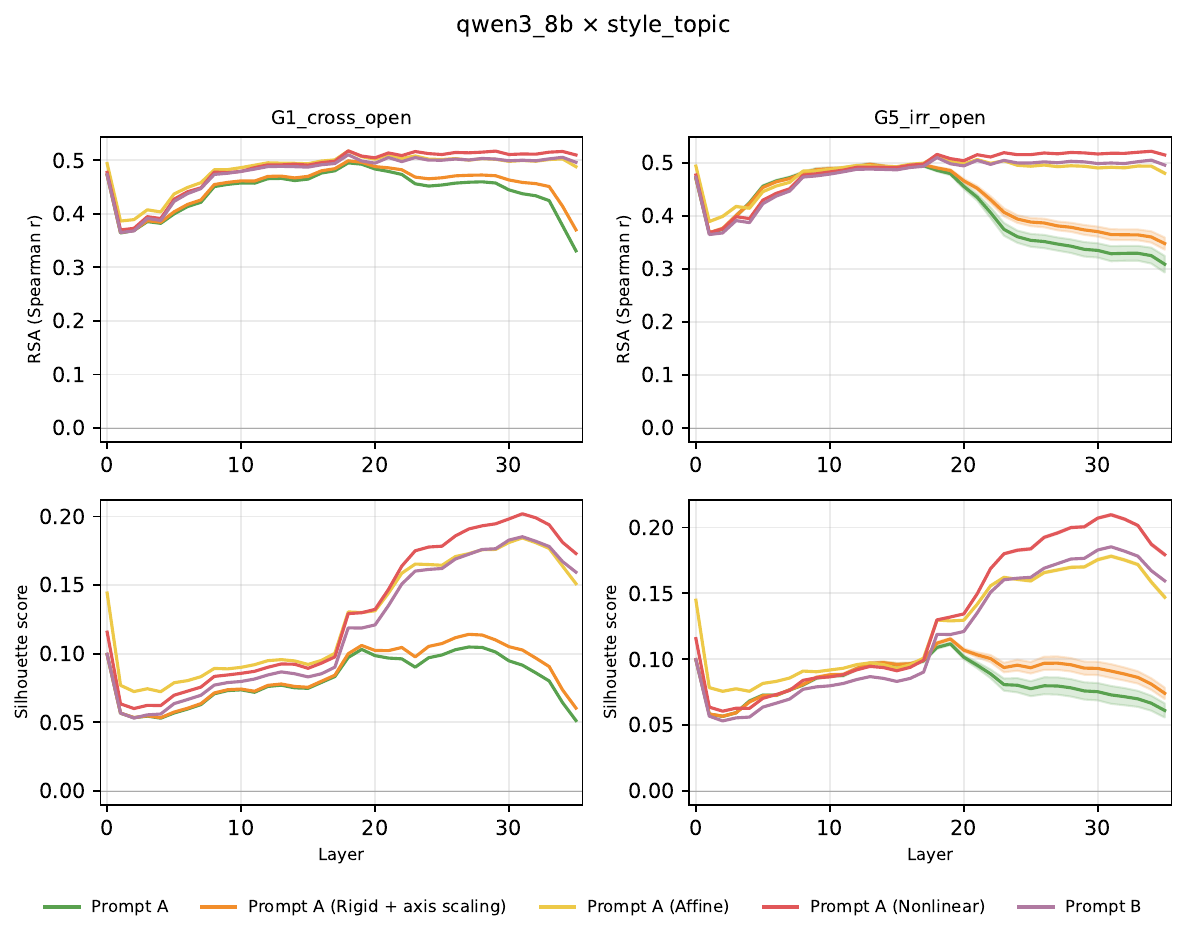}
  \caption{Layerwise RDM correlation (left) 
    and silhouette score (right) for WritingStyle (2/2) under prompt $A$ vs.\ prompt $B$ for for Qwen3-8B.}
  \label{fig:rsa_writingstyle}
\end{figure}

\begin{figure}[h]
  \centering
  \includegraphics[width=\linewidth]{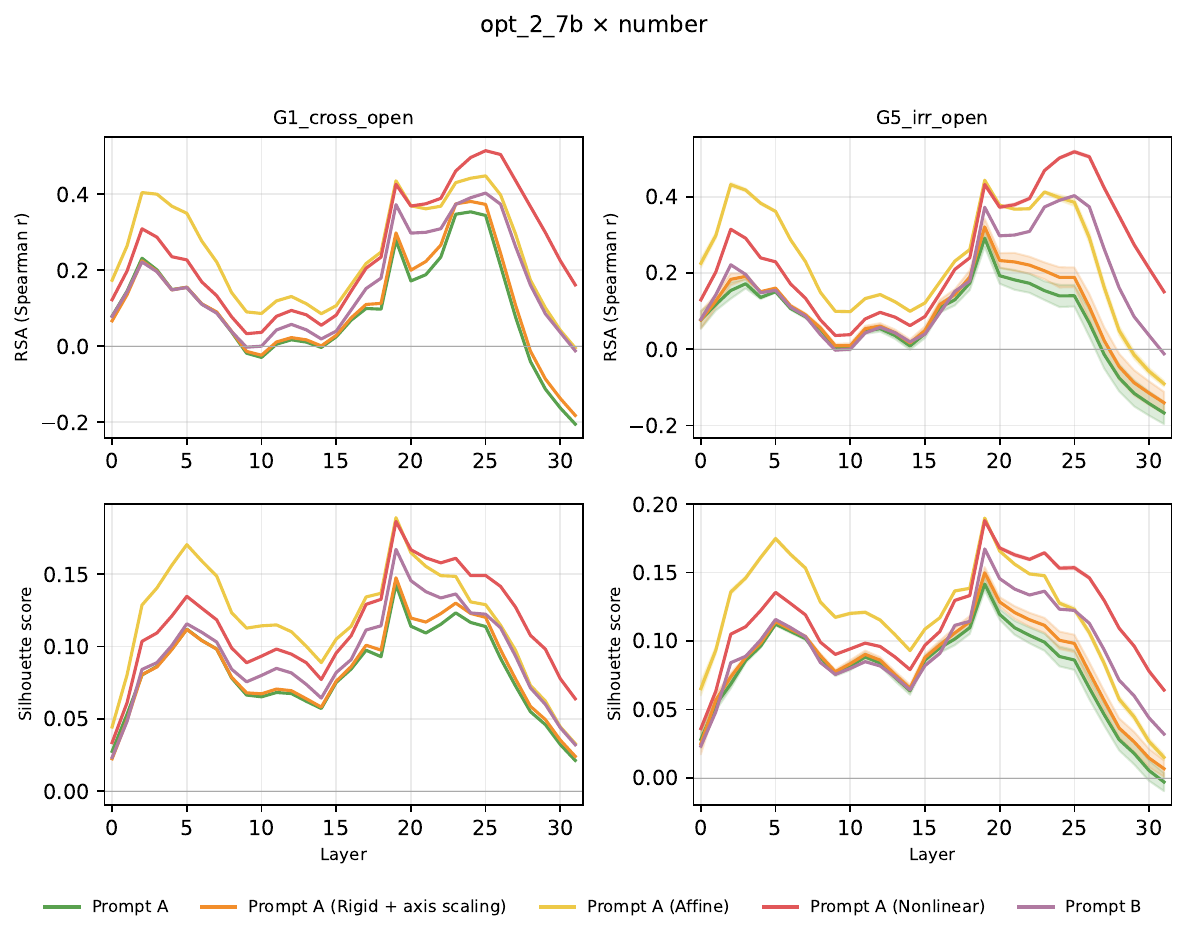}\\[2pt]
  \includegraphics[width=\linewidth]{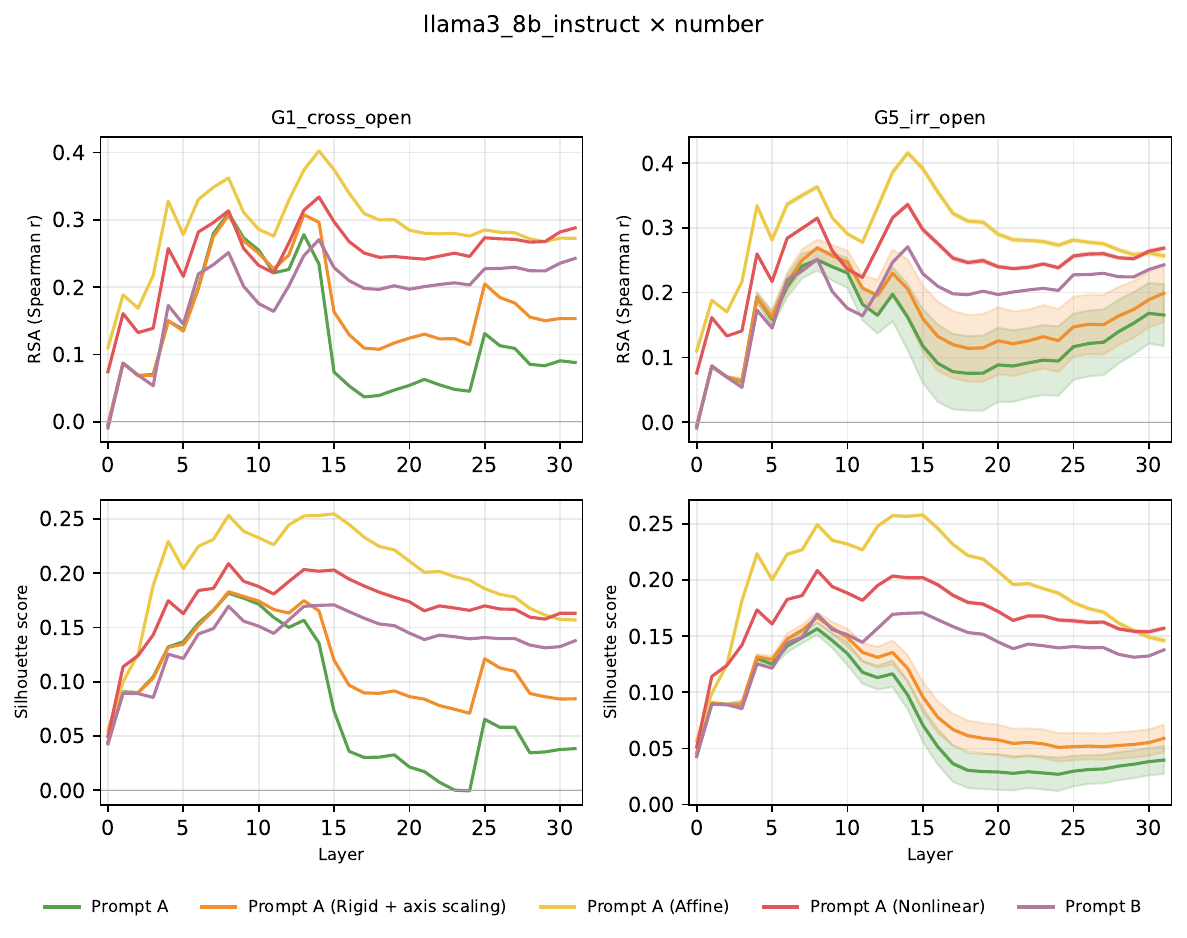}
  \caption{Layerwise RDM correlation (left) 
    and silhouette score (right) for Number (1/2) under prompt $A$ vs.\ prompt $B$ for OPT-2.7B (top) and Llama-3-8B-Instruct (bottom). }
  \label{fig:rsa_number}
\end{figure}

\begin{figure}[h]
  \centering
  \includegraphics[width=\linewidth]{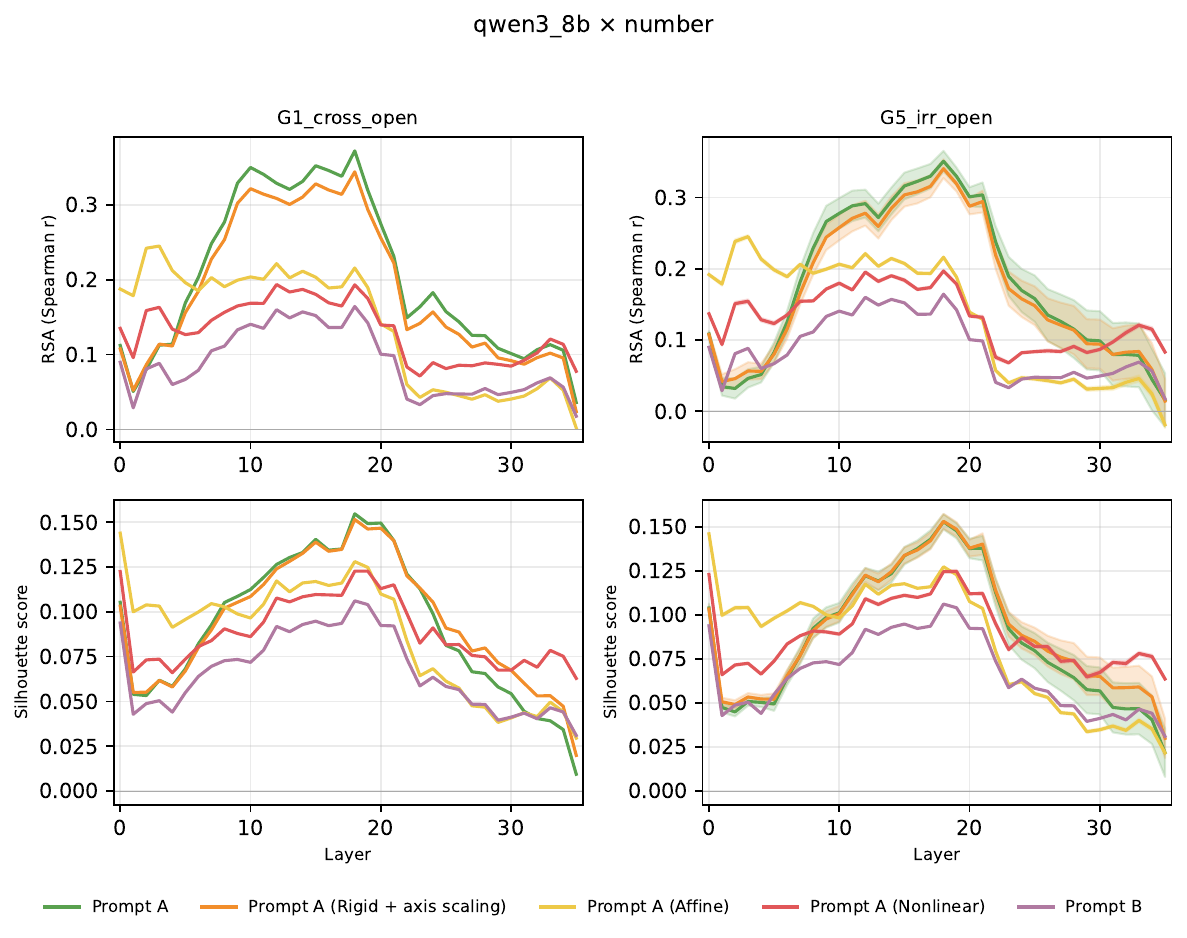}
  \caption{Layerwise RDM correlation (left) 
    and silhouette score (right) for Number (2/2) under prompt $A$ vs.\ prompt $B$ for for Qwen3-8B.}
  \label{fig:rsa_number}
\end{figure}

\begin{figure}[h]
  \centering
  \includegraphics[width=\linewidth]{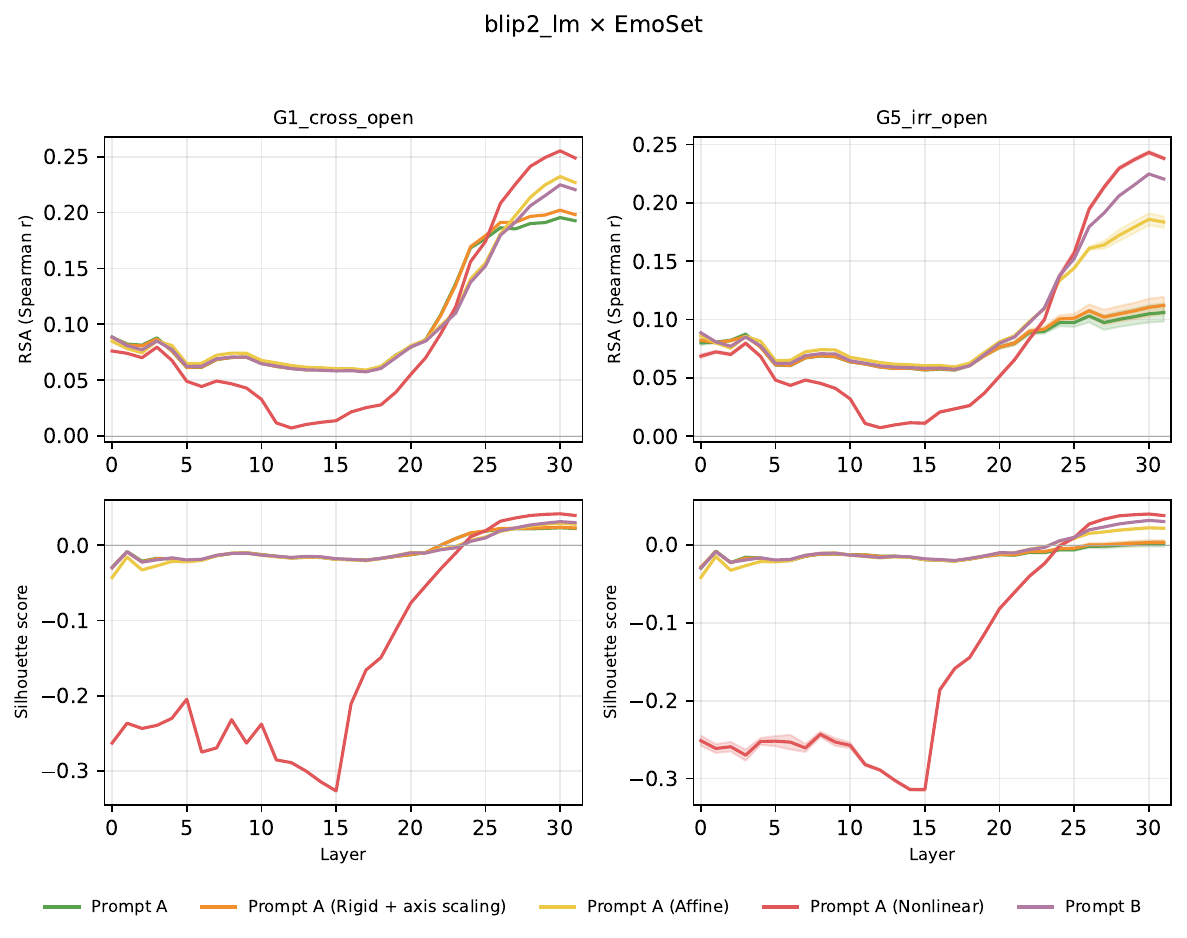}\\[2pt]
  \includegraphics[width=\linewidth]{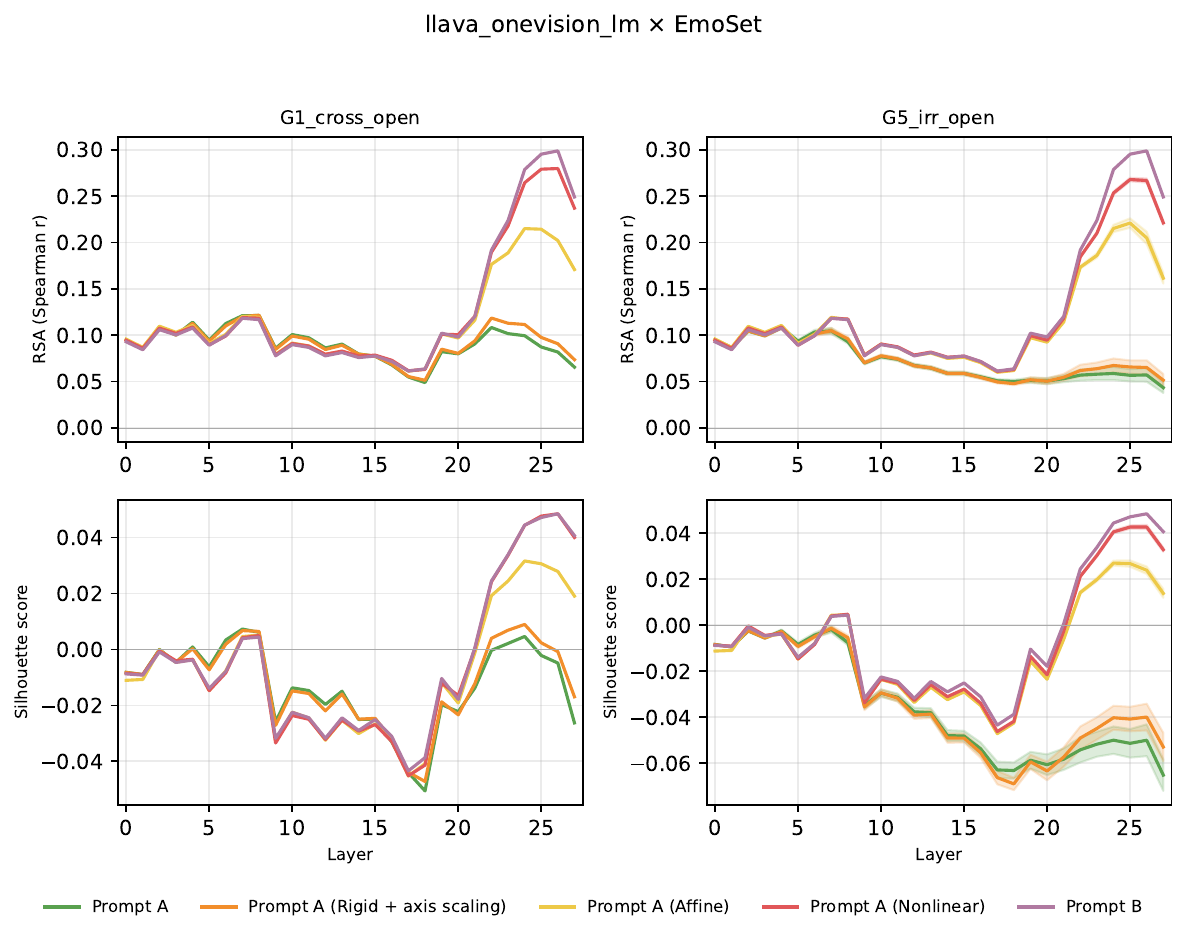}
  \caption{Layerwise RDM correlation (left) 
    and silhouette score (right) for EmoSet (1/2) under prompt $A$ vs.\ prompt $B$ for OPT-2.7B (top) and Llama-3-8B-Instruct (bottom). }
  \label{fig:rsa_emoset}
\end{figure}

\begin{figure}[h]
  \centering
  \includegraphics[width=\linewidth]{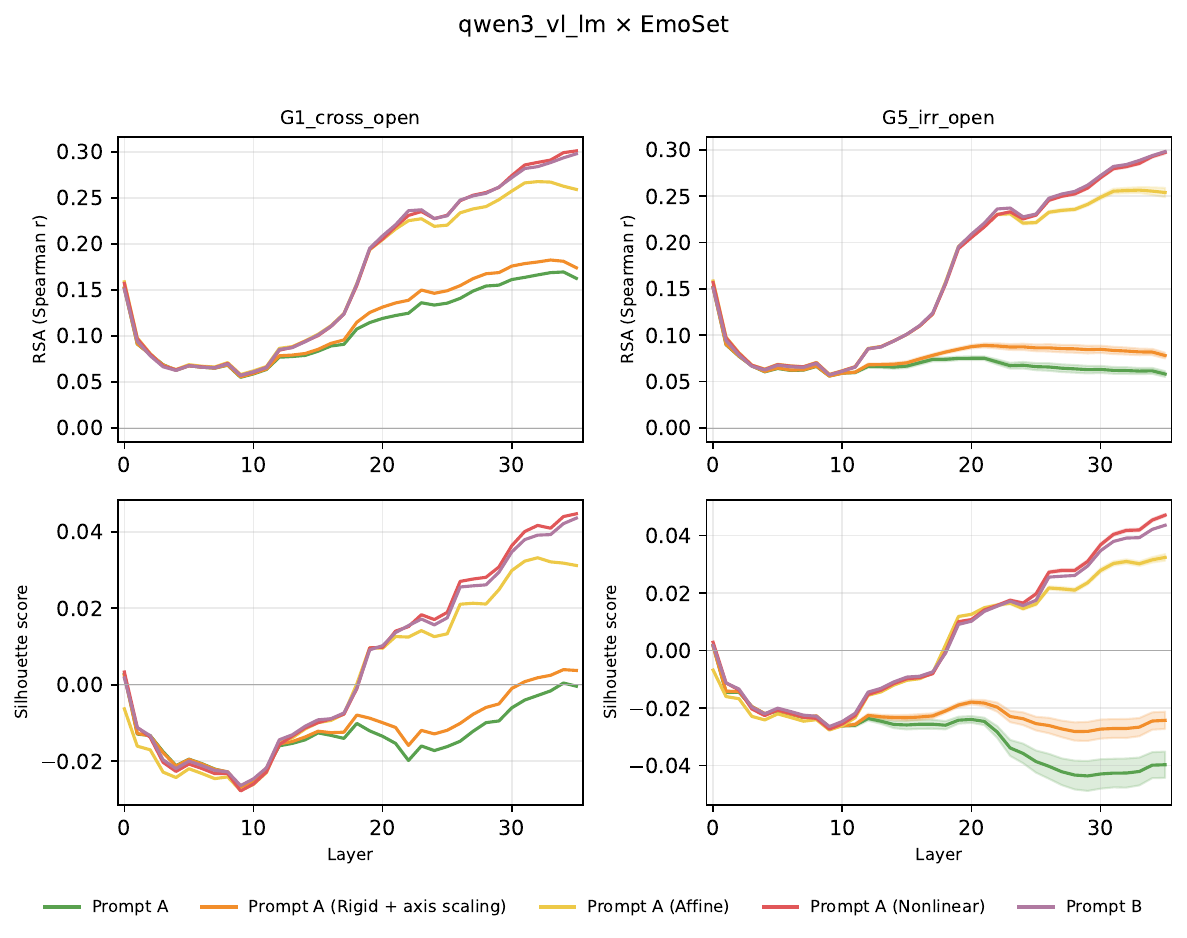}
  \caption{Layerwise RDM correlation (left) 
    and silhouette score (right) for EmoSet (2/2) under prompt $A$ vs.\ prompt $B$ for for Qwen3-8B.}
  \label{fig:rsa_emoset}
\end{figure}

\begin{figure}[h]
  \centering
  \includegraphics[width=\linewidth]{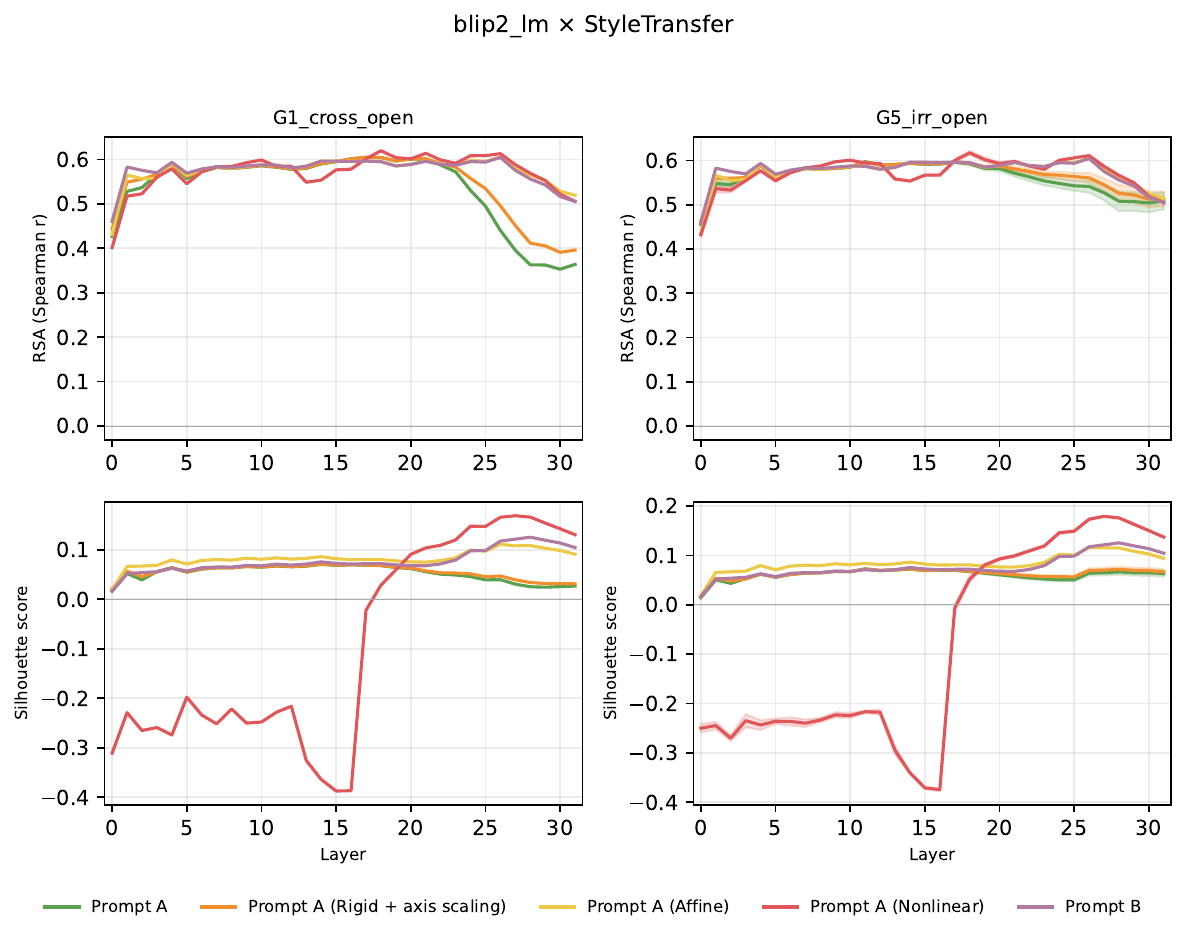}\\[2pt]
  \includegraphics[width=\linewidth]{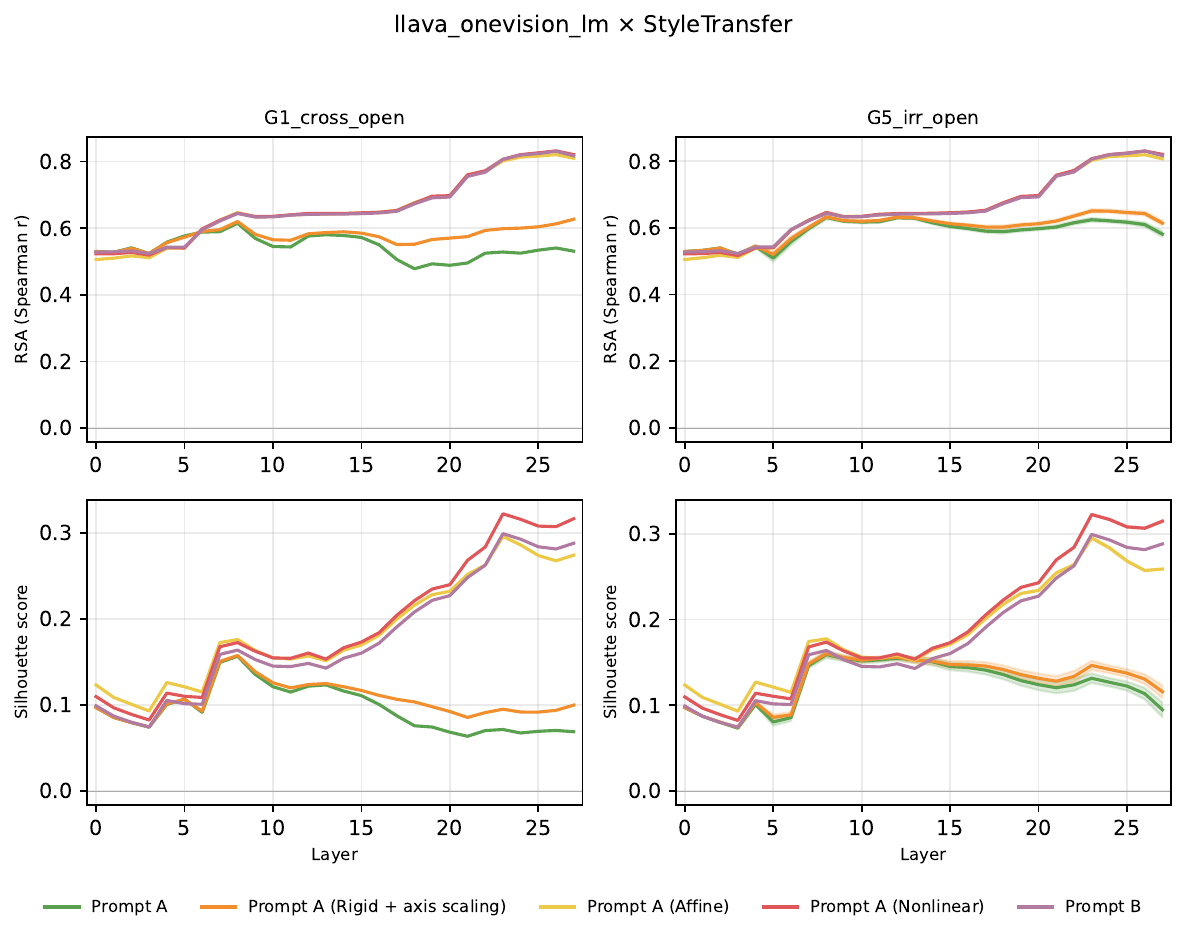}
  \caption{Layerwise RDM correlation (left) 
    and silhouette score (right) for StyleTransfer (1/2) under prompt $A$ vs.\ prompt $B$ for OPT-2.7B (top) and Llama-3-8B-Instruct (bottom). }
  \label{fig:rsa_styletransfer}
\end{figure}

\begin{figure}[h]
  \centering
  \includegraphics[width=\linewidth]{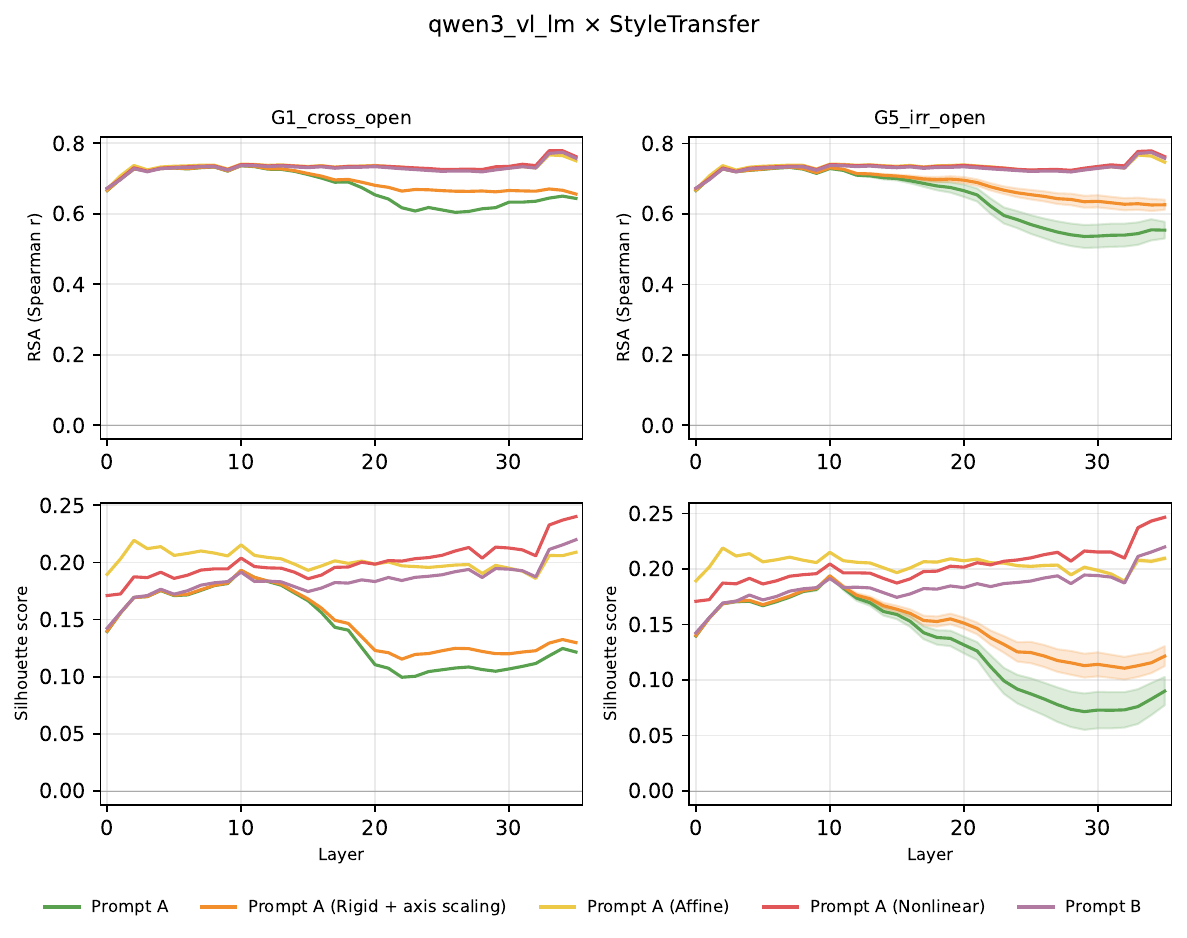}
  \caption{Layerwise RDM correlation (left) 
    and silhouette score (right) for StyleTransfer (2/2) under prompt $A$ vs.\ prompt $B$ for for Qwen3-8B.}
  \label{fig:rsa_styletransfer}
\end{figure}

\begin{figure}[h]
  \centering
  \includegraphics[width=\linewidth]{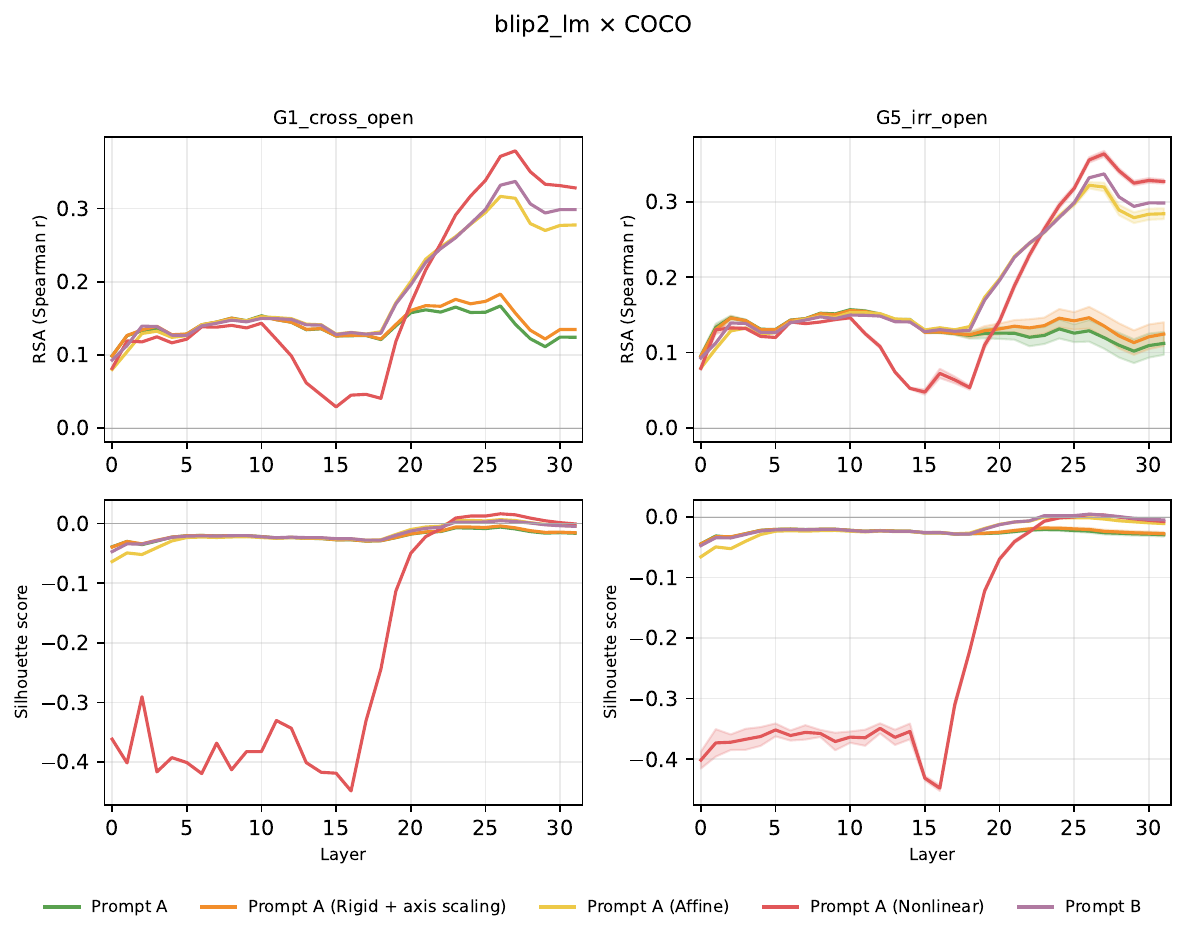}\\[2pt]
  \includegraphics[width=\linewidth]{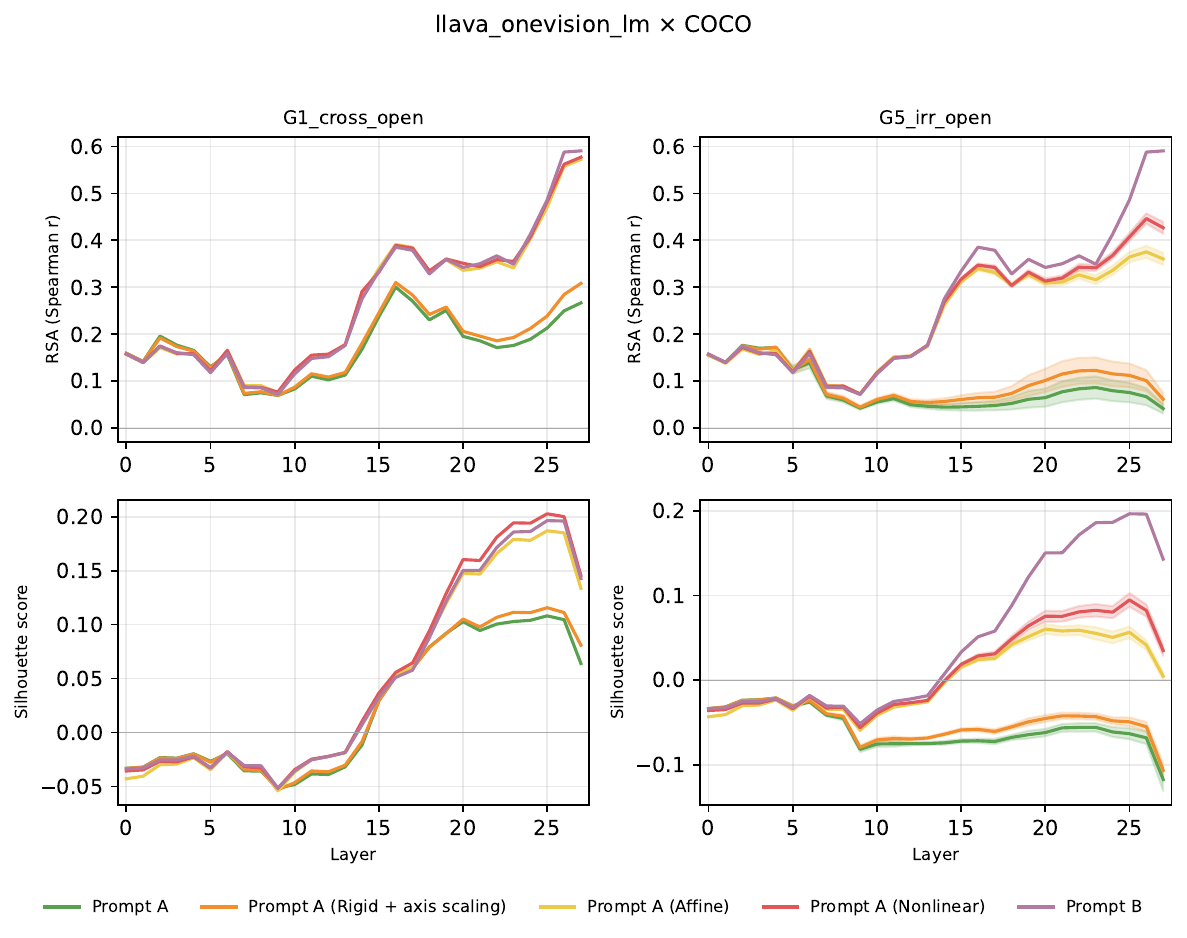}
  \caption{Layerwise RDM correlation (left) 
    and silhouette score (right) for COCO (1/2) under prompt $A$ vs.\ prompt $B$ for OPT-2.7B (top) and Llama-3-8B-Instruct (bottom).}
  \label{fig:rsa_coco}
\end{figure}

\begin{figure}[h]
  \centering
  \includegraphics[width=\linewidth]{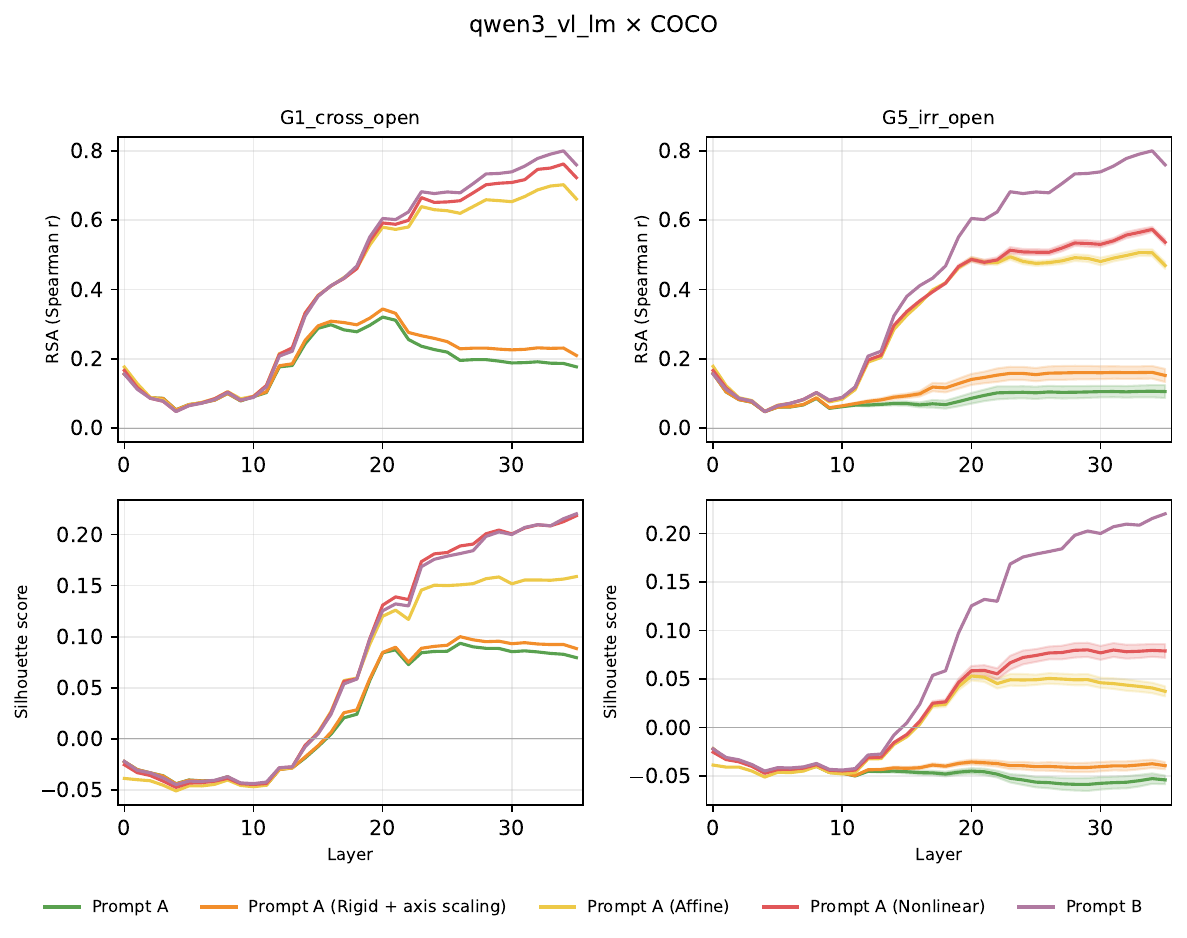}
  \caption{Layerwise RDM correlation (left) 
    and silhouette score (right) for COCO (2/2) under prompt $A$ vs.\ prompt $B$ for for Qwen3-8B.}
  \label{fig:rsa_coco}
\end{figure}

\clearpage
\section{Decomposition results for individual prompt-pair groups}
\label{app:r2bars}

For each (model, dataset) cell we show the full five-tier incremental
cross-validated $R^{2}$ at every layer as a stacked bar. Each segment
is $\Delta R^{2}_{k}=R^{2}_{k}-R^{2}_{k-1}$ for one tier in the nested
chain (translation $T$, rigid transformation with uniform scaling $O_u$, rigid transformation with axis-wise scaling $O_{a}$, affine $L$, nonlinear $N$).
These visualizations complement the summary
Fig.~\ref{fig:r2_decomposition} and make visible the
\emph{strategy-level} differences across model families discussed in
\S\ref{sec:results_r2_decomposition}. 

\begin{figure}[h]
  \centering
  \includegraphics[width=\linewidth]{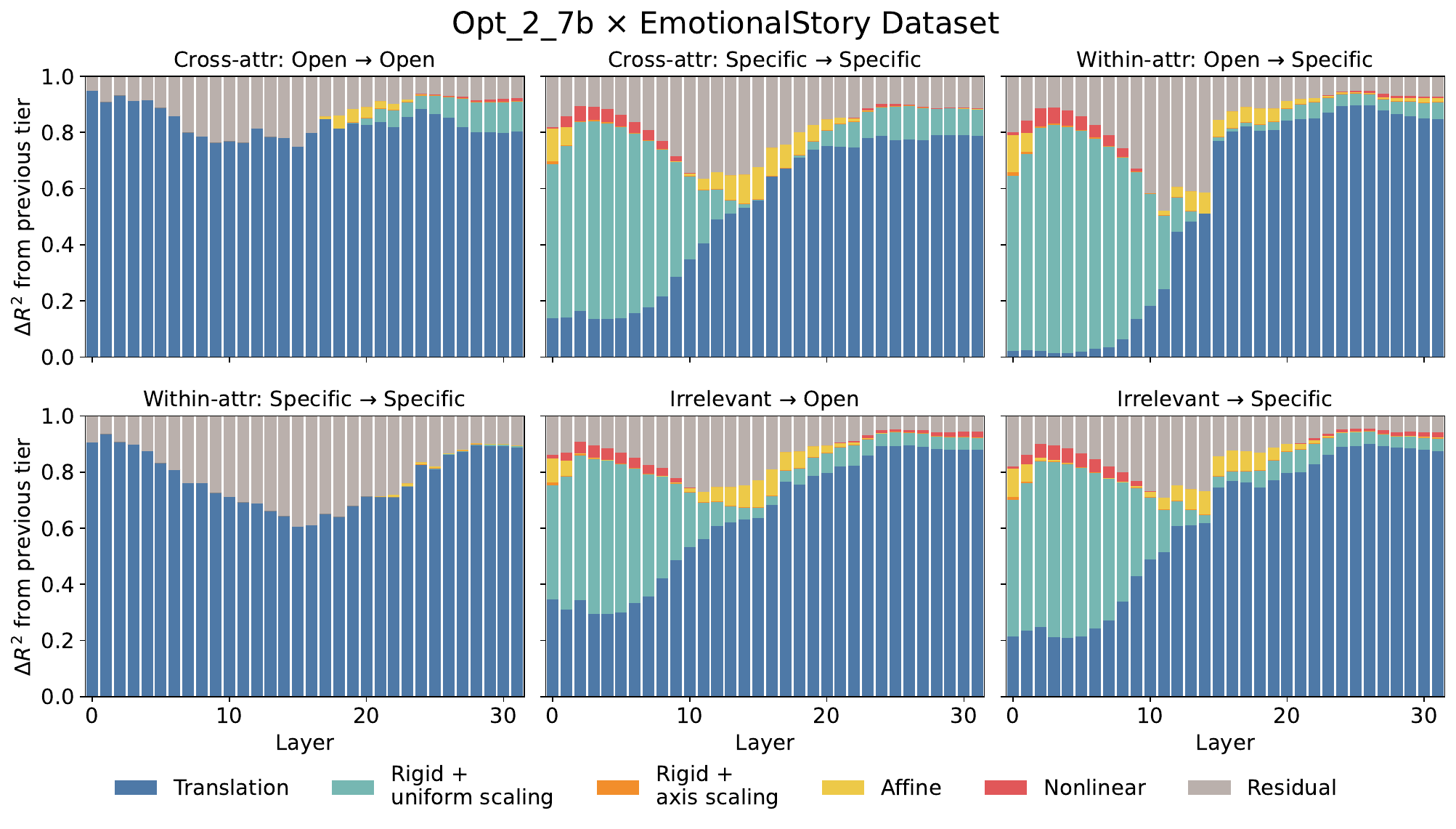}\\[2pt]
  \includegraphics[width=\linewidth]{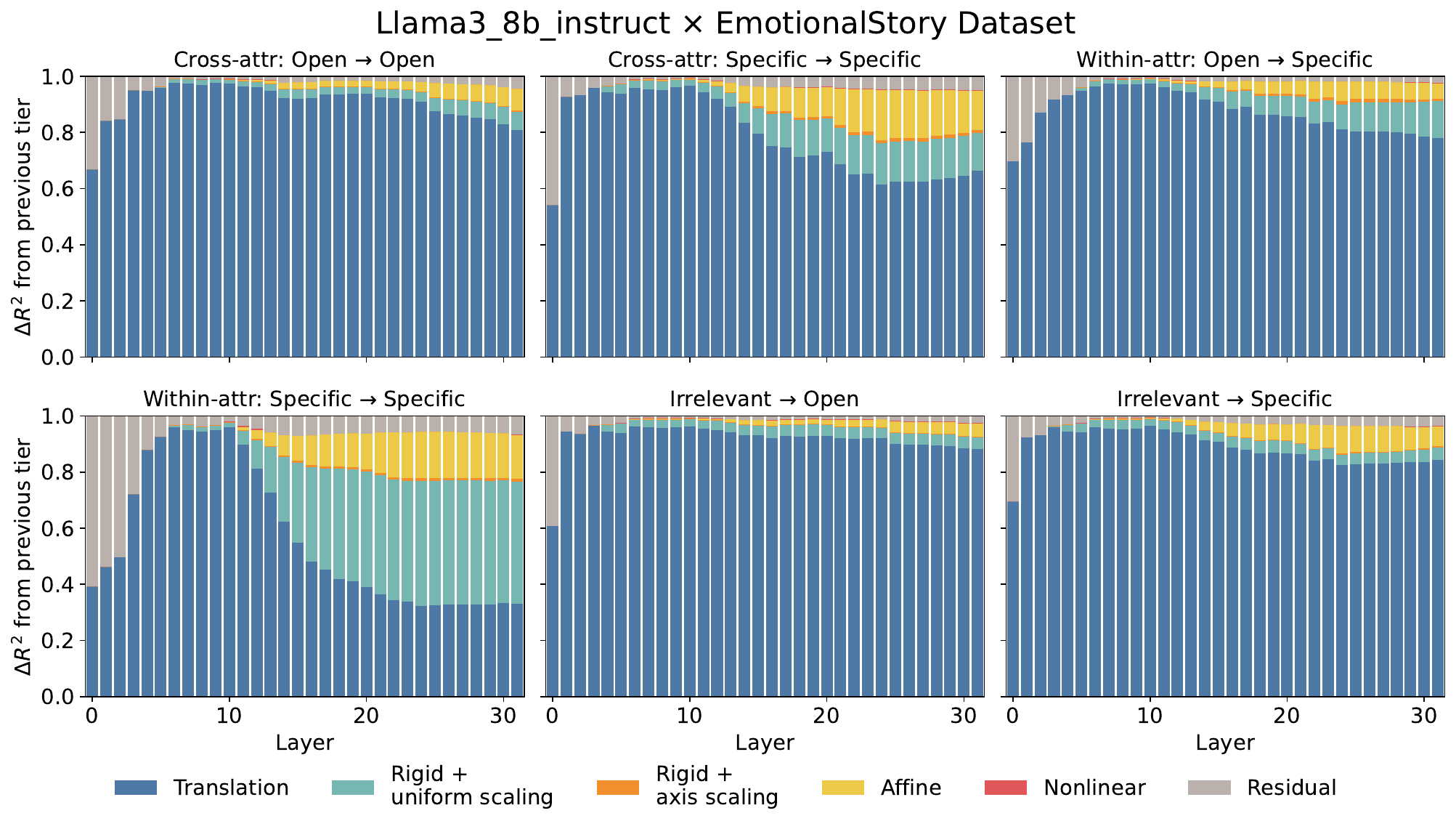}\\[2pt]
  \includegraphics[width=\linewidth]{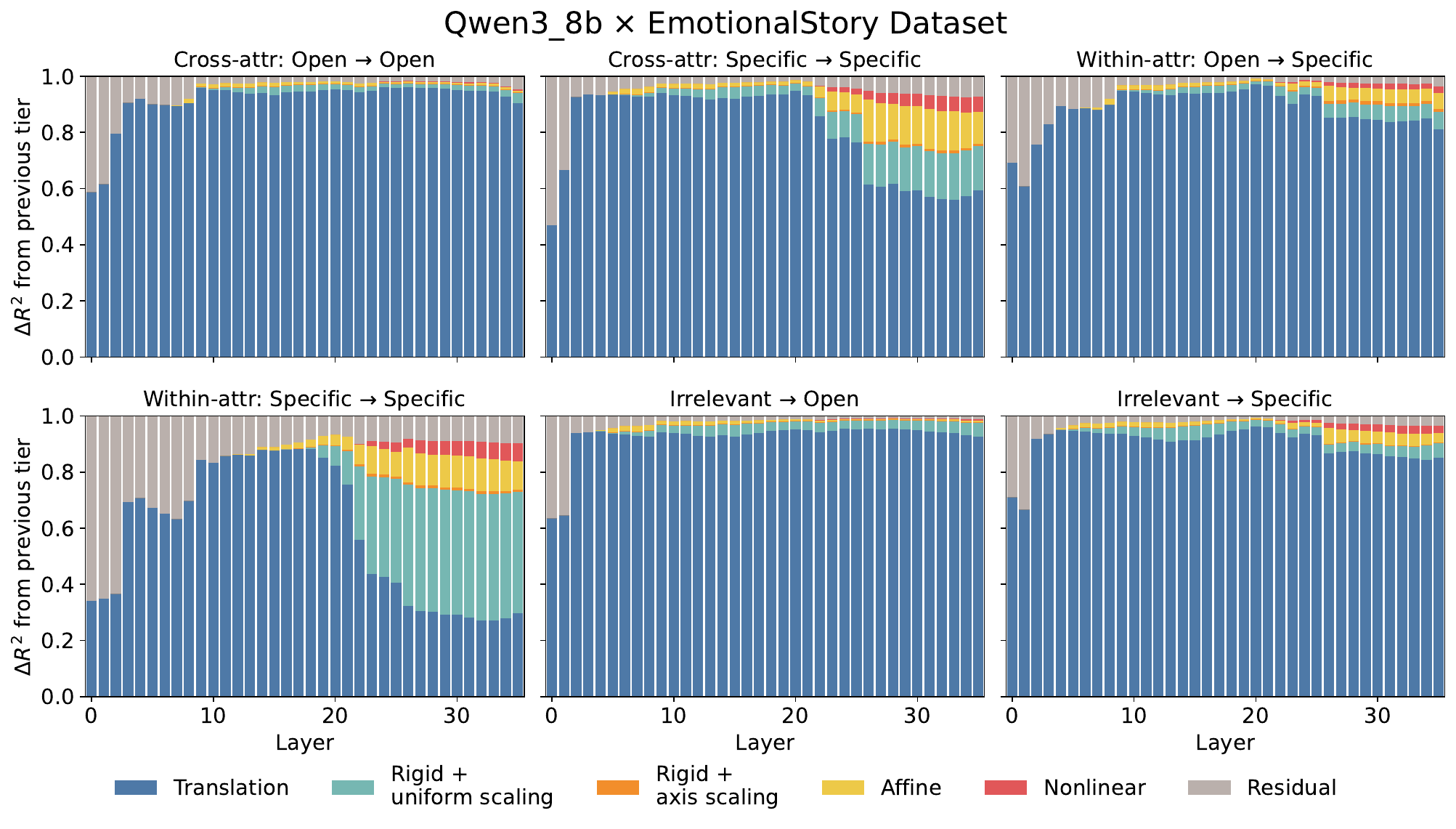}
  \caption{Incremental $R^{2}$ of each transformation for EmotionalStory dataset across
  layers for OPT-2.7B (top), Llama-3-8B-Instruct (middle), Qwen3-8B
  (bottom).}
  \label{fig:r2bar_emotionalstory}
\end{figure}

\begin{figure}[h]
  \centering
  \includegraphics[width=\linewidth]{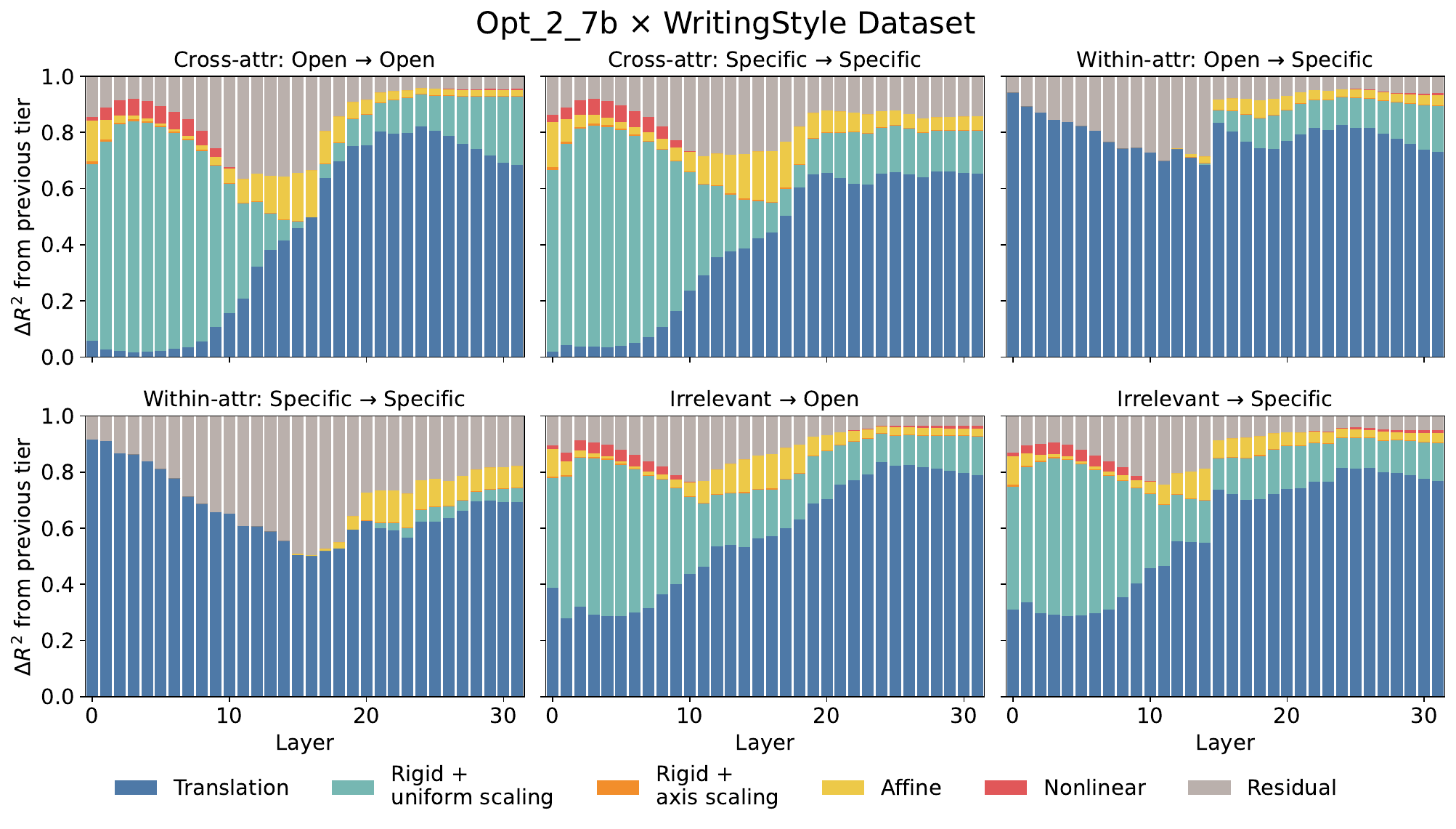}\\[2pt]
  \includegraphics[width=\linewidth]{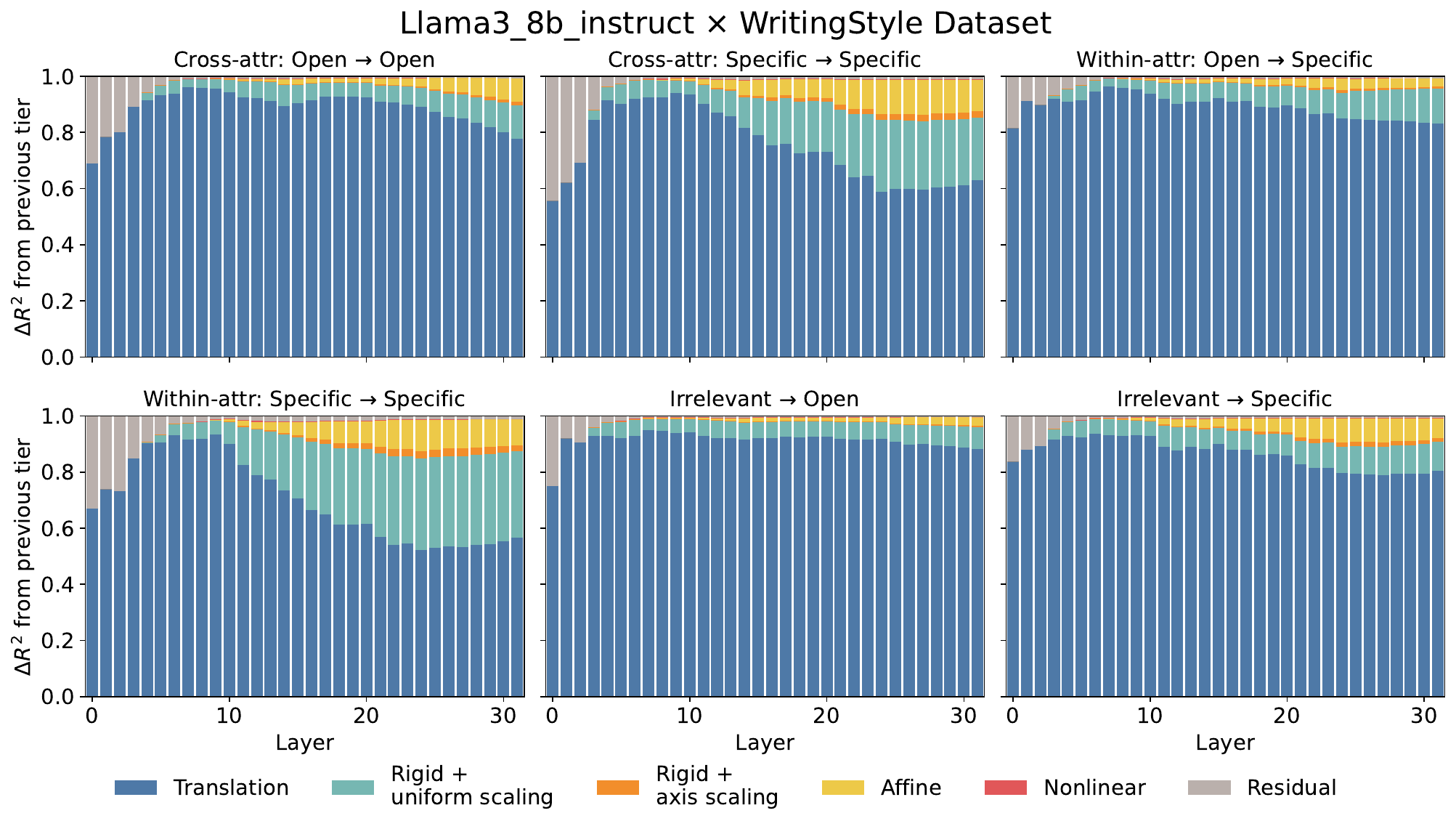}\\[2pt]
  \includegraphics[width=\linewidth]{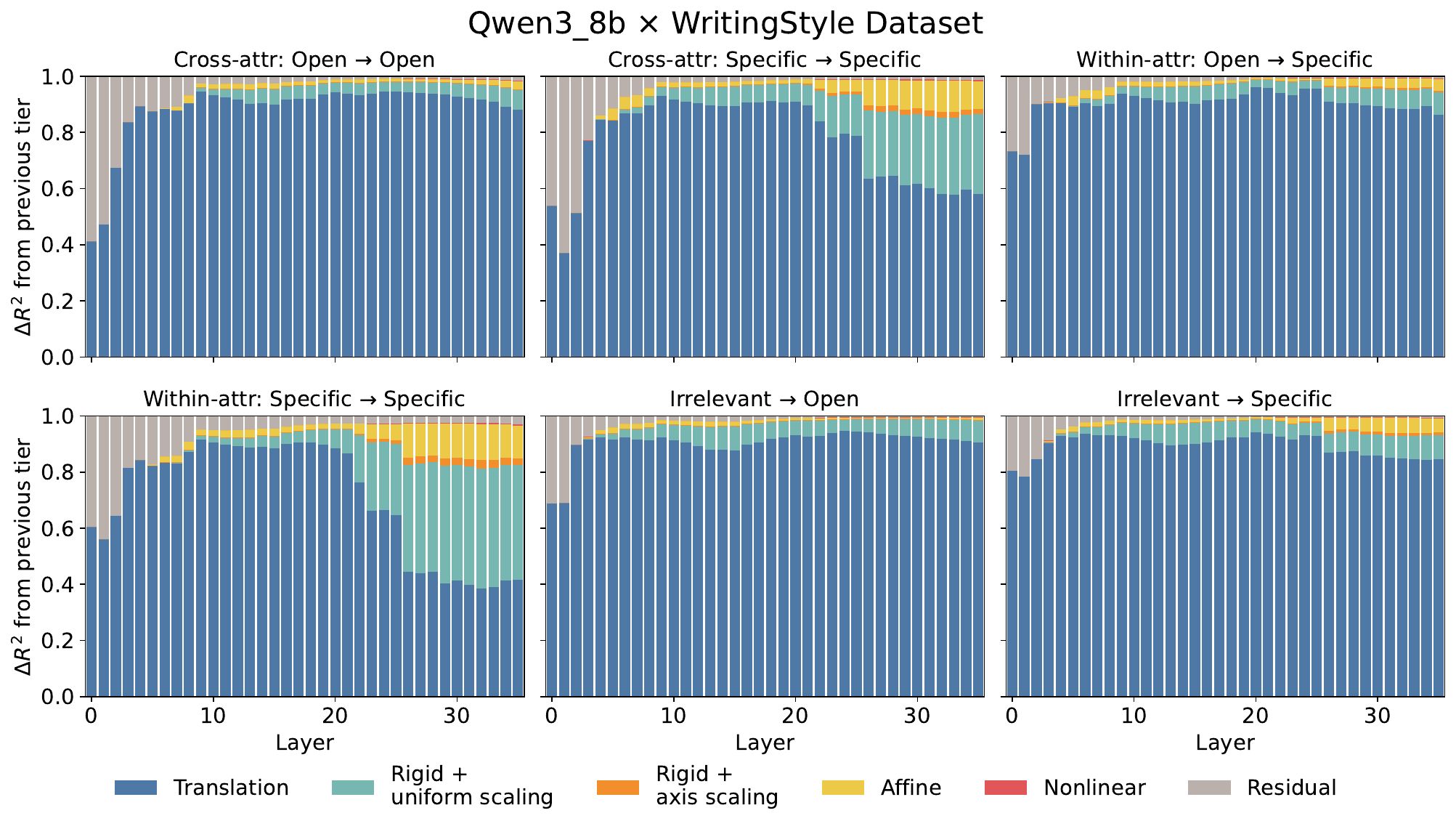}
  \caption{Incremental $R^{2}$ of each transformation for WritingStyle across
  layers for OPT-2.7B (top), Llama-3-8B-Instruct (middle), Qwen3-8B
  (bottom).}
  \label{fig:r2bar_writingstyle}
\end{figure}

\begin{figure}[h]
  \centering
  \includegraphics[width=\linewidth]{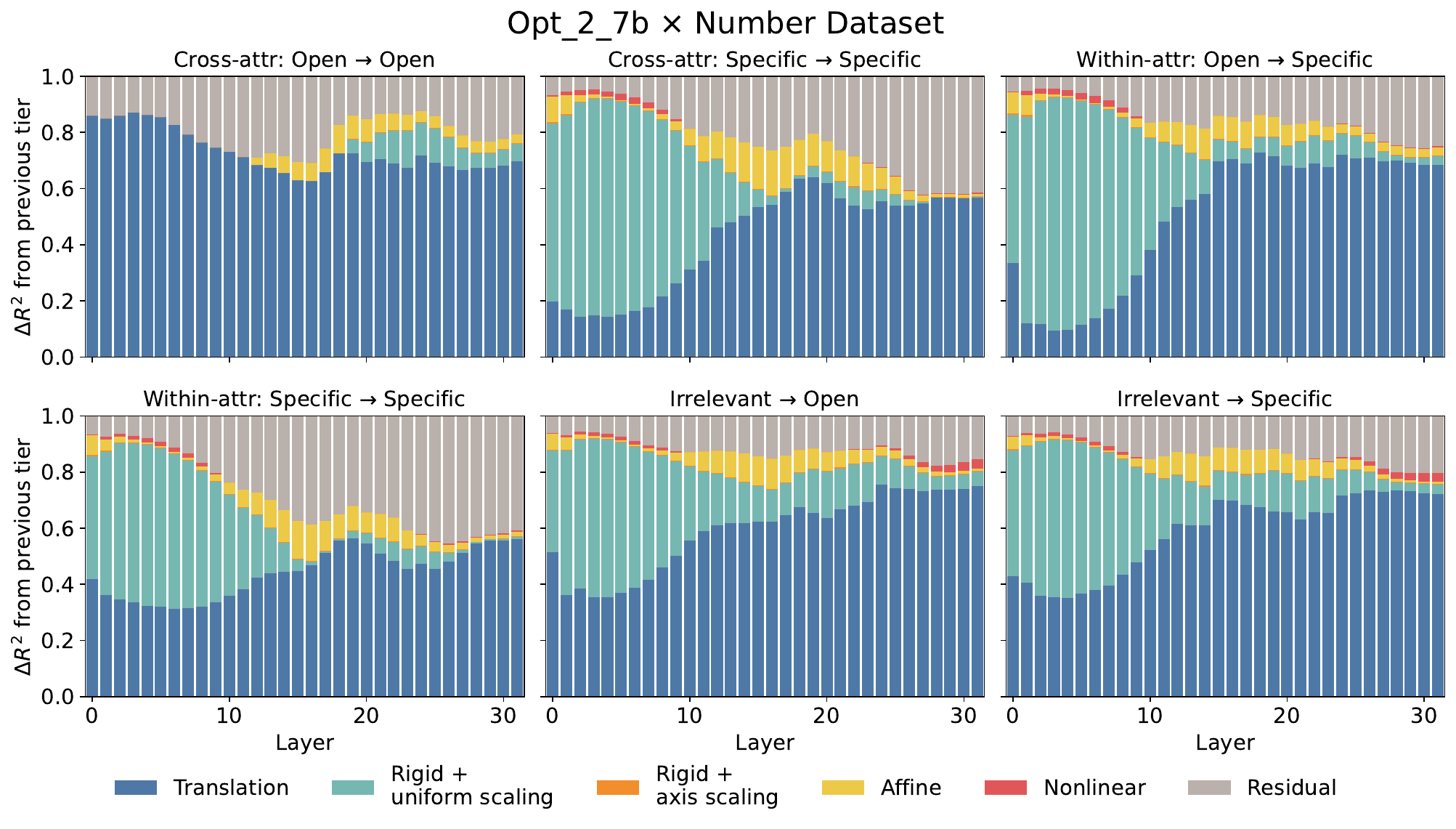}\\[2pt]
  \includegraphics[width=\linewidth]{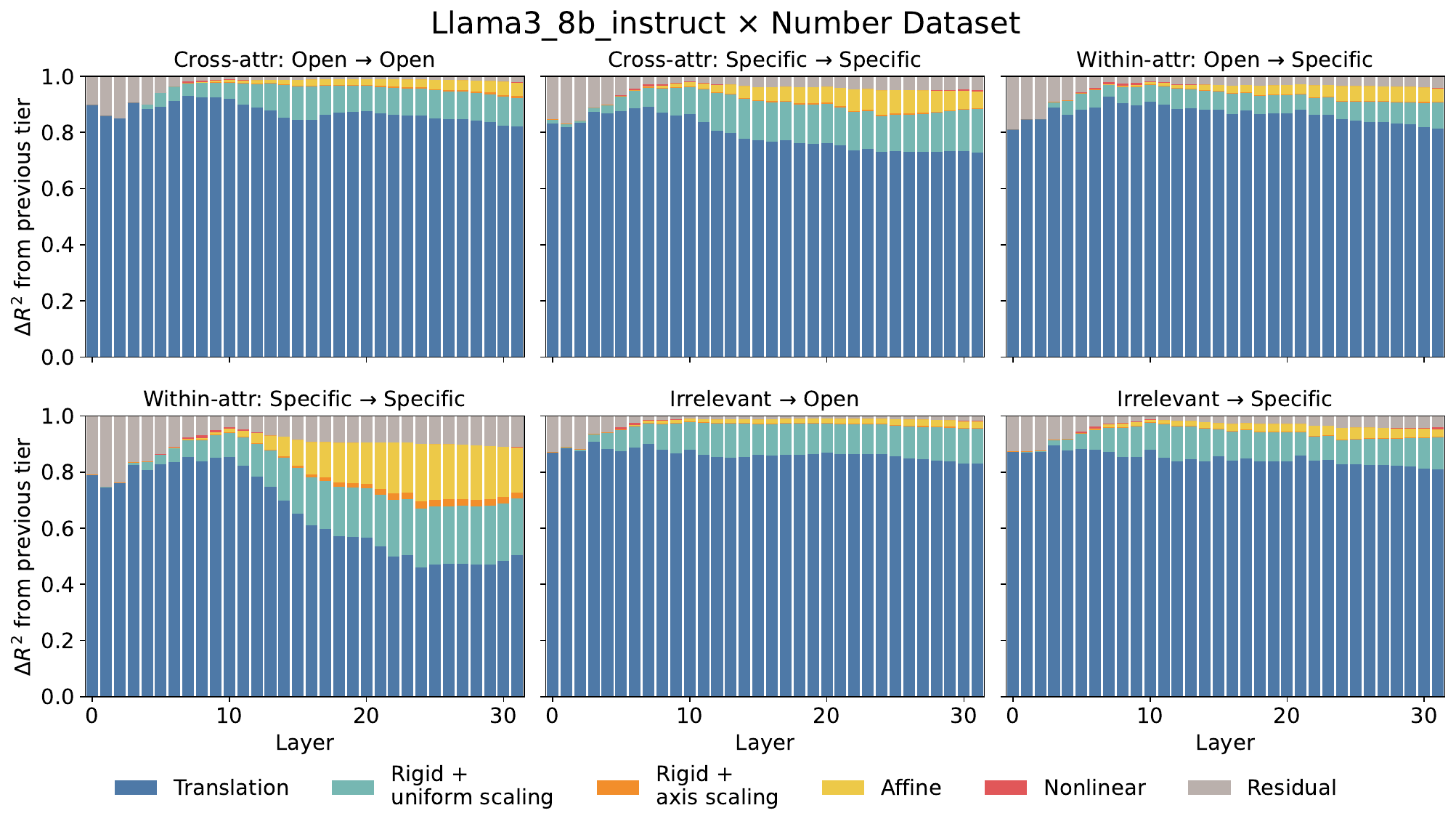}\\[2pt]
  \includegraphics[width=\linewidth]{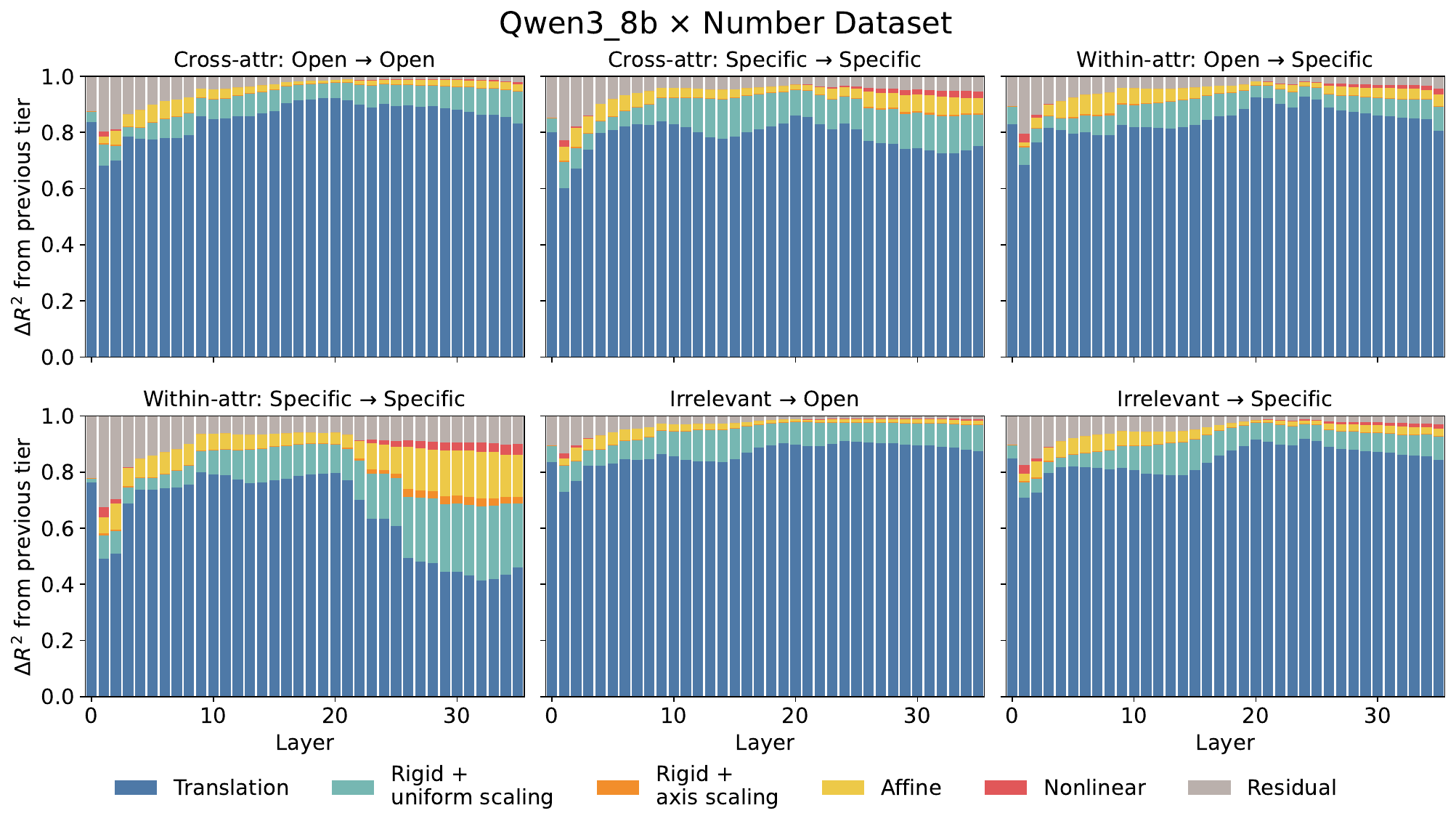}
  \caption{Incremental $R^{2}$ of each transformation for Number across layers
  for OPT-2.7B (top), Llama-3-8B-Instruct (middle), Qwen3-8B (bottom).}
  \label{fig:r2bar_number}
\end{figure}

\begin{figure}[h]
  \centering
  \includegraphics[width=\linewidth]{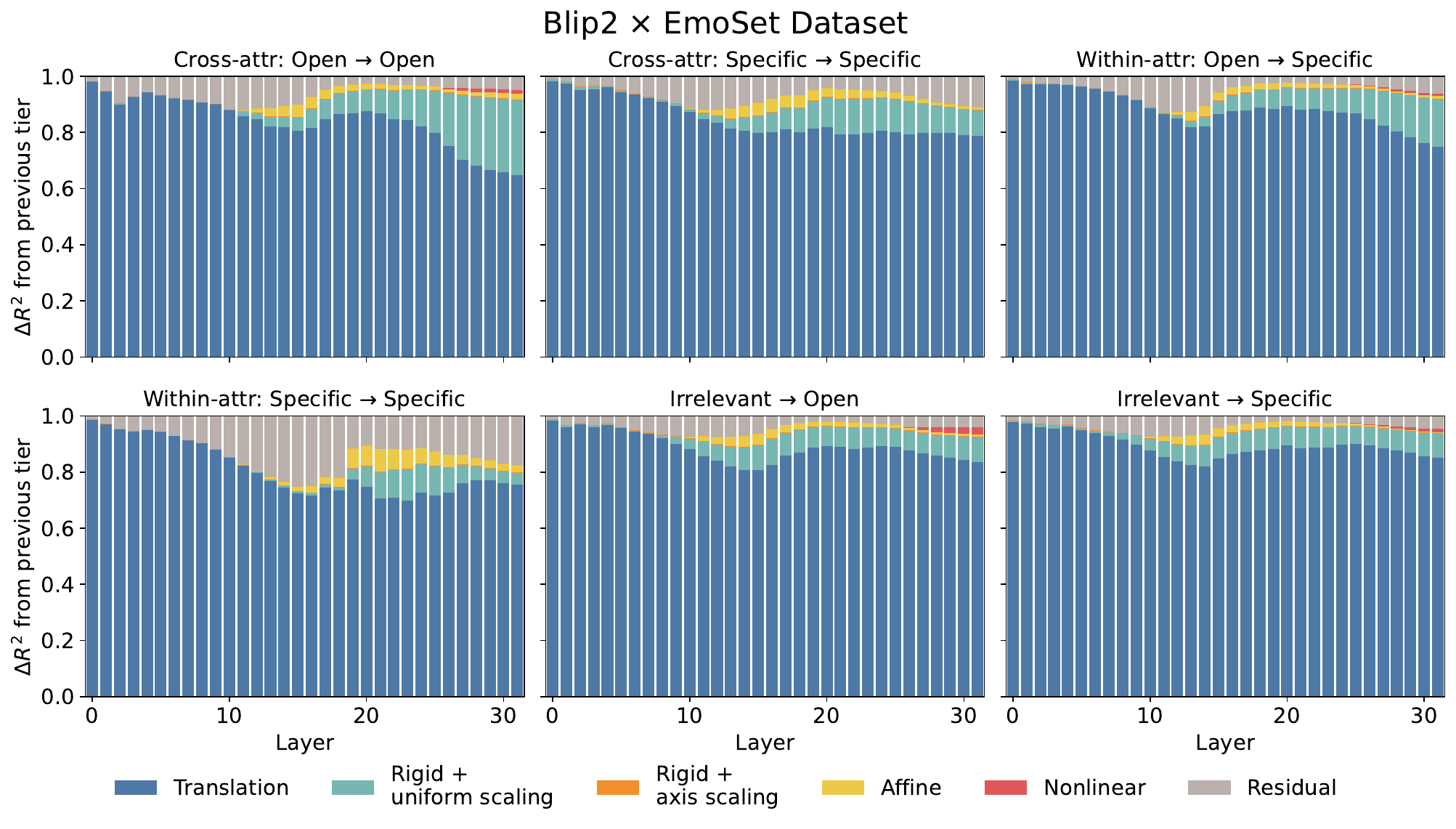}\\[2pt]
  \includegraphics[width=\linewidth]{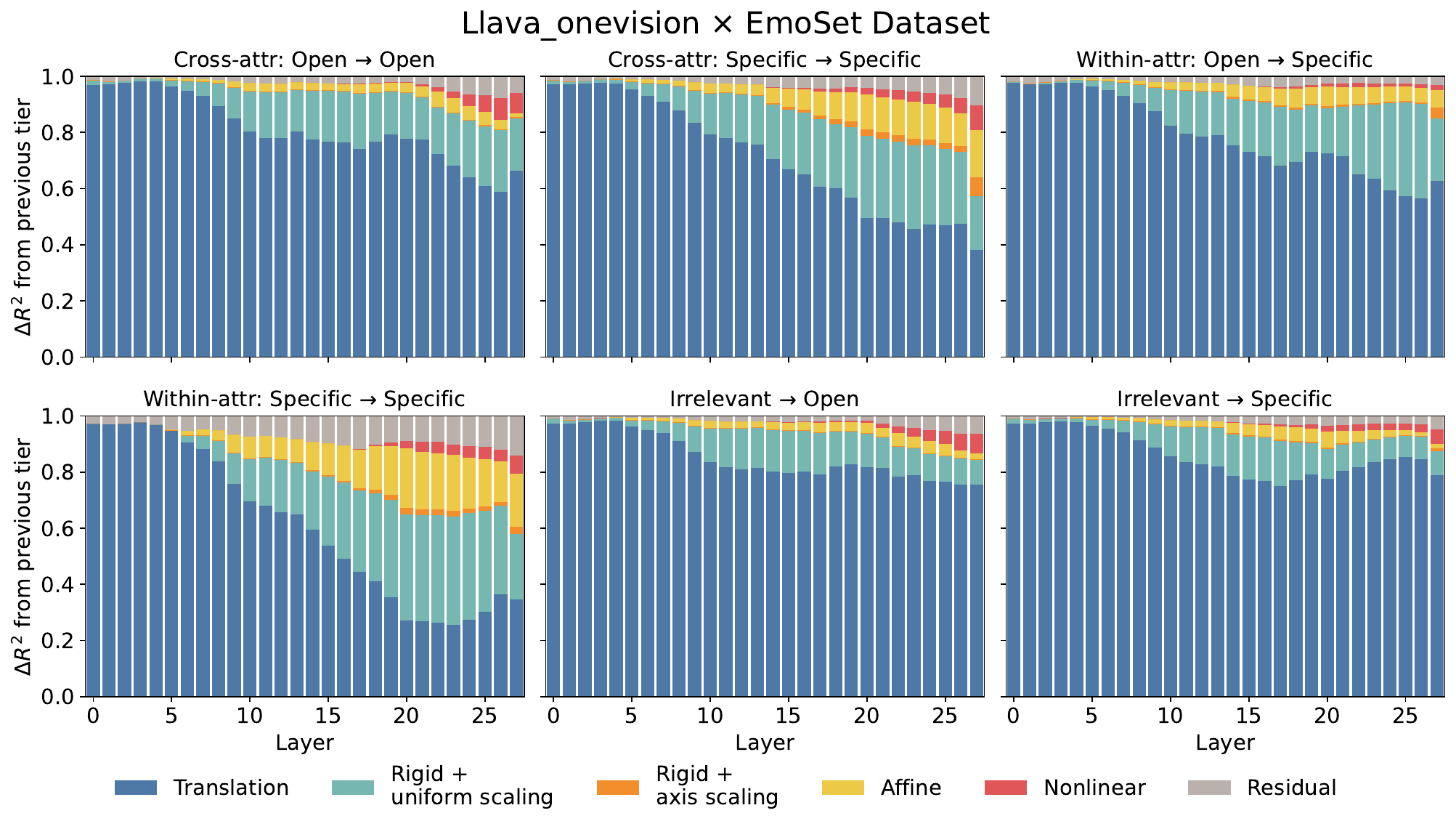}\\[2pt]
  \includegraphics[width=\linewidth]{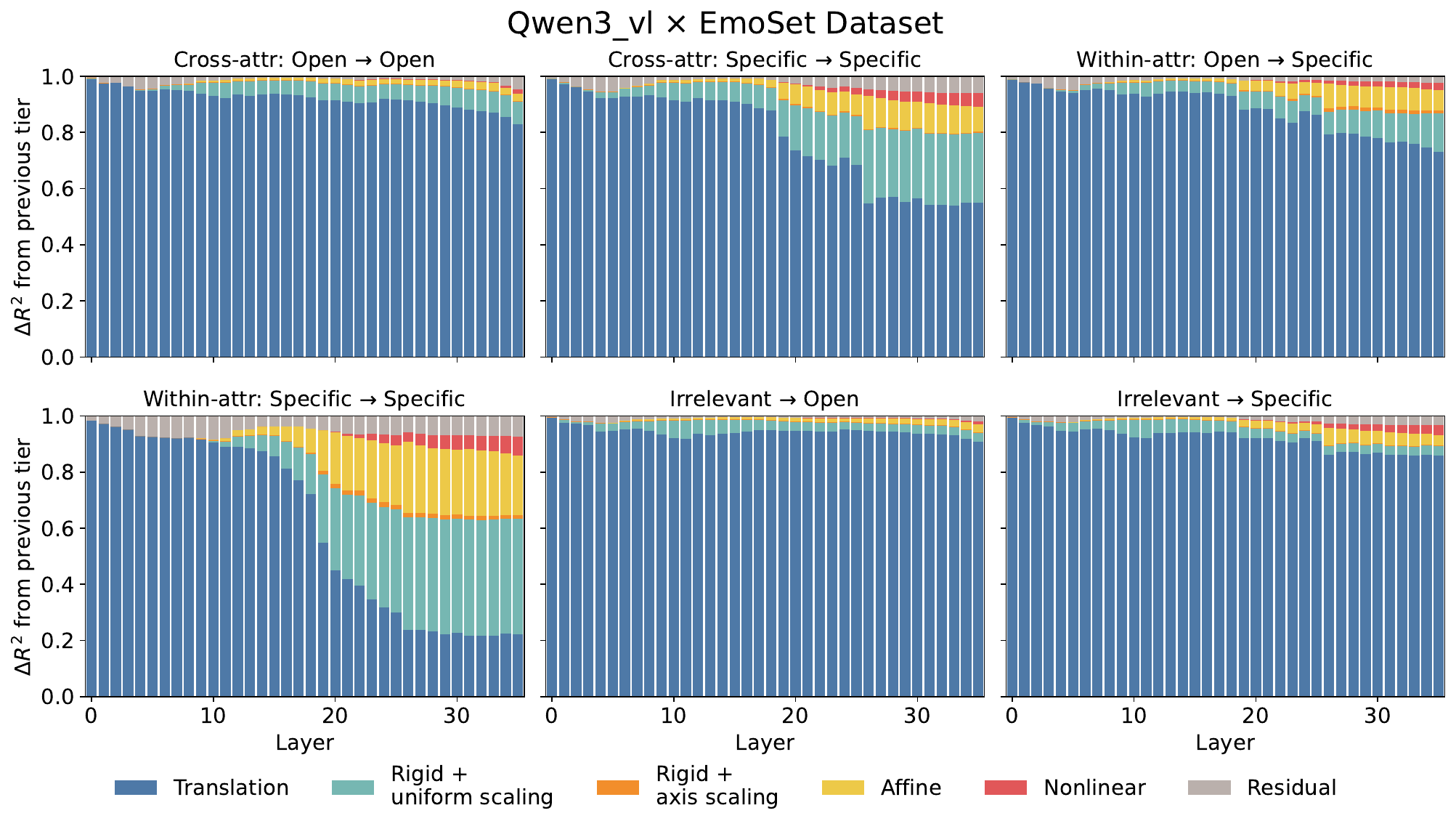}
  \caption{Incremental $R^{2}$ of each transformation for EmoSet across layers
  for BLIP-2 (top), LLaVA-OneVision-7B (middle), Qwen3-VL-8B (bottom).}
  \label{fig:r2bar_emoset}
\end{figure}

\begin{figure}[h]
  \centering
  \includegraphics[width=\linewidth]{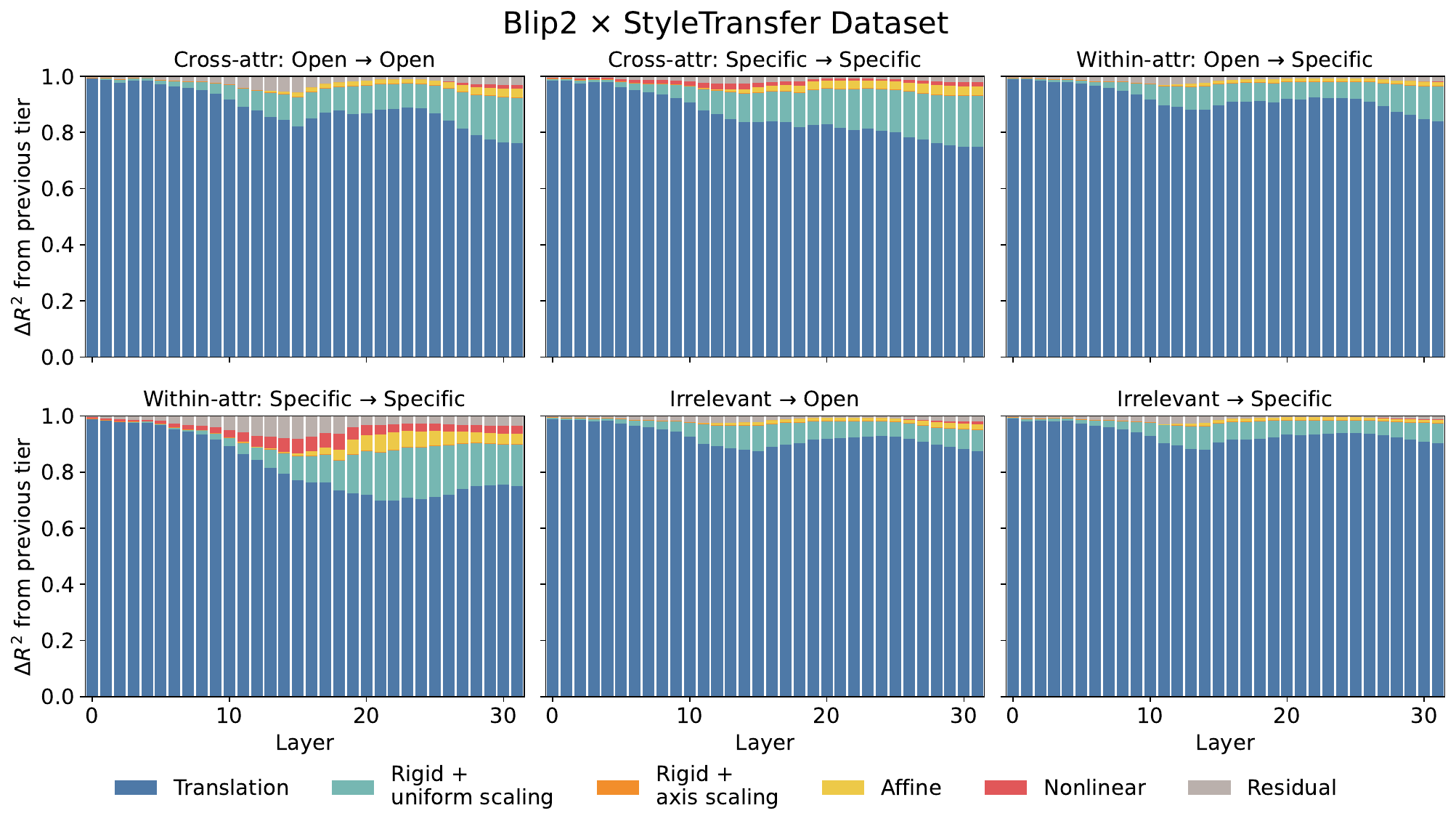}\\[2pt]
  \includegraphics[width=\linewidth]{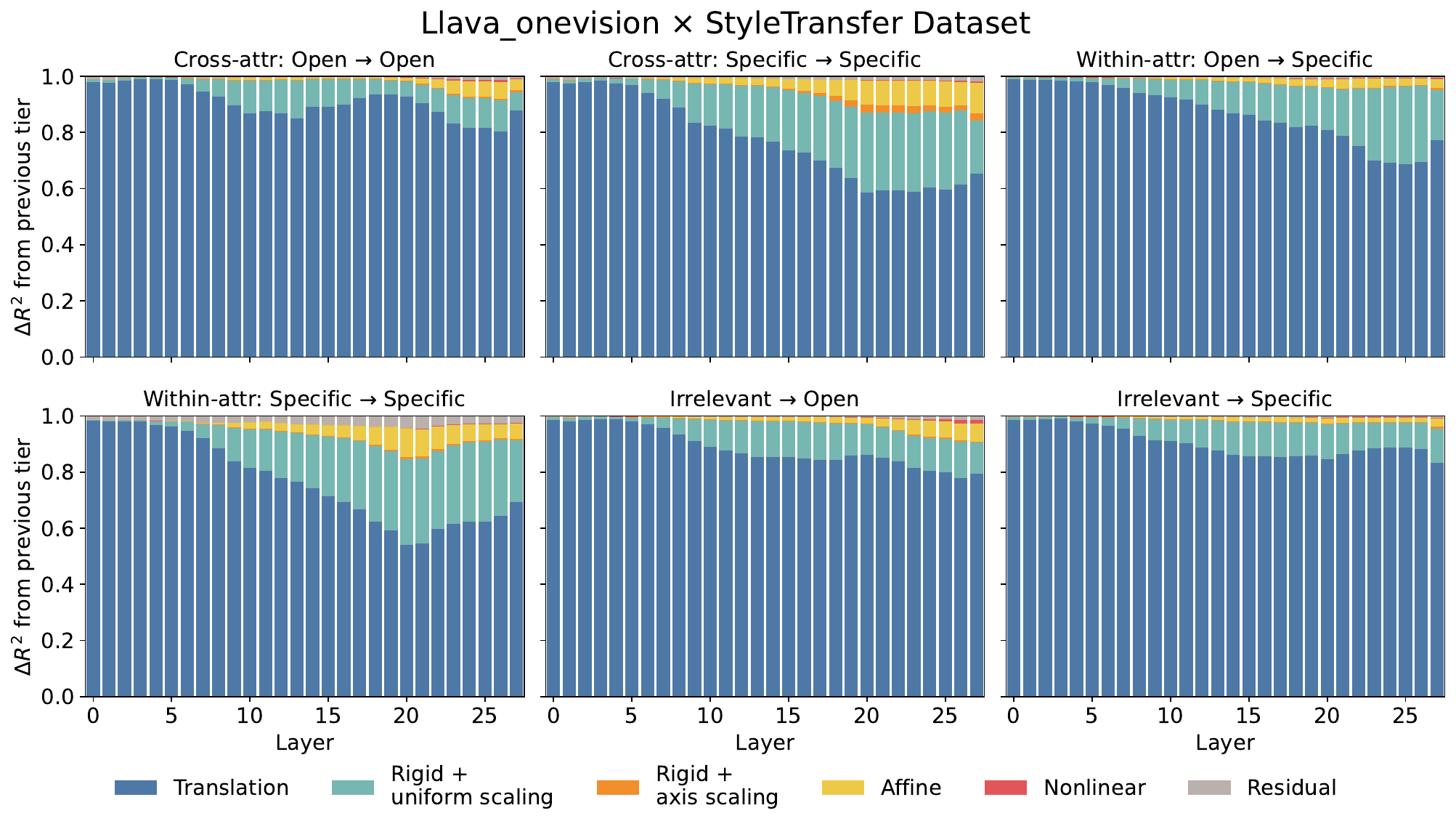}\\[2pt]
  \includegraphics[width=\linewidth]{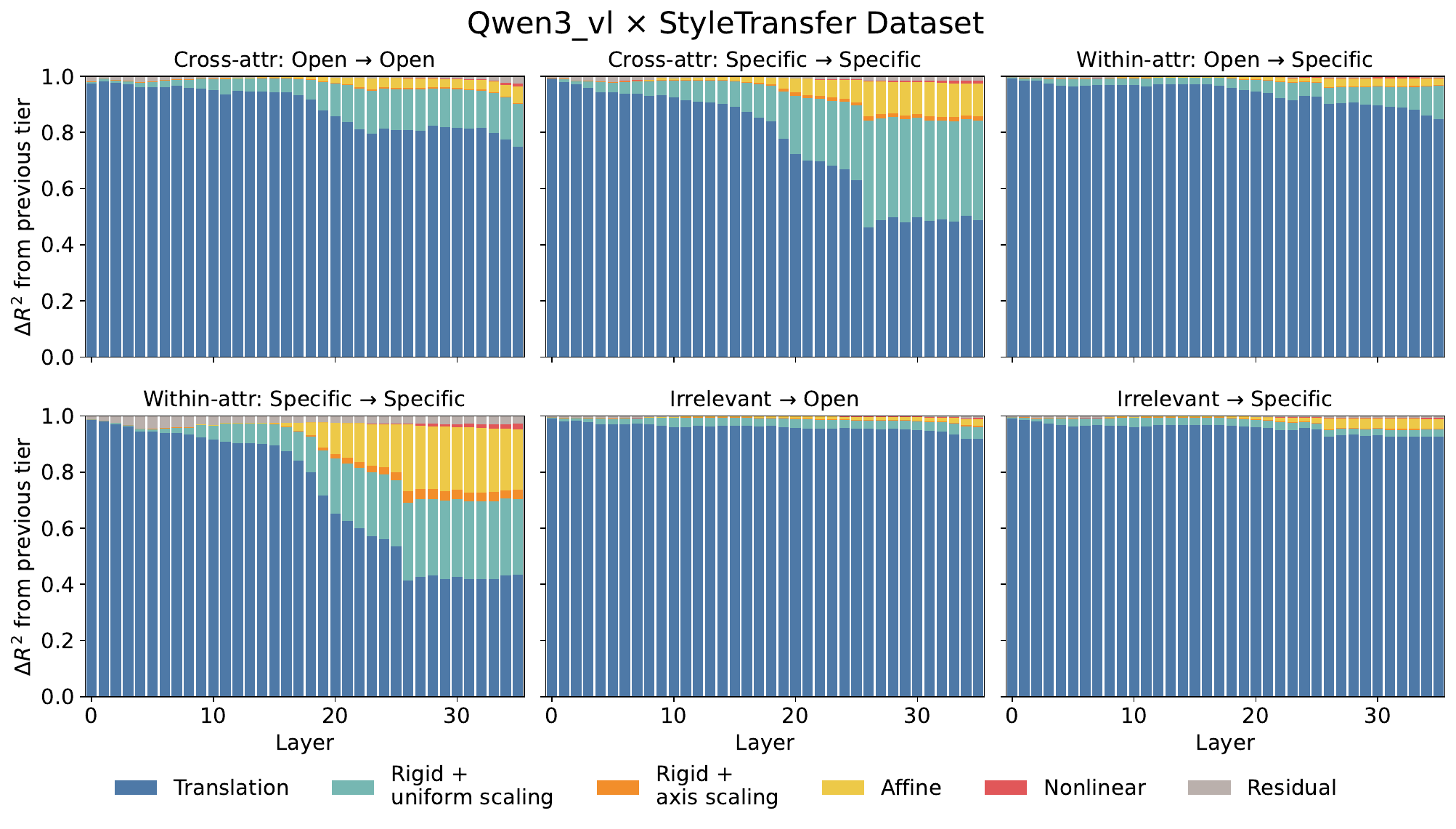}
  \caption{Incremental $R^{2}$ of each transformation for StyleTransfer across
  layers for BLIP-2 (top), LLaVA-OneVision-7B (middle), Qwen3-VL-8B
  (bottom).}
  \label{fig:r2bar_styletransfer}
\end{figure}

\begin{figure}[h]
  \centering
  \includegraphics[width=\linewidth]{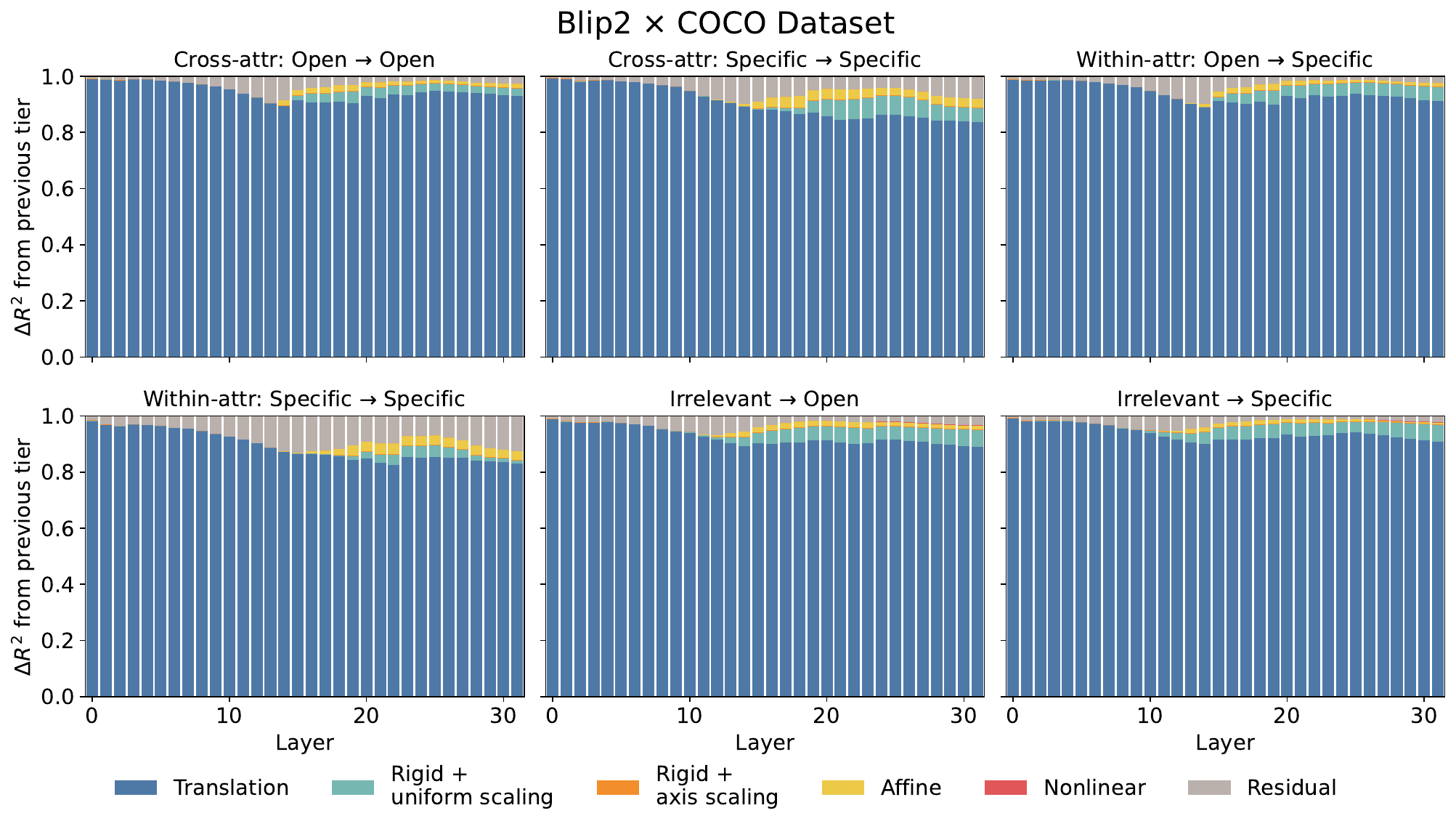}\\[2pt]
  \includegraphics[width=\linewidth]{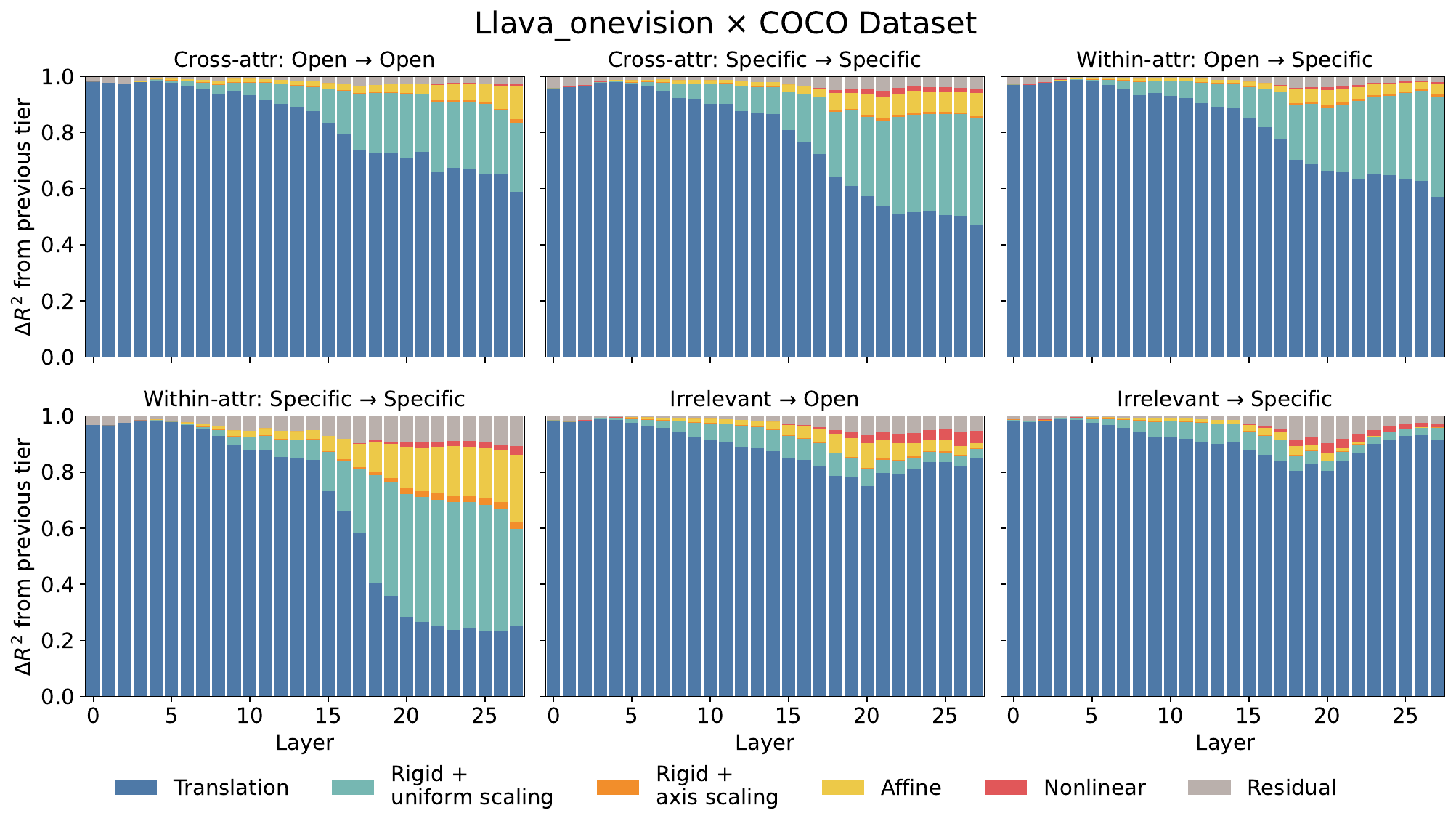}\\[2pt]
  \includegraphics[width=\linewidth]{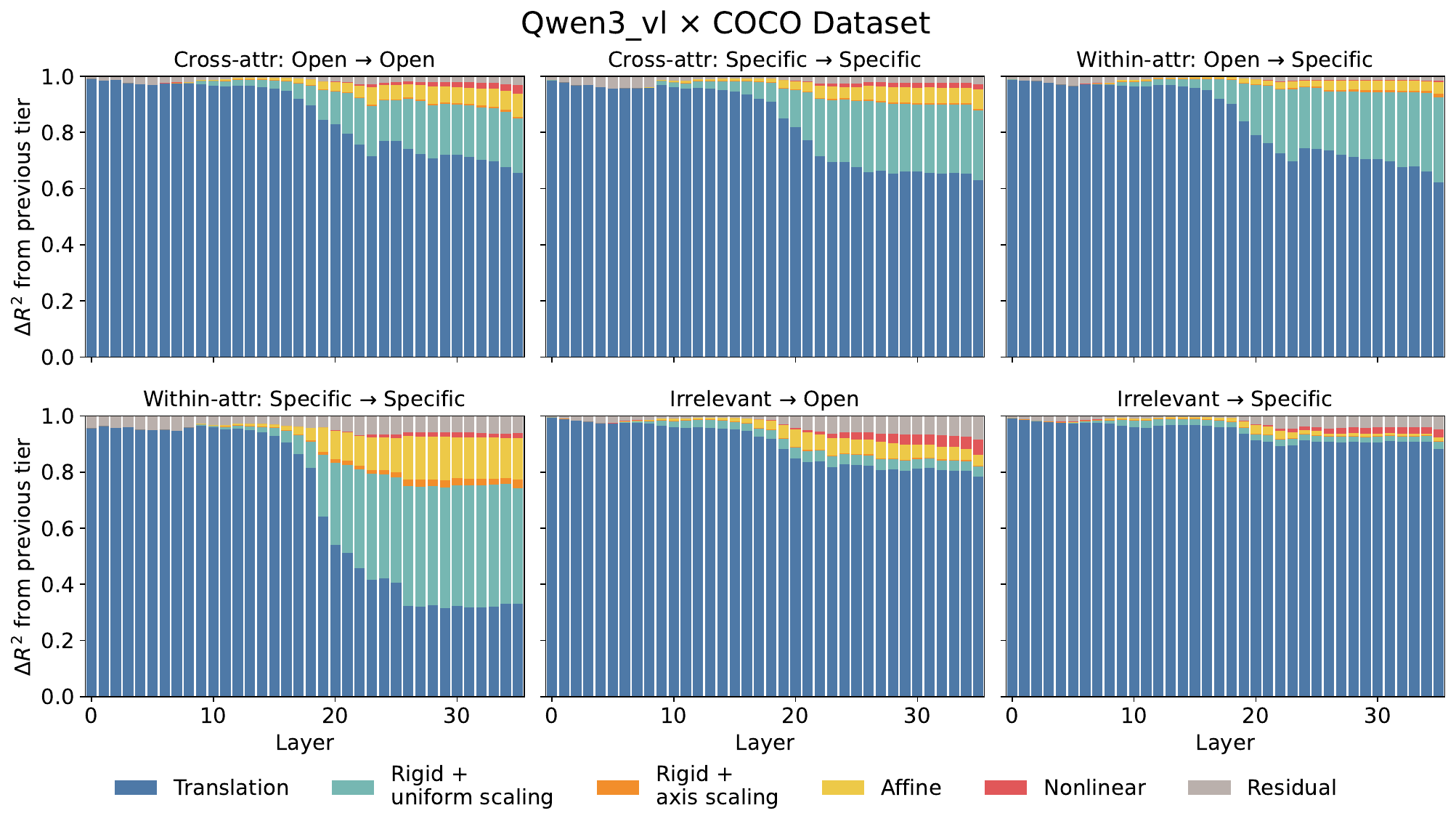}
  \caption{Incremental $R^{2}$ of each transformation for COCO across layers for
  BLIP-2 (top), LLaVA-OneVision-7B (middle), Qwen3-VL-8B (bottom).}
  \label{fig:r2bar_coco}
\end{figure}

\clearpage
\section{Generalization to prompt paraphrases and out-of-distribution (OOD) datasets}
\label{app:para}

We ask whether the fitted transformations \(\widehat{f}_{k}\) capture
task-dependent representational structure that generalizes beyond the
specific prompt--stimulus pairings used for fitting. We evaluate two
forms of generalization: (i) \emph{prompt paraphrasing}, in which the
canonical target prompt \(B\) is replaced by semantically equivalent
rewordings that query the same attribute, and (ii) \emph{input
distribution shift}, in which transformations fitted on one dataset are
evaluated on another dataset.

\subsection*{Prompt paraphrasing}

\paragraph{Setup.}
We fix the source prompt \(A\) and the canonical target prompt \(B\)
from the cross-attribute open--open pairing group (G1). We then construct
three semantic paraphrases \(\{B'_i\}_{i=1}^{3}\) of \(B\), each
preserving the queried attribute while varying the surface form of the
instruction. For each paraphrase \(B'_i\), we fit each tier-\(k\)
transformation \(\widehat{f}_{k}^{(i)}\) on the paraphrased prompt pair
\((\mathbf{X}^{A},\mathbf{X}^{B'_i})\) using the training stimuli, and
evaluate the fitted transformation against the canonical target
representations \(\mathbf{X}^{B}\) on held-out stimuli. We use the same
5-fold cross-validation protocol as in the main experiments
(Section~\ref{sec:experiments}).

For paraphrase \(i\), the cumulative paraphrase-generalization score of
tier \(k\) is

\[
  R^{2,(i)}_{k} =
  \frac{
  \big\|\mathbf{X}^{B}-\mathbf{X}^{A}\big\|_{F}^{2}
  -
  \big\|\mathbf{X}^{B}-\widehat{f}_{k}^{(i)}(\mathbf{X}^{A})\big\|_{F}^{2}
  }{
  \big\|\mathbf{X}^{B}-\mathbf{X}^{A}\big\|_{F}^{2}
  },
\]

where both norms are computed over held-out stimuli and scores are
averaged across folds. As a reference ceiling, we also fit and evaluate
the canonical pair \((\mathbf{X}^{A},\mathbf{X}^{B})\) under the same
cross-validation procedure. If the learned transformation captures the
target attribute independently of the exact wording of \(B\), the
paraphrase-generalization curves should approach the canonical reference
curve.

We evaluate this analysis in one vision-language setting and one
language-only setting: LLaVA-OneVision-7B on StyleTransfer, with
canonical target prompt \(B=\) ``What artistic style does the image
belong to?'' and three paraphrases querying \emph{style}; and
Llama3-8B-Instruct on EmotionalStory, with canonical target prompt
\(B=\) ``What emotion does this text express?'' and three paraphrases
querying \emph{emotion}.

\begin{figure}[h]
  \centering
  \includegraphics[width=\linewidth]{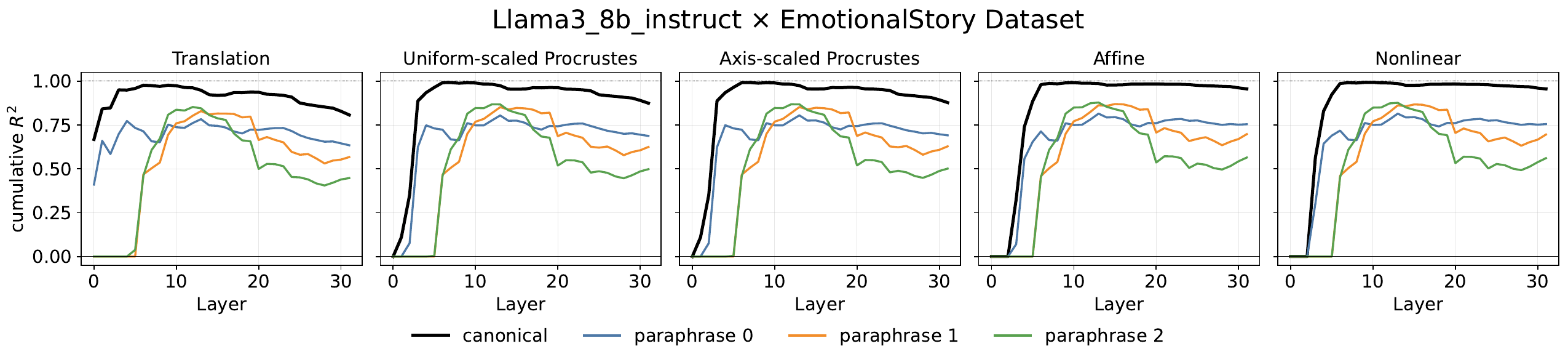}
  \caption{Cumulative cross-validated $R^{2}$ of each transformation under prompt paraphrasing, evaluated on the canonical $\mathbf{X}^{B}$ on held-out stimuli for Llama3-8B-Instruct on EmotionalStory (source $A$\,=\,topic, canonical target $B$\,=\,emotion;
  three paraphrases of emotion). }
  \label{fig:para_llm}
\end{figure}

\begin{figure}[h]
  \centering
  \includegraphics[width=\linewidth]{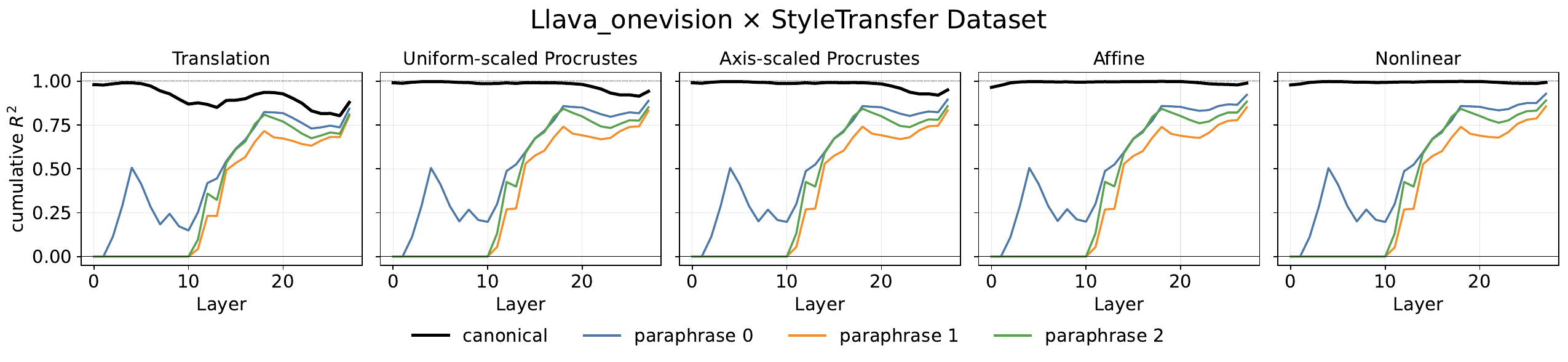}
  \caption{Cumulative cross-validated $R^{2}$ of each transformation under prompt paraphrasing, evaluated on the canonical $\mathbf{X}^{B}$ on held-out stimuli
  for LLaVA-OneVision-7B on StyleTransfer(source $A$\,=\,scene,
  canonical target $B$\,=\,style; three paraphrases of style).}
  \label{fig:para_vlm}
\end{figure}

\paragraph{Results.}
For Llama3-8B-Instruct on EmotionalStory
(Fig.~\ref{fig:para_llm}), the paraphrase-generalization curves closely
track the canonical reference curve, except in the earliest layers. This
suggests that, across most layers, the fitted transformations capture a
prompt-induced representational change that is largely stable across
semantically equivalent rewordings of the target prompt. For
LLaVA-OneVision-7B on StyleTransfer (Fig.~\ref{fig:para_vlm}), the
paraphrase curves approach the canonical reference primarily in deeper
layers. This indicates that paraphrase-invariant transformations emerge
more strongly at later stages of LLaVA, whereas
earlier layers remain more sensitive to the surface form of the prompt.

\subsection*{Evaluation on out-of-distribution (OOD) datasets}

\paragraph{Setup.}
We next test whether transformations fitted on one stimulus distribution
generalize to a distinct dataset from the same modality. For each
\((\mathrm{model}, D_{\mathrm{src}}, D_{\mathrm{tgt}})\) triple, we fit
each tier-\(k\) transformation \(\widehat{f}_{k}^{\mathrm{src}}\) using
all stimuli from the source dataset \(D_{\mathrm{src}}\). We then apply
the fitted transformation to all stimuli in the target dataset
\(D_{\mathrm{tgt}}\) under the same prompt pair and compute the
per-tier incremental contribution \(\Delta R^{2}_{k}\). As in the
paraphrase analysis, we evaluate one vision-language setting
(LLaVA-OneVision-7B: StyleTransfer \(\leftrightarrow\) EmoSet) and one
language-only setting (Llama3-8B-Instruct: WritingStyle
\(\leftrightarrow\) EmotionalStory).

\begin{figure}[h]
  \centering
  \includegraphics[width=0.95\linewidth]{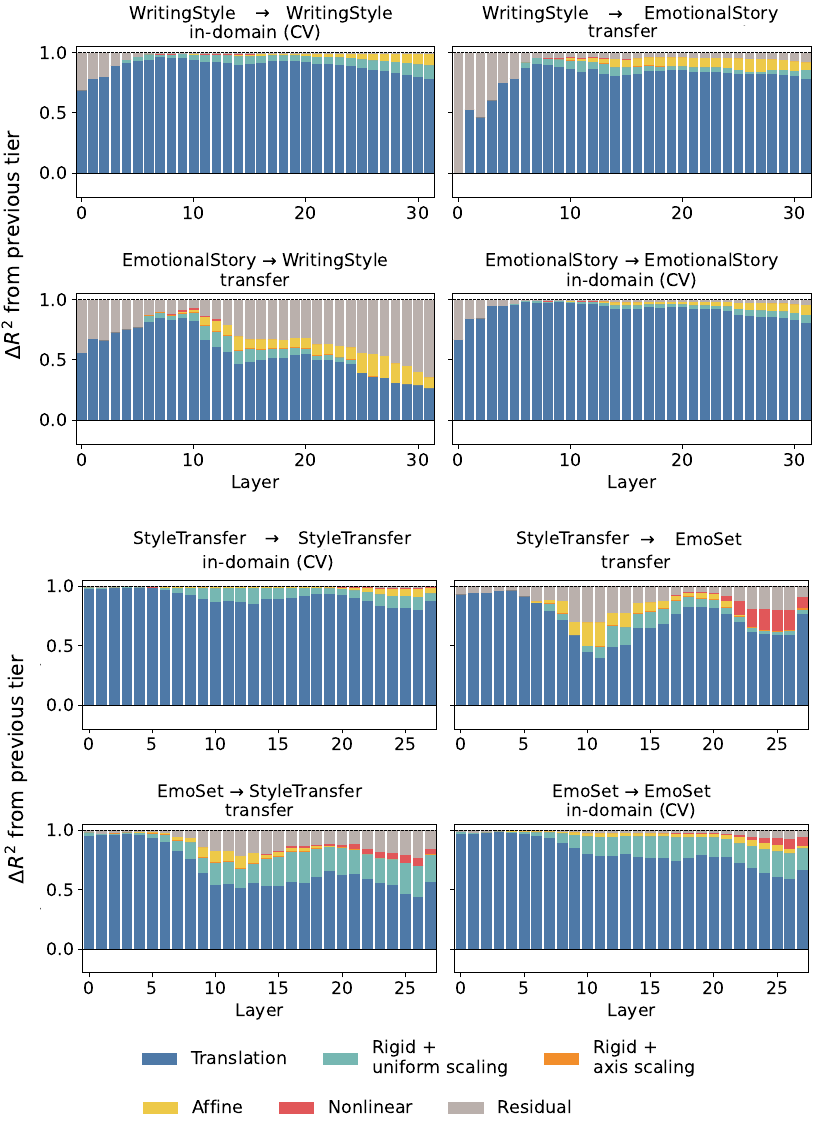}
  \caption{
  Out-of-distribution (OOD) generalization of the nested geometric
  decomposition. Top: Llama3-8B-Instruct. Bottom: LLaVA-OneVision-7B.
  Rows denote the source dataset used to fit the transformation, and
  columns denote the target dataset used for evaluation. Diagonal panels
  show matched in-domain 5-fold cross-validation; off-diagonal panels
  show cross-dataset transfer. Each bar shows the per-layer stacked
  decomposition of incremental \(\Delta R^{2}_{k}\) into transformation
  tiers and residual for cross-attribute open--open prompt pairs.
  }
  \label{fig:cross_dataset}
\end{figure}

For comparison, the diagonal entries in Fig.~\ref{fig:cross_dataset}
show matched in-domain performance estimated by 5-fold cross-validation
within each dataset, whereas the off-diagonal entries show cross-dataset
transfer from \(D_{\mathrm{src}}\) to \(D_{\mathrm{tgt}}\).

\paragraph{Results.}
Transformations fitted on one dataset retain substantial explanatory
power when transferred to a different dataset from the same modality, but
the decomposition across transformation tiers changes under distribution
shift (Fig.~\ref{fig:cross_dataset}). In the language-only setting,
transfer from EmotionalStory to WritingStyle shows the strongest
degradation: the translation component decreases markedly in deeper
layers relative to the in-domain condition, with the loss primarily
redistributed to the affine tier and the residual. The reverse direction,
WritingStyle to EmotionalStory, transfers more robustly, suggesting an
asymmetry in how dataset-specific structure contributes to the fitted transformation.

In the vision-language setting, the in-domain decompositions are
dominated by translation across layers. Under cross-dataset transfer,
however, a larger fraction of the explained and unexplained variance is
assigned to affine, nonlinear, and residual components. Thus,
while a substantial component of the fitted transformation is shared
across datasets, the relative contribution of simple translation versus
higher-order transformations remains dataset-dependent.

Together with the paraphrase analysis, these results indicate that the
nested geometric hierarchy captures prompt-dependent representational
structure that is partly invariant to prompt rewording and partly
transferable across input distributions. At the same time, the tier-wise
allocation of variance is sensitive to the stimulus distribution,
suggesting that prompt-induced representational changes contain both
task-level and stimulus-distribution-specific components.

\clearpage
\section{Dimensionality of Procustes alignment}
\label{app:rank10}

\begin{figure}[h]
  \centering
  \includegraphics[width=\linewidth]{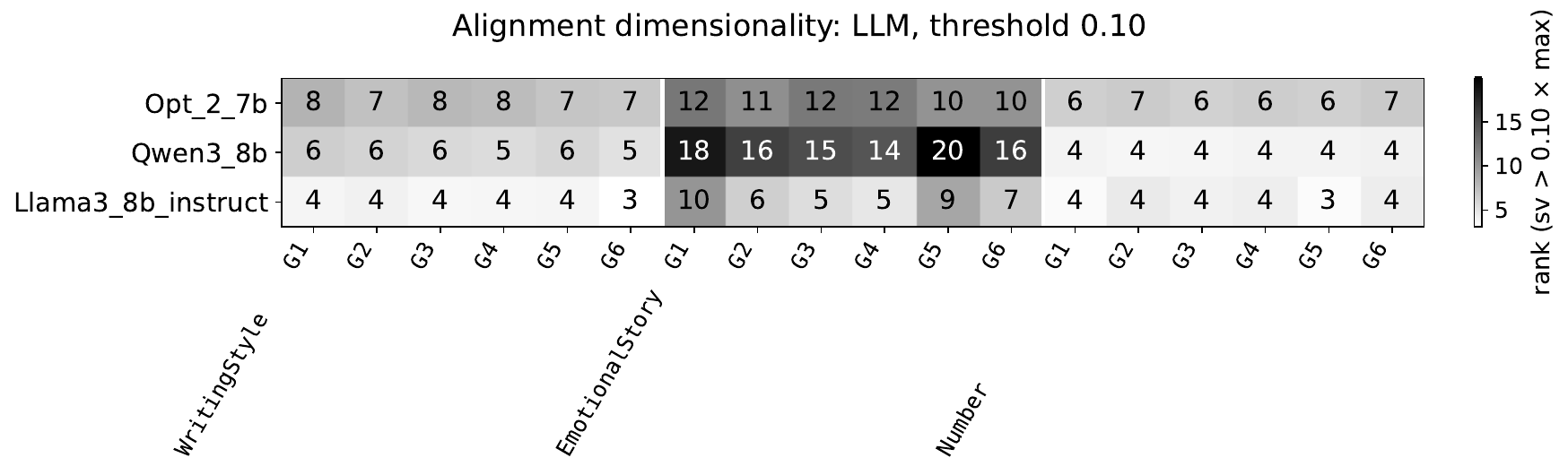}\\[2pt]
  \includegraphics[width=\linewidth]{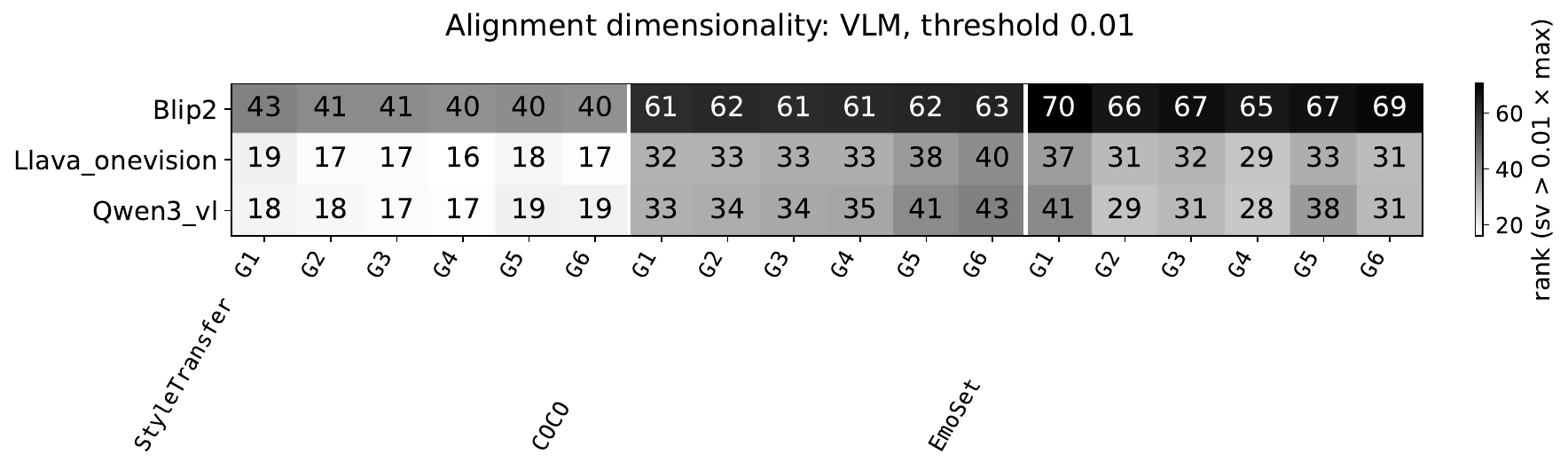}\\[2pt]
  \includegraphics[width=\linewidth]{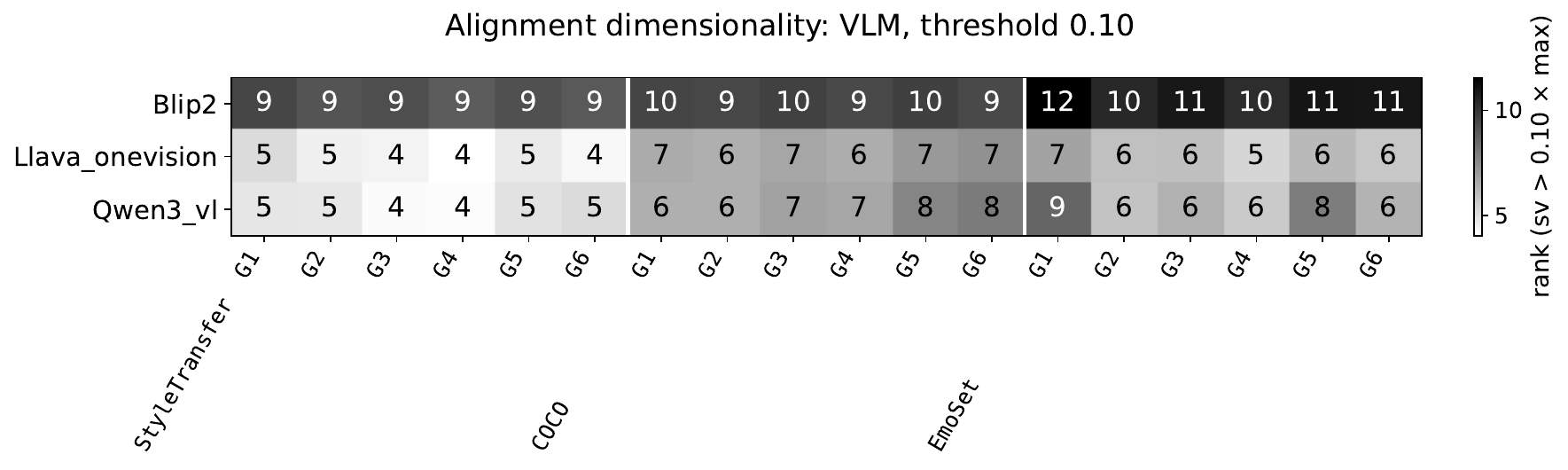}
  \caption{Alignment dimensionality. (top) the centered cross-prompt cross-covariance at the stricter 10\%-of-$\sigma_{\max}$ threshold (LLM, all six prompt-pair groups). (bottom and middke ) llm. two thresholds. the layout across (model $\times$ dataset $\times$
  group) is the same as in Fig.~\ref{fig:strategy_rank}\textbf{b}.}
  \label{fig:rank10}
\end{figure}

Fig.~\ref{fig:strategy_rank}\textbf{b} (bottom) shows the
alignment dimensionality using a 1\%
relative threshold on the singular values of cross-covariance $\widetilde{\mathbf{X}}^{A\,\top}\widetilde{\mathbf{X}}^{B}$ 
($\sigma_{i}>0.01\,\sigma_{\max}$). For completeness we also report
the stricter 10\% threshold
($\sigma_{i}>0.10\,\sigma_{\max}$) in
Fig.~\ref{fig:rank10} for LLMs. 

\clearpage
\section{Example generated texts after interventions}
\label{app:behavior}

We provide qualitative examples of model generations after replacing an
internal representation with the output of each fitted transformation tier.
For a stimulus \(s_i\) and layer \(\ell\), we run the model with prompt
\(A\), replace the residual-stream activation at the final input-prompt
token by \(\widehat{f}_{k}(\Phi^{A}(s_i))\), and then greedily decode
50--100 tokens from the modified state. We compare seven conditions:
\textbf{prompt \(A\)} without intervention; \textbf{translation};
\textbf{rigid\_uni}, corresponding to an orthogonal transformation plus
uniform scaling; \textbf{rigid\_axis}, corresponding to an orthogonal
transformation plus axis-wise scaling; \textbf{affine};
\textbf{nonlinear}; and \textbf{prompt \(B\)}, an oracle activation
reference in which the prompt-\(A\) residual stream at the same layer and
token is replaced by the held-out representation \(\Phi^{B}(s_i)\). This
oracle condition preserves the remaining prompt-\(A\) context and is
therefore distinct from running the model end-to-end under prompt \(B\).

\paragraph{G1 / LLaVA-OneVision-7B \(\times\) COCO.}

Prompt \(A\): \emph{``Are there people in this image?''} \quad Prompt\(B\): \emph{``How many people are in this image?''} This prompt pair tests whether an intervention can shift the model from binary detection to numerical counting. At the mid-depth layer \(\ell=22\), higher-tier transformations replace the prompt-\(A\) yes/no response format with count-like answers (Table~\ref{tab:texts_g1_llava_coco}).

\begin{table}[!h]
\centering
\caption{G1 / LLaVA-OneVision-7B $\times$ COCO, prompt pair
\emph{detect}$\,\to\,$\emph{count} at $\ell=22$.}
\label{tab:texts_g1_llava_coco}
\footnotesize
\begin{tabular}{@{}c|lllllll@{}}
\toprule
Stim & prompt $A$ & translation & rigid\_uni & rigid\_axis & affine & nonlinear & prompt $B$ \\
\midrule
8  & yes & two  & two  & two  & two   & two   & two  \\
12 & no  & no   & 0    & 0    & 0     & 0     & 0    \\
13 & no  & no   & 0    & 0    & zero  & 0     & zero \\
15 & yes & five & five & five & five  & five  & five \\
25 & yes & four & four & four & four  & four  & four \\
29 & yes & two  & two  & two  & 10    & 10    & 10   \\
41 & yes & one  & one  & one  & one   & one   & one  \\
\bottomrule
\end{tabular}
\end{table}

\paragraph{G1 / Qwen3-VL-8B \(\times\) COCO.}

We use the same detection-to-counting prompt pair and apply interventions

at layer \(\ell=27\) (Table~\ref{tab:texts_g1_qwen_coco}). Because

Qwen3-VL often produces longer descriptive responses, we report only the

first sentence in each cell. Translation and the orthogonal-plus-uniform

scaling tier largely preserve the detection framing of prompt \(A\)

(e.g., ``Yes, there are people \(\ldots\)''), whereas affine and nonlinear

interventions more consistently shift the output toward a numerical

count.

\begin{table}[!h]
\centering
\caption{G1 / Qwen3-VL-8B $\times$ COCO, prompt pair
\emph{detect}$\,\to\,$\emph{count} at $\ell=27$ (first sentence shown
per cell). Cells with identical text within a row are grouped.}
\label{tab:texts_g1_qwen_coco}
\footnotesize
\begin{tabular}{@{}cp{0.85\linewidth}@{}}
\toprule
Stim & Generated text per level group (groups separated by ``$\mid$'') \\
\midrule
25 & \emph{prompt $A$}: ``Yes, there are people in this image.'' $\mid$
     \emph{T, $O_u$, $O_a$}: ``Based on the image provided, yes, there are people in this image.'' $\mid$
     \emph{L, N, prompt $B$}: ``There are four people in this image.'' \\
\midrule
69 & \emph{prompt $A$}: ``Yes, there are people in this image.'' $\mid$
     \emph{T, $O_u$}: ``Based on the image provided, yes, there are people in this image.'' $\mid$
     \emph{$O_a$, L, N, prompt $B$}: ``There are two people in the image.'' \\
\midrule
94 & \emph{prompt $A$}: ``Yes, there are two people in this image.'' $\mid$
     \emph{T}: ``Based on the visual information in the image, yes, there are people in this image.'' $\mid$
     \emph{$O_u$, $O_a$, L, N, prompt $B$}: ``There are two people in this image.'' \\
\midrule
211 & \emph{prompt $A$}: ``Yes, there are people in this image.'' $\mid$
      \emph{T, $O_u$, $O_a$}: ``Based on the image provided, yes, there are people in this image.'' $\mid$
      \emph{L, N, prompt $B$}: ``There are three people in this image.'' \\
\midrule
237 & \emph{prompt $A$}: ``Yes, there are people in this image.'' $\mid$
      \emph{T}: ``Based on the image provided, yes, there are people in this image.'' $\mid$
      \emph{$O_u$, $O_a$, L, N, prompt $B$}: ``There are five people in this image.'' \\
\bottomrule
\end{tabular}
\end{table}

\paragraph{G5 / LLaVA-OneVision-7B \(\times\) COCO: irrelevant ``capital'' prompt.}
Prompt \(A\): \emph{``What is the capital of France?''} \quad Prompt
\(B\): \emph{``How many people are in this image?''}
This prompt pair tests whether an intervention can induce an
image-dependent counting response even when the source prompt is
task-irrelevant. Without intervention, the model answers the source
question with ``Paris.'' After applying intermediate- and higher-tier
transformations, the generated response shifts toward the count requested
by prompt \(B\) (Table~\ref{tab:texts_g5_llava_capital}). This suggests
that the fitted transformations can inject target-task structure beyond
the semantic content explicitly requested by the source prompt.

\begin{table}[!h]
\centering
\caption{G5 / LLaVA-OneVision-7B $\times$ COCO, prompt pair
\emph{capital}$\,\to\,$\emph{count} at $\ell=22$.}
\label{tab:texts_g5_llava_capital}
\footnotesize
\begin{tabular}{@{}c|lllllll@{}}
\toprule
Stim & prompt $A$ & translation & rigid\_uni & rigid\_axis & affine  & nonlinear & prompt $B$ \\
\midrule
8  & paris & two & two   & two  & two   & two  & two  \\
12 & paris & two & zero  & zero & 0     & 0    & 0    \\
13 & paris & two & two   & two  & 0     & zero & zero \\
15 & paris & two & five  & five & five  & five & five \\
25 & paris & two & four  & four & four  & four & four \\
29 & paris & two & two   & two  & two   & two  & 10   \\
41 & paris & two & One   & One  & three & one  & one  \\
\bottomrule
\end{tabular}
\end{table}

\paragraph{G5 / LLaVA-OneVision-7B \(\times\) COCO: irrelevant ``arithmetic'' prompt.}
Prompt \(A\): \emph{``What is 2+2?''} \quad Prompt \(B\):
\emph{``How many people are in this image?''}
This condition provides a second task-irrelevant source prompt. The
unmodified model produces the arithmetic answer ``4,'' whereas
interventions can shift the response toward an image-dependent count
(Table~\ref{tab:texts_g5_llava_arith}). 

\begin{table}[!h]
\centering
\caption{G5 / LLaVA-OneVision-7B $\times$ COCO, prompt pair
\emph{arithmetic}$\,\to\,$\emph{count} at $\ell=22$.}
\label{tab:texts_g5_llava_arith}
\footnotesize
\begin{tabular}{@{}c|lllllll@{}}
\toprule
Stim & prompt $A$ & translation & rigid\_uni & rigid\_axis & affine & nonlinear & prompt $B$ \\
\midrule
8  & 4 & two & two  & two  & two  & two   & two  \\
12 & 4 & two & zero & zero & 0    & 0     & 0    \\
13 & 4 & two & 0    & 0    & 0    & 0     & 0    \\
15 & 4 & two & five & five & five & five  & five \\
25 & 4 & 4   & 4    & 4    & 4    & 4     & 4    \\
29 & 4 & two & two  & two  & two  & three & 10   \\
41 & 4 & two & 1    & two  & 1    & 2     & 1    \\
\bottomrule
\end{tabular}
\end{table}


\clearpage
\section{Asset licenses and credits}
\label{app:licenses}

\paragraph{Pretrained models.} All models used in this paper are
publicly available HuggingFace checkpoints, cited via their original
papers and listed with the URL of the specific revision and the
license under which we used them:
\begin{itemize}
  \item LLaVA-OneVision-7B \citep{li2024llavaonevision} --
        \url{https://huggingface.co/llava-hf/llava-onevision-qwen2-7b-ov-hf}
        -- Apache 2.0.
  \item Qwen3-VL-8B-Instruct \citep{bai2025qwen3vl} --
        \url{https://huggingface.co/Qwen/Qwen3-VL-8B-Instruct}
        -- Apache 2.0.
  \item Qwen3-8B \citep{yang2025qwen3} --
        \url{https://huggingface.co/Qwen/Qwen3-8B}
        -- Apache 2.0.
  \item BLIP-2 (OPT-2.7B backbone) \citep{li2023blip2} --
        \url{https://huggingface.co/Salesforce/blip2-opt-2.7b}
        -- BSD-3-Clause.
  \item Meta-Llama-3-8B-Instruct \citep{grattafiori2024llama3} --
        \url{https://huggingface.co/meta-llama/Meta-Llama-3-8B-Instruct}
        -- Meta Llama 3 Community License Agreement.
  \item OPT-2.7B \citep{zhang2022opt} --
        \url{https://huggingface.co/facebook/opt-2.7b}
        -- OPT-175B License Agreement (research use; the same agreement
        covers all OPT model sizes).
\end{itemize}

\paragraph{Image datasets.}
\begin{itemize}
  \item COCO val2017 \citep{lin2014coco} -- annotations released under
        CC BY 4.0; the underlying images are subject to Flickr Terms
        of Use. We use a 1{,}000-image subset balanced by
        supercategory.
  \item EmoSet \citep{yang2023emoset} -- non-commercial research use
        only, no redistribution permitted; used in compliance with the
        dataset's terms of use. We use a 1{,}600-image
        $8\!\times\!4$ emotion-by-content subset.
  \item StyleTransfer \citep{boger2025style} -- 1{,}920 photographs
        rendered in seven styles, made available alongside the
        published paper in \emph{Nature Human Behaviour} 9 (2025)
        2497--2509.
\end{itemize}

\paragraph{Curated text stimulus sets.} The three LLM prompt-pair
sets (\textsc{EmotionalStory}, \textsc{WritingStyle}, \textsc{Number})
were authored by us. They consist of factual, style, and numeric
prompts and contain no personal or sensitive content. They will be
released alongside the analysis code under CC BY 4.0.

\clearpage  

\end{document}